\documentclass[10pt,twocolumn,letterpaper]{article}

\usepackage[pagenumbers]{cvpr} %

\usepackage[dvipsnames]{xcolor}

\usepackage{times}

\usepackage{epsfig}
\usepackage{graphicx}
\usepackage{amsmath}
\usepackage{amssymb}
\usepackage{xcolor} %

\definecolor{lightyellow}{RGB}{255,255,180} %

\usepackage{colortbl}
\usepackage{booktabs} %
\usepackage{multirow}
\usepackage{makecell}
\usepackage{float}
\usepackage{enumitem} %
\usepackage{arydshln}
\usepackage{bbding}
\usepackage{animate}
\usepackage{grffile}
\usepackage{siunitx} %
\usepackage{tabularx}
\usepackage{bbm}
\usepackage{caption} 
\usepackage{subcaption}
\usepackage{pifont}     %
\usepackage{graphicx}
\usepackage{tikz}
\usepackage{amsthm}

\definecolor{cvprblue}{rgb}{0.21,0.49,0.74}
\usepackage[pagebackref,breaklinks,colorlinks,citecolor=cvprblue]{hyperref}

\def\paperID{12552} %
\def\confName{CVPR}
\def\confYear{2024}

\title{NeRF \emph{On-the-go}: Exploiting Uncertainty for Distractor-free NeRFs in the Wild}

\author{
Weining Ren$^{1*}$ \quad 
Zihan Zhu$^{1}$${\footnotemark}$ \quad 
Boyang Sun$^{1}$ \quad 
Jiaqi Chen$^{1}$ \quad 
Marc Pollefeys$^{1,2}$ \quad 
Songyou Peng$^{1,3}$ \vspace{0.2em}\\
$^{1}$ETH Z\"urich\qquad
$^{2}$Microsoft\qquad
$^{3}$MPI for Intelligent Systems, T\"ubingen \qquad\\
\href{https://rwn17.github.io/nerf-on-the-go/}{https://rwn17.github.io/nerf-on-the-go/}
}

\newcommand{\bC}{\mathbf{C}}

\newcommand{\bff}{\mathbf{f}} %

\newcommand{\br}{\mathbf{r}}

\newcommand{\figref}[1]{Fig.~\ref{#1}}
\newcommand{\secref}[1]{Sec.~\ref{#1}}

\newcommand{\eqnref}[1]{Eq.~\eqref{#1}}
\newcommand{\tabref}[1]{Table~\ref{#1}}

\def\mc{\mathcal}

\makeatletter
\DeclareRobustCommand\onedot{\futurelet\@let@token\@onedot}
\def\@onedot{\ifx\@let@token.\else.\null\fi\xspace}
\def\eg{e.g\onedot}

\def\wrt{wrt\onedot}

\def\etal{et~al\onedot}

\makeatother

\newcommand{\boldparagraph}[1]{\vspace{0.2em}\noindent{\bf #1.}}

\renewcommand{\paragraph}[1]{\boldparagraph{#1}}

\definecolor{darkgreen}{rgb}{0,0.7,0}
\definecolor{newyellow}{rgb}{1,0.8,0.05}
\definecolor{newgreen}{rgb}{0.2,0.8,0.2}

\newcommand{\ours}{Ours\xspace}
\newcommand{\RobustNeRF}{RobustNeRF~\cite{robustnerf}\xspace}
\newcommand{\RobustNeRFStar}{RobustNeRF$^*$~\cite{robustnerf}\xspace}
\newcommand{\NeRFW}{NeRF-W~\cite{nerfw}\xspace}
\newcommand{\hanerf}{Ha-NeRF~\cite{chen2022hallucinated}\xspace}
\newcommand{\ddnerf}{D$^2$NeRF~\cite{ddnerf}\xspace}

\newcommand{\mipNeRFthreesixty}{Mip-NeRF 360~\cite{mipnerf360}\xspace}
\newcommand{\lpips}{\scalebox{0.8}{LPIPS$\downarrow$}}
\newcommand{\mssim}{\scalebox{0.8}{MS-SSIM$\uparrow$}}
\newcommand{\ssim}{\scalebox{0.8}{SSIM$\uparrow$}}
\newcommand{\psnr}{\scalebox{0.8}{PSNR$\uparrow$}}

\makeatletter
\def\adl@drawiv#1#2#3{%
        \hskip.5\tabcolsep
        \xleaders#3{#2.5\@tempdimb #1{1}#2.5\@tempdimb}%
                #2\z@ plus1fil minus1fil\relax
        \hskip.5\tabcolsep}
\newcommand{\cdashlinelr}[1]{%
  \noalign{\vskip\aboverulesep
           \global\let\@dashdrawstore\adl@draw
           \global\let\adl@draw\adl@drawiv}
  \cdashline{#1}
  \noalign{\global\let\adl@draw\@dashdrawstore
           \vskip\belowrulesep}}
\makeatother

\begin{document}

\twocolumn[{%
\renewcommand\twocolumn[1][]{#1}%
\maketitle
\vspace{-3em}
\begin{center}
    \includegraphics[width=1\textwidth]{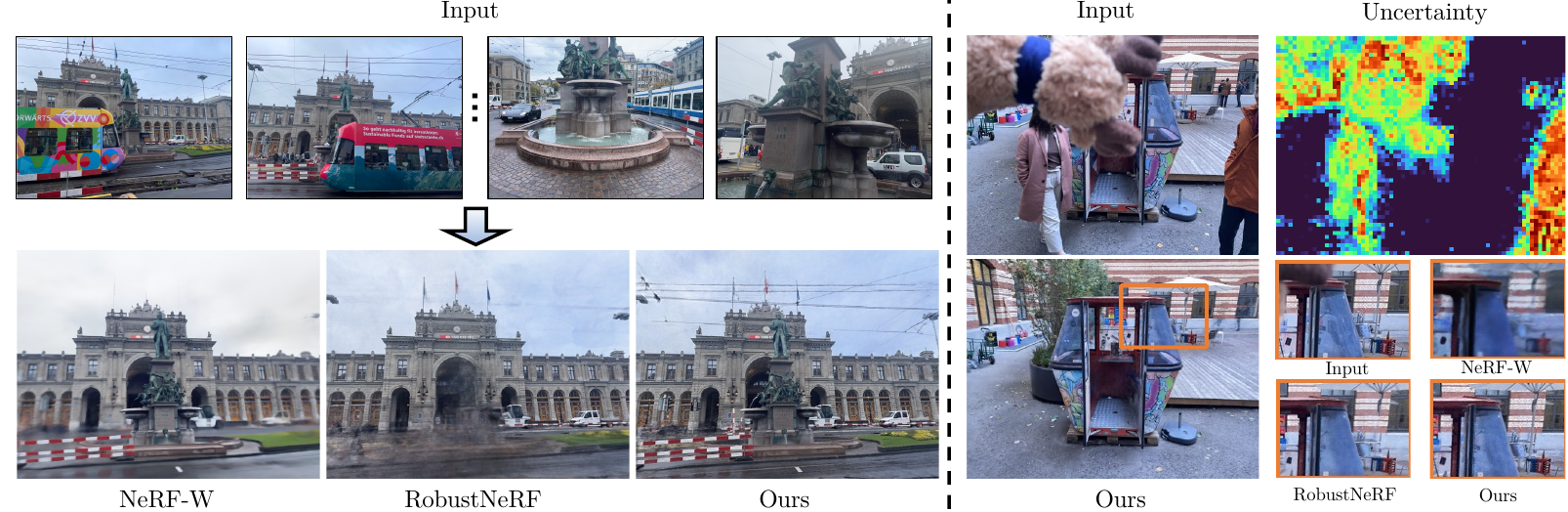}
    \vspace{-1.5em}
    \captionof{figure}{\textbf{NeRF \textit{On-the-go}.}
    Given casually captured image sequences or videos in the wild as inputs, the goal of this paper is to train a NeRF for static scenes and effectively remove all dynamic elements in the scenes (cars, trams, pedestrians, etc), i.e. distractors.
    Unlike existing methods such as \NeRFW and \RobustNeRF, which produce imperfect results,
    our method leverages the predicted uncertainty maps to effectively remove those distractors. This results in high-fidelity novel view synthesis on challenging dynamic scenes.
    }
    \label{fig:teaser}
\end{center}
}]
\footnotetext{* Equal contribution.}

\begin{abstract}
\vspace{-1.0em}
Neural Radiance Fields (NeRFs) have shown remarkable success in synthesizing photorealistic views from multi-view images of static scenes, but face challenges in dynamic, real-world environments with distractors like moving objects, shadows, and lighting changes. Existing methods manage controlled environments and low occlusion ratios but fall short in render quality, especially under high occlusion scenarios. In this paper, we introduce NeRF On-the-go, a simple yet effective approach that enables the robust synthesis of novel views in complex, in-the-wild scenes from only casually captured image sequences. Delving into uncertainty, our method not only efficiently eliminates distractors, even when they are predominant in captures, but also achieves a notably faster convergence speed. Through comprehensive experiments on various scenes, our method demonstrates a significant improvement over state-of-the-art techniques. This advancement opens new avenues for NeRF in diverse and dynamic real-world applications.
\end{abstract}
\vspace{-2.0em}
    
\section{Introduction}
\label{sec:intro}

Novel View Synthesis (NVS) tackles the challenge of rendering a scene from previously unobserved viewpoints.
Neural radiance fields (NeRFs)~\cite{nerf} have emerged as a groundbreaking paradigm for this task. 
This is because a NeRF can produce geometrically consistent and photorealistic renderings, even for complex scenarios with thin structures and semi-transparent objects.

Training a NeRF model requires a set of RGB images with given camera poses, and demands manual adjustments of camera settings, such as focal length, exposure, and white balance. More crucially, vanilla NeRFs operate under the assumption that the scene should remain completely static during the capture process, without any \emph{distractors} such as moving objects, shadows, or other dynamic elements~\cite{robustnerf}. 
Nevertheless, the real world is inherently dynamic, making this distractor-free requirement often unrealistic to meet.
Additionally, removing distractors from the captured data is non-trivial. 
The process involves per-pixel annotation for each image, a procedure that is very labor-intensive, especially for lengthy captures of large scenes.
This underscores a key limitation in the practical application of NeRFs in dynamic, real-world environments.

Recently, several works~\cite{martin2021nerf,Tancik2022CVPR,Rematas2022CVPR,ddnerf} have attempted to address the challenges. \cite{Rematas2022CVPR} and \cite{Tancik2022CVPR} use pre-trained semantic segmentation models for specific moving objects, but the model fails to segment undefined object classes. 
NeRF-W~\cite{martin2021nerf} optimizes pixel-wise uncertainty from randomly initialized embedding by volume rendering. Such a design is suboptimal since it neglects the prior information of the image and entangles the uncertainty with radiance field reconstruction. As a result, they need to introduce transient embeddings to account for distractors. The addition of a new degree of freedom complicates system tuning, leading to a Pareto-optimal scenario as discussed in~\cite{robustnerf}.
Dynamic NeRF methods like D$^2$NeRF~\cite{ddnerf} can decompose static and dynamic scenes for video input, but underperform with sparse image inputs.
More recently, RobustNeRF~\cite{robustnerf} models distractors as outliers and demonstrates impressive results in controlled and simple scenarios.
Nevertheless, its performance significantly drops in complex, in-the-wild scenes.
Interestingly, RobustNeRF also underperforms in scenarios without any distractors.
This leads to a compelling research question:
\begin{center}
\emph{Can we build a NeRF for in-the-wild scenes from casually captured images, regardless of the ratio of distractors?}
\end{center}

Toward this goal, we introduce NeRF \emph{On-the-go}, a versatile plug-and-play module designed for effective distractor removal, allowing rapid NeRF training from any casually captured images.
Our method is grounded in three key aspects.
First, we utilize DINOv2 features~\cite{oquab2023dinov2} for their robustness and spatial-temporal consistency in feature extraction, from which a small multi-layer perception (MLP) predicts per-sample pixel uncertainty. 
Second, our method leverages a structural similarity loss to improve uncertainty optimization, enhancing the distinction between foreground distractors and the static background. Third, we incorporate estimated uncertainty into NeRF's image reconstruction objective using a decoupled training strategy, which significantly enhances distractor elimination, particularly in high occlusion scenes.
Our method demonstrates robustness across a wide range of scenarios, from confined indoor scenes with small objects to complex, large-scale street view scenes, and can effectively handle varying levels of distractors.
Notably, we find that our \textit{On-the-go} module can also significantly accelerate NeRF training up to one order of magnitude, compared with RobustNeRF.
This efficiency, combined with its straightforward integration with modern NeRF frameworks, makes NeRF \textit{On-the-go} an accessible and powerful tool for enhancing NeRF training in dynamic real-world settings.

\section{Related Work}
\label{sec:related}

\paragraph{Uncertainty in Scene Reconstruction}
Uncertainty has proven to enhance the robustness and reliability of a wide range of tasks such as monocular depth prediction~\cite{hornauer2022gradient, poggi2020uncertainty}, semantic segmentation~\cite{Huang2018ECCV,Mukhoti2021ARXIV}, and simultaneous localization and mapping (SLAM)~\cite{Yang2020CVPR,Merrill2022CVPR,Costante2020TRO,Sandstrom2023ICCVW}. In general, uncertainty can be divided into two categories: epistemic and aleatoric~\cite{kendall2017uncertainties}. In the specific context of scene reconstruction, epistemic uncertainty generally arises from data limitations, such as restricted viewpoints. For instance, \cite{shen2023estimating} utilizes ensemble learning to quantify epistemic uncertainty for exploring unobserved regions in next-best-view (NBV) planning for NeRF.
Goli~\etal~\cite{goli2023bayes} establishes a volumetric uncertainty field to remove the floaters from NeRF. On the other hand, aleatoric uncertainty comes from the inherent randomness of the data, such as the noise of measurement, appearance changes, and distractors in the scene. There are works~\cite{pan2022activenerf, jin2023neu, ran2023neurar} that utilize aleatoric uncertainty as a guiding principle for active learning and NBV planning for better NeRF training. Similarly, DebSDF~\cite{Xiao2023ARXIV} improves indoor scene reconstruction through an uncertainty map to mitigate the noise from monocular prior. 

Closely related to us, NeRF-W~\cite{nerfw} was pioneering to eliminate transient objects and address variable illumination in unstructured internet photo collections, achieved by introducing transient and appearance embeddings. 
Follow-up works like Ha-NeRF~\cite{chen2022hallucinated} hallucinates NeRFs from unconstrained tourism images, while Neural Scene Chronology~\cite{lin2023neural} reconstructs temporal-varying chronology from time-stamped Internet photos.
Building upon previous formulation for aleatoric uncertainty, 
we innovate by integrating DINOv2 features into uncertainty prediction, which enhances the quality of predicted uncertainty.
In a recent work, Kim~\etal~\cite{kim2023up} also presents a similar DINO-based uncertainty prediction approach, but directly adapts for NeRF-W~\cite{nerfw} to a pose-free condition.
In contrast, we focus on refining NeRF training to effectively handle distractors from casually-captured image sequences.

\paragraph{SLAM and SfM in Dynamic Scenes}
Handling dynamic scenes has been studied for years in the literature of SLAM and SfM.
Classical methods exclude pixels associated with dynamic objects with robust kernel function~\cite{mur2017orb, engel2017direct} or RANSAC~\cite{schoenberger2016sfm, schoenberger2016mvs}. However, such hand-craft features are effective in scenarios with a low occlusion ratio but struggle at in-the-wild scenes. To address this, recent advances have integrated additional information. This includes external segmentation or detection modules for pre-defined classes~\cite{yu2018ds, yang2019cubeslam, ye2023pvo, zhang2020vdo}, utilization of optical or scene flow~\cite{barsan2018robust, ye2022deflowslam, shen2023dytanvo, esparza2022stdyn, teed2021droid, zhao2022particlesfm}, and geometry-based approaches using clustering and epipolar line distance~\cite{yu2018ds, bescos2018dynaslam, huang2019clusterslam}.

\paragraph{NeRF in Dynamic Scenes}
Recent NeRF methods focus on reconstructing both static and dynamic components from a video sequence~\cite{li2022neural,hypernerf,ddnerf, xian2021space,gao2021dynamic, nsff,wang2021neural, du2021neural}
enabling novel view synthesis at arbitrary timestamps. Although primarily designed for video inputs, these methods often underperform with photo collection sequences~\cite{robustnerf}. Additionally, separating static and dynamic components can be time-consuming and requires extensive hyperparameter tuning. A notable example in this realm is EmerNeRF~\cite{yang2023emernerf}, which also employs the DINOv2~\cite{oquab2023dinov2} features. However, they use them for enhanced scene decomposition, while we use them as a strong prior knowledge for distractor removal.

RobustNeRF, to our knowledge the only method that also targets static scene reconstruction from dynamic scenes, uses Iteratively Reweighted Least Squares for outlier verification. Compared with it, our method can deal with more complex scenes with various levels of occlusions.

\section{Method}
\label{sec:method}
\begin{figure}[t!]
    \centering
    \footnotesize
    \setlength{\tabcolsep}{1.5pt}
    \includegraphics[width=\columnwidth]{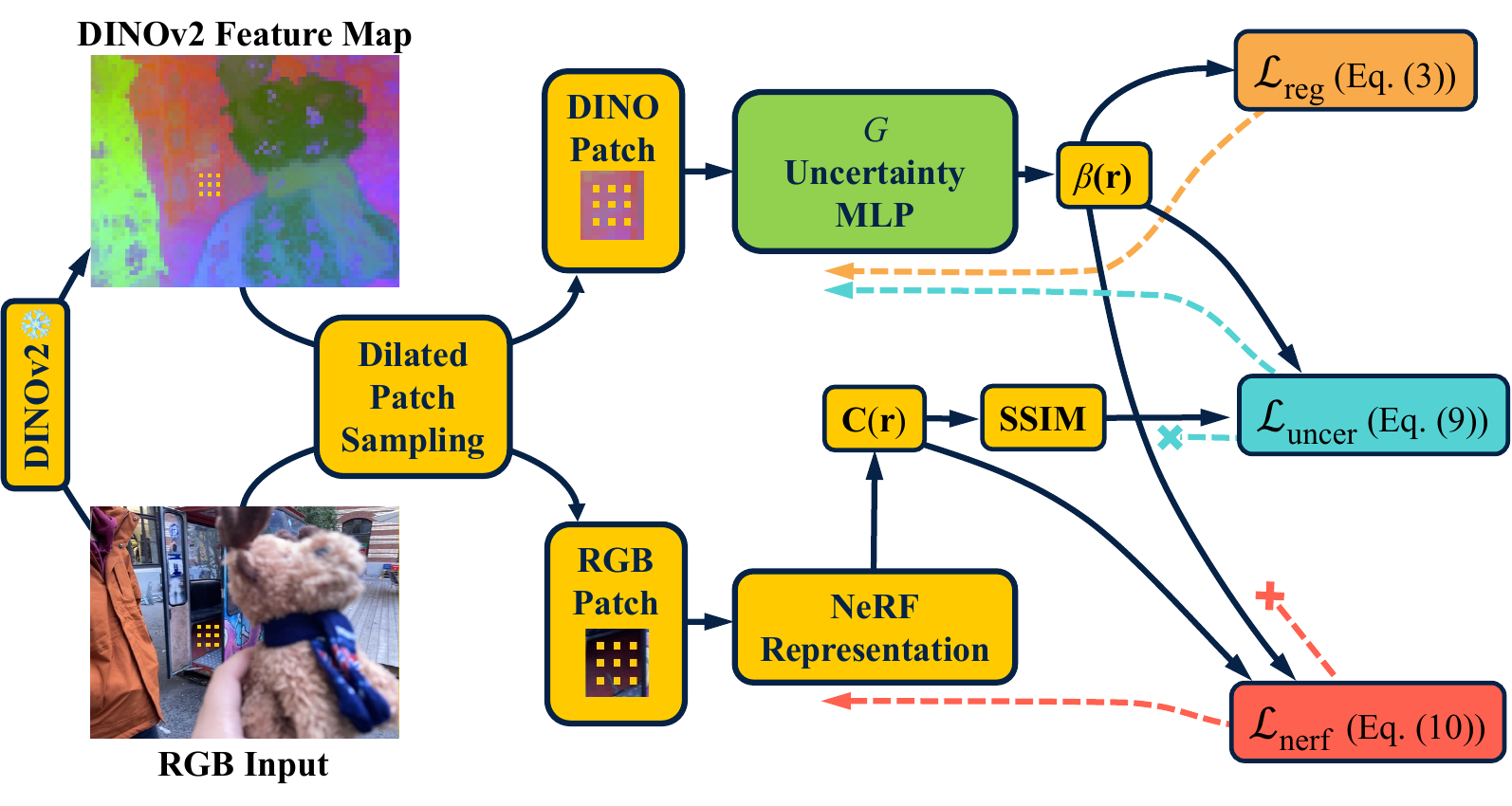}
    \caption{\textbf{Pipeline.}  A pre-trained DINOv2 network extracts feature maps from posed images, followed by a dilated patch sampler that selects rays. The uncertainty MLP $G$ then takes the DINOv2 features of these rays as inputs to generate the uncertainties \(\beta(\br)\). 
    Three losses (on the right) are used to optimize $G$ and the NeRF model.
    Note that the training process is facilitated by detaching the gradient flows as indicated by the colored dashed lines.}
    \label{fig:pipeline}
\end{figure}

We start by showing how to utilize per-pixel DINO features for uncertainty prediction (\secref{method:dino}). 
Subsequently, we show a novel approach for learning uncertainty to remove distractors in NeRF (\secref{method:uncertainty}). 
We further introduce our decoupled optimization scheme for uncertainty prediction and NeRF (\secref{method:optimization}).
Finally, we illustrate why sampling method is important in distractor-free NeRF training (\secref{method:dilated_patch}).
An overview of our pipeline is depicted in \figref{fig:pipeline}.

\newcommand{\uni}{\beta (\br)}

\subsection{Uncertainty Prediction with DINOv2 Features}
\label{method:dino}

Our primary objective is to effectively identify and eliminate recurring distractors--those that appear across multiple images.
To achieve this, we take advantage of DINOv2~\cite{oquab2023dinov2} features, which have shown to be able to maintain spatial-temporal consistency across views.

We begin with extracting DINOv2 features for each input RGB image.
Next, these features serve as inputs to a small MLP to predict the uncertainty value for each pixel. 
To further enforce the consistency of our uncertainty prediction, we incorporate a regularization term.

\paragraph{Image Feature Extraction} 
For RGB images with a resolution of \( H \times W \), we derive per-pixel features through a pre-trained DINOv2 feature extractor \(\mathcal{E}\): 
 \begin{equation}
   \mc{F}_i = \mathcal{E}(\mc{I}_i),\ \mathcal{E} \in \mathbb{R}^{H \times W \times 3} \to \mathbb{R}^{H \times W \times C}
 \end{equation}
where \( i \) spans all training images, and \( C \) denotes feature dimension. 
This module also upsamples the feature maps to the original resolution by nearest-neighbor sampling. 

\paragraph{Uncertainty Prediction} 
Once we obtain the 2D DINOv2 feature maps, we proceed to determine the uncertainty of each sampled ray $\br$. We first query its corresponding feature \( \mathbf{f} = \mc{F}_i(\br)\), and then input it to a shallow MLP to estimate the uncertainty for this ray $\beta(\mathbf{r}) = G(\mathbf{f})$, where \(G\) is the uncertainty MLP.
In the subsequent sections, we will demonstrate how this predicted uncertainty $\beta(\mathbf{r})$ is integrated into the optimization process as a weighting function, which plays a crucial role in refining the NeRF model, particularly in handling and mitigating the impact of distractors in the scene.

\paragraph{Uncertainty Regularization}
To enforce spatial-temporal consistency in uncertainty predictions,  we introduce a regularization term based on the cosine similarity of feature vectors within a minibatch.
Specifically, for each sampled ray \(\br\), we define a neighbor set $\mathcal{N}(\mathbf{r})$ consisting of rays in the same batch whose associated feature vectors exhibit high similarity to the feature $\bff$ of \(\br\). 
This neighbor set is formed by selecting rays that meet a specified cosine similarity threshold $\eta$:
\begin{equation*}
\mathcal{N}(\mathbf{r}) = \{\mathbf{r}^{\prime} | \cos(\mathbf{f}, \mathbf{f}^{\prime}) > \eta \}
\end{equation*}
where $\mathbf{f}^{\prime}$ is the associated feature of $\mathbf{r}^{\prime}$.
The refined uncertainty for a ray $\br$ is computed as the average of $\mathcal{N}(\mathbf{r})$:
\begin{equation}
    \Bar{\beta}(\br) = \frac{1}{|\mathcal{N}(\br)|}\sum_{r\prime \in \mathcal{N}(\br)} \beta(\br^{\prime}) 
\end{equation}
To reinforce consistency, we introduce a regularization term that penalizes the variance of uncertainty within $\mathcal{N}(\mathbf{r})$:
\begin{equation}
\mathcal{L}_\text{reg}(\br) =\frac{1}{|\mathcal{N}(\br)|} \sum_{r\prime \in \mathcal{N}(\br)}(\Bar{\beta}(\br) - \beta(\br^{\prime}))^2 .
\label{eq:uncer_reg}
\end{equation}
This regularization aims to smooth out abrupt changes in uncertainty predictions across similar features from rays across images, thereby enhancing the overall robustness and consistency of the uncertainty estimation process.

\subsection{Uncertainty for Distractor Removal in NeRF}
\label{method:uncertainty}
We hypothesize that pixels correlating with dynamic elements (distractors) should have high uncertainty, whereas static regions should have low uncertainty.
This premise allows us to effectively integrate predicted uncertainty into NeRF training objectives, aiming to progressively filter out distractors for enhanced novel view synthesis.

We will analyze the potential issue of the classical way of incorporating uncertainty into the loss function for NeRF. Finally, we introduce a simple yet effective modification, to incorporate uncertainty, for robust distractor removal. 

\paragraph{Uncertainty Convergence Analysis\label{method:Converge}}
Uncertainty prediction has been widely used in different fields, including NeRF-based novel view synthesis.
For example, in the seminal work NeRF in the Wild~\cite{nerfw}, their loss is written as \footnote{We omit their regularization term for transient density.}: 
\begin{equation}
    \mathcal{L}(\mathbf{r})=\frac{\| \bf{C}(\br) - \hat{\bf{C}}(\br) \|^2}{2 \beta^2(\br)}+\lambda_1 \log \beta(\br)
    \label{eq:l2_loss}
\end{equation}
Here, $\bC(\br)$ and $\hat{\bC}(\br)$ represent the input and rendered RGB values.
The uncertainty \(\beta(\br)\) is treated as a weight function. The regularization term is crucial for balancing the first term and preventing the trivial solution where \(\beta(\br) = \infty \). 

Here we present a simple analysis to understand how the uncertainty changes \wrt the loss function, we first take the partial derivative \wrt \(\beta(\br)\):
\begin{equation}
    \frac{d \mathcal{L}(\br)}{d \beta (\br)}=- \frac{\| \bf{C}(\br) - \hat{\bf{C}}(\br) \|^2}{\uni^3}+\lambda_1 \frac{1}{\beta (\br)}
\end{equation}
Setting this derivative to 0, we derive the closed-form solution for the optimal uncertainty:
\begin{equation}\label{eqa:uncer}
   \frac{d \mathcal{L}(\br)}{d \beta (\br)} = 0 \Rightarrow \uni=\sqrt{\frac{1}{\lambda_1}}    \| \bf{C}(\br) - \hat{\bf{C}}(\br) \|
\end{equation}
This reveals an important relationship between uncertainty prediction and the error between the rendered and input colors.
Specifically, the optimal uncertainty is directly proportional to this error term.

\begin{figure}[t!]
    \centering
    \setlength{\tabcolsep}{1.5pt}
    \begin{tabular}{cccc}
      \begin{subfigure}{0.28\columnwidth}
        \includegraphics[width=\linewidth]{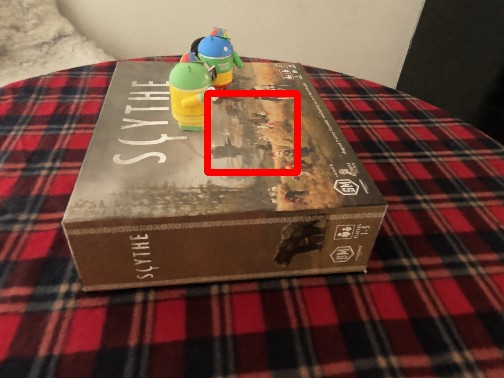}
        \caption{Rendering}
        \label{fig:sub_ssim1}
      \end{subfigure} &
      \begin{subfigure}{0.28\columnwidth}
        \includegraphics[width=\linewidth]{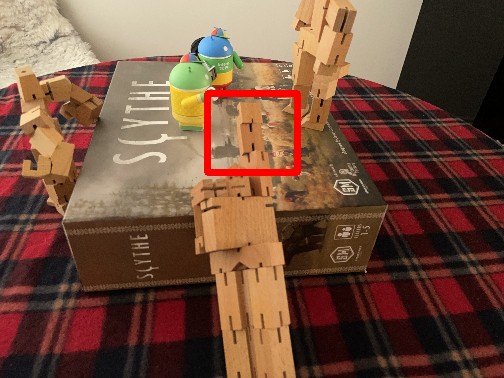}
        \caption{Input}
        \label{fig:sub_ssim2}
      \end{subfigure} &
      \begin{subfigure}{0.28\columnwidth}
        \includegraphics[width=\linewidth]{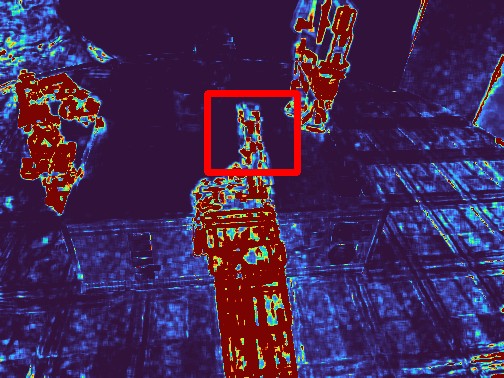}
        \caption{SSIM Error}
        \label{fig:sub_ssim3}
      \end{subfigure} &
      \multirow{2}{*}[2cm]{\begin{subfigure}{0.06\columnwidth}
        \includegraphics[width=0.8\linewidth, height=8.7\linewidth]{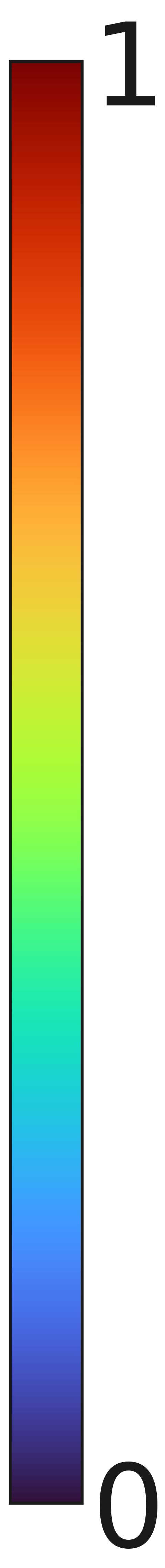}
        \label{fig:large}
    \end{subfigure}} \\
      \begin{subfigure}{0.28\columnwidth}
        \includegraphics[width=\linewidth]{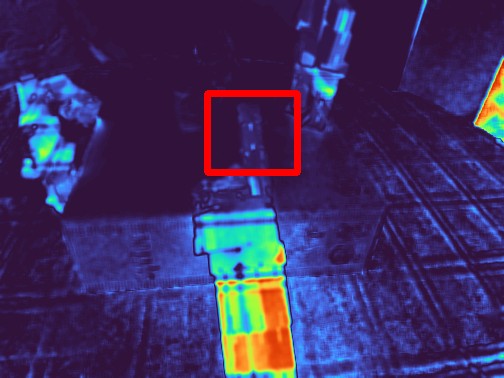}
        \caption{Luminance Error}
        \label{fig:sub_ssim4}
      \end{subfigure} &
      \begin{subfigure}{0.28\columnwidth}
        \includegraphics[width=\linewidth]{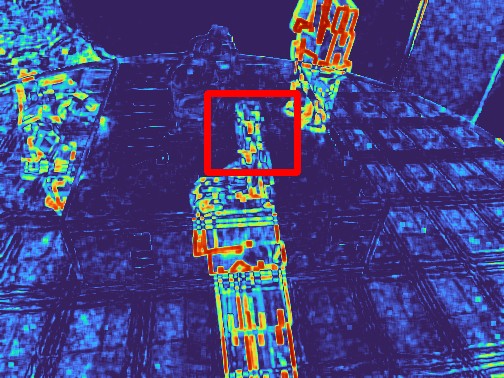}
        \caption{Contrast Error}
        \label{fig:sub_ssim5}
      \end{subfigure} &
      \begin{subfigure}{0.28\columnwidth}
        \includegraphics[width=\linewidth]{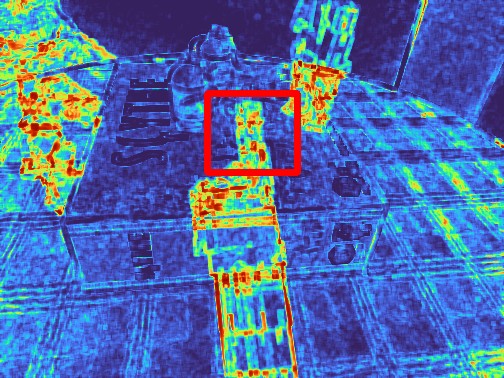}
        \caption{Structure Error}
        \label{fig:sub_ssim6}
      \end{subfigure} & \\
    \end{tabular}
    \vspace{-0.5em}
    \caption{\textbf{SSIM Can Effectively Distinguish Distractors.} In this scene from~\cite{robustnerf}, the 3 wooden robots are the dynamic elements. 
    SSIM pinpoints distractors by leveraging discrepancies in three measurements including luminance, contrast, and structure. Conversely, relying solely on the $\ell_2$ error between RGB values (luminance error) proves challenging, especially when the distractors and background have similar colors. The color bar on the right side indicates the correspondence for error interpretation.
    }
    \label{fig:ssim}
  \end{figure}

However, a challenge arises when employing the $\ell_2$ loss as shown in~\eqnref{eq:l2_loss}, particularly when the color of distractors and background is close (as illustrated in \figref{fig:ssim} (d)). 
In such cases, the predicted uncertainty in those regions will also be low according to~\eqnref{eqa:uncer}.
This impedes the effectiveness of uncertainty-based distractor removal, and leads to cloud artifacts in the rendered images. 

Recognizing the limitation inherent in the $\ell_2$ RGB loss, we propose a new loss for better uncertainty learning, so that the predicted uncertainty can discriminate between distractors and static background more effectively.

\paragraph{SSIM-Based Loss for Enhanced Uncertainty Learning}
The structural similarity index (SSIM) is comprised of three measurements: luminance, contrast, and structure similarities. These components capture local structural and contractual differences, which is crucial for distinguishing between scene elements.
This is verified in \figref{fig:ssim}, where SSIM is effective in detecting distractors by incorporating these three components together.
An SSIM loss can be formulated as:
\begin{equation}
\begin{aligned}
\mathcal{L}_\text{SSIM} &= 1 - \operatorname{SSIM}(P, \hat{P})\\ &=
1 - L(P, \hat{P}) \cdot C(P, \hat{P}) \cdot S(P, \hat{P})
\end{aligned}\label{eq:ssim}
\end{equation}
where $P$ and $\hat{P}$ are patches sampled from the input and rendered images \(\bf{C} (\br)\) and \(\hat{\bf{C}}(\br)\), respectively.
$L, C, S$ refer to the luminance, contrast, and structure similarities between $P$ and $\hat{P}$. 
We further modify~\eqnref{eq:ssim} as:
\begin{equation}
\mathcal{L}_\text{SSIM} = (1-L(P, \hat{P}))\cdot(1-C(P, \hat{P}))\cdot(1-S(P, \hat{P}))
\label{eq:ssim_new}
\end{equation}

Compared to~\eqnref{eq:ssim}, our reformulation in~\eqnref{eq:ssim_new} places greater emphasis on the differences between dynamic and static elements. 
Consequently, this enhances the disparity in uncertainty, facilitating more effective optimization of uncertainty.
The mathematical proof and comparisons between~\eqnref{eq:ssim}  and~\eqnref{eq:ssim_new} are included in the supplements.

Building on this updated SSIM formulation, we introduce a new loss tailored for uncertainty learning:
\begin{equation}
\mathcal{L}_\text{uncer}(\br) = \frac{\mathcal{L}_\text{SSIM}}{2\uni^2} + \lambda_1 \log \uni \\
\label{eq:uncer}
\end{equation}
This loss is a simple modification of~\eqnref{eq:l2_loss}, adapted for better uncertainty learning. $\mathcal{L}_\text{uncer}$ is specifically applied to train the uncertainty estimation MLP $G$.
This is crucial as it allows us to decouple the training of the NeRF model from uncertainty prediction.
Such decoupling ensures that the learned uncertainty is robust to various types of distractors.
Please refer to \tabref{tab:ablation} for an ablation for $\mathcal{L}_\text{uncer}$.

Note that a recent work S3IM~\cite{xie2023s3im} also uses SSIM for NeRF training, but their loss is tailored for static scenes, whereas ours is designed for better uncertainty learning. Also, S3IM employs stochastic sampling to identify non-local structural similarities, while we use dilated sampling to focus on local structures for distractor removal.

\subsection{Optimization\label{method:optimization}}
As mentioned above, it is crucial to separately optimize the uncertainty prediction module and NeRF model.
For optmization of the uncertainty prediction MLP, we employ $\mathcal{L}_\text{uncer}$ in~\eqnref{eq:uncer} and $\mathcal{L}_\text{reg}$ in~\eqnref{eq:uncer_reg}.
In parallel, we train the NeRF model with the following:
\begin{equation}
\mathcal{L}_\text{nerf}(\br) =  \frac{\| \bf{C}(\br) - \hat{\bf{C}}(\br) \|^2}{2 \beta^2(\br)}
\end{equation}
This loss, essentially~\eqnref{eq:l2_loss} without the regularization term, is used because $\mathcal{L}_\text{uncer}$ already prevents trivial solutions for uncertainty ($\beta(\br) = \infty$).
The parallel training process is facilitated by detaching the gradient flow from $\mathcal{L}_\text{uncer}$ to NeRF representation, and $\mathcal{L}_\text{nerf}$ to the uncertainty MLP $G$ as illustrated in \figref{fig:pipeline}.
Note that we also follow RobustNeRF~\cite{robustnerf} and include the interval loss and distortion loss from Mip-NeRF 360~\cite{mipnerf360} for NeRF training, which we omit here for simplicity.
Our overall objectives integrate all losses together, denoted as:
\begin{equation}
 \lambda_{2}\mathcal{L}_\text{nerf}(\br) + \lambda_{3}\mathcal{L}_\text{uncer}(\br) + \lambda_{4}\mathcal{L}_\text{reg}(\br)
\end{equation}
where each term is weighted by a corresponding $\lambda$.

\subsection{Dilated Patch Sampling\label{method:dilated_patch}}
\begin{figure}[t!]
  \centering
  \begin{subfigure}{0.30\columnwidth}
    \includegraphics[width=\linewidth]{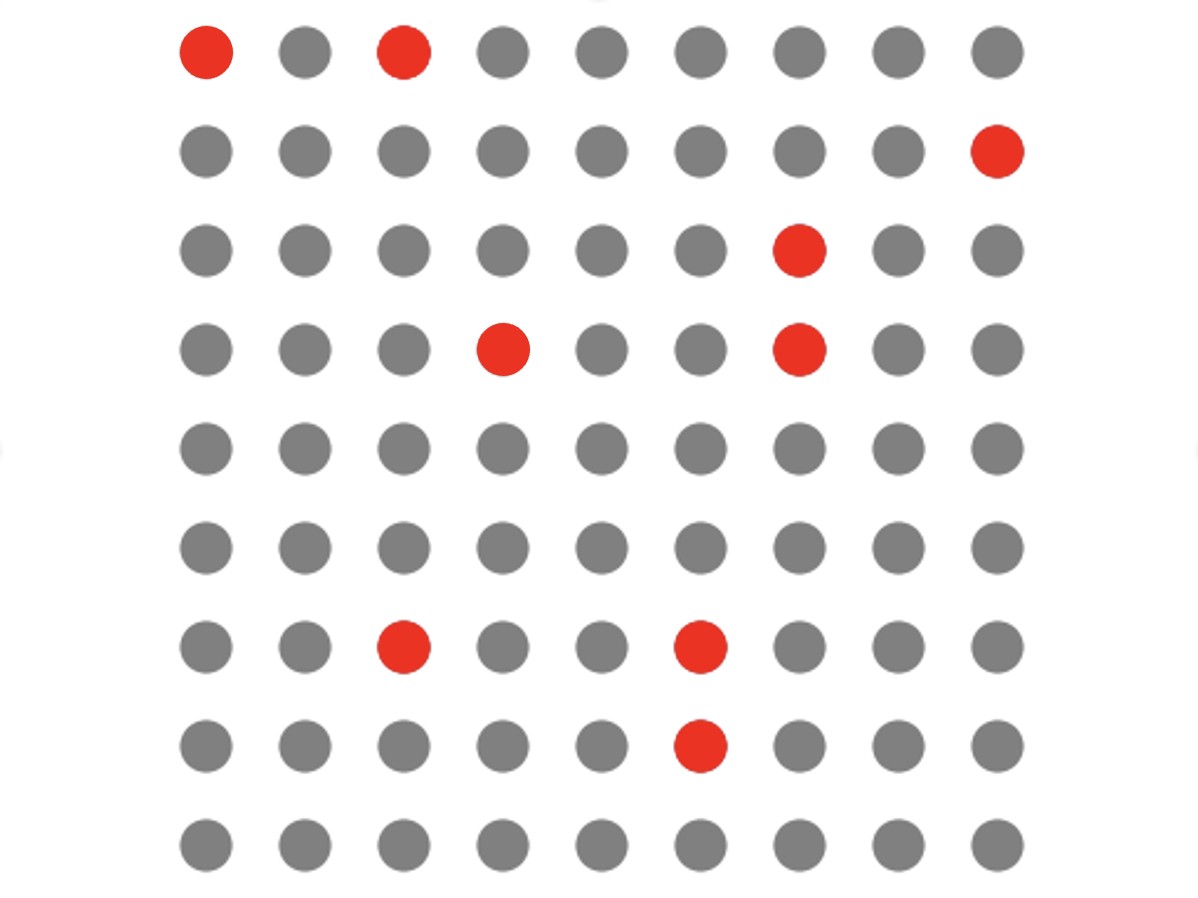}
    \caption*{(a) Random~\cite{nerfw}}
    \label{fig:sub_1}
  \end{subfigure}\hspace{1mm}
  \begin{subfigure}{0.30\columnwidth}
    \includegraphics[width=\linewidth]{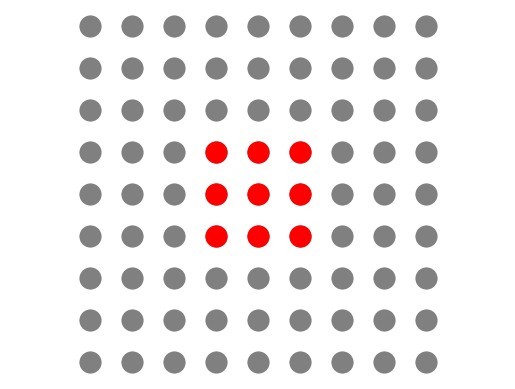}
    \caption*{(b) Patch~\cite{robustnerf}}
    \label{fig:sub2}
  \end{subfigure}\hspace{1mm}
  \begin{subfigure}{0.30\columnwidth}
    \includegraphics[width=\linewidth]{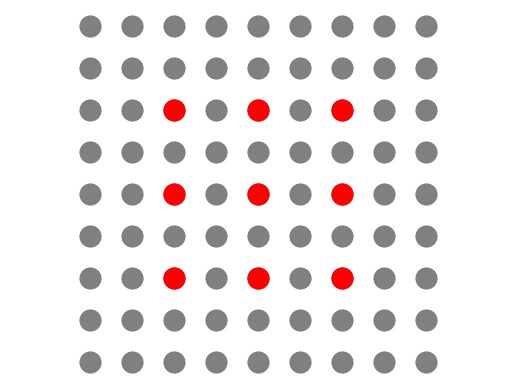}
    \caption*{(c) \textbf{Dilated Patch}}
    \label{fig:sub3}
  \end{subfigure}
  \caption{\textbf{Comparison of Different Ray Sampling Strategies.} 
  In contrast to random sampling and patch sampling, dilated patch sampling can improve training efficiency and uncertainty learning.
  }
  \label{fig:dilated_patch}
  \vspace{-0.5em}
\end{figure}

In this section, we delve into the ray sampling strategy, a key factor in the efficacy of NeRF training, particularly in achieving distractor-free results.

RobustNeRF has demonstrated the efficacy of patch-based ray sampling (\figref{fig:dilated_patch} (b)) over random sampling (\figref{fig:dilated_patch} (a)).
However, this approach has its limitations, primarily due to the small size of the sampled patches (e.g. $16\times16$). 
Especially when the batch size is small due to the constraint of GPU memory, this small context can restrict the network's learning capacity to remove distractors, impacting optimization stability and convergence speed.

To tackle the issue, we utilize dilated patch sampling~\cite{schwarz2020graf, xu2022sinnerf, mihajlovic2022keypointnerf, jain2021putting, xu2023desrf, wang2023neural}, depicted in~\figref{fig:dilated_patch} (c).
This strategy involves sampling rays from a dilated patch.
By enlarging the patch size, we can significantly increase the amount of contextual information available in each training iteration.

Our empirical findings in~\tabref{tab:ablation_dilation} show that dilated patch sampling not only accelerates the training process, but also yields superior performance in distractor removal.

\section{Experiments}
\label{sec:exp}

\boldparagraph{RobustNeRF Dataset}
There are four sequences with toys-on-the-table settings. However, note that we are unable to include the \textit{Crab} scene since it is not released. 
Meanwhile, we put comparisons on~\textit{Baby Yoda} scene in supplements, since each image in this sequence contains a distinct set of distractors, which is different from our setting.

\boldparagraph{\textit{On-the-go} Dataset}
To rigorously evaluate our approach in real-world indoor and outdoor settings, we captured a dataset that contains 12 casually captured sequences, including 10 outdoor and 2 indoor scenes, with varying ratios of distractors (from 5\% to over 30 \%).
For quantitative evaluation, we select 6 sequences representing different occlusion rates, as shown in~\figref{fig:self_sample}. More details and results for this dataset are available in supplements.

\begin{figure}
  \centering
  \footnotesize
  \setlength{\tabcolsep}{0.5pt}
  \newcommand{\sz}{0.25}
  \newcommand{\sza}{0.1729} %
  \resizebox{\linewidth}{!}{
  \begin{tabular}{cc|cc}
    \includegraphics[width=\sz\linewidth]{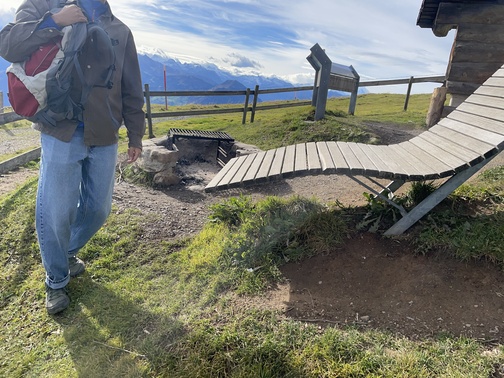} &
    \includegraphics[width=\sz\linewidth]{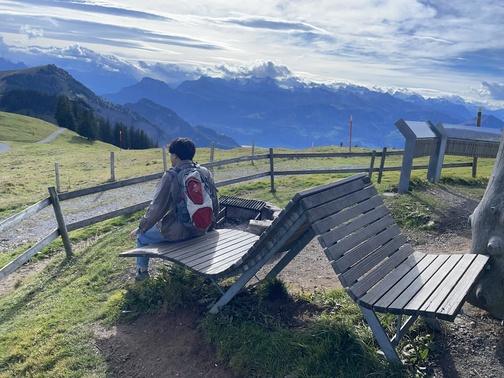} &
    \includegraphics[width=\sz\linewidth]{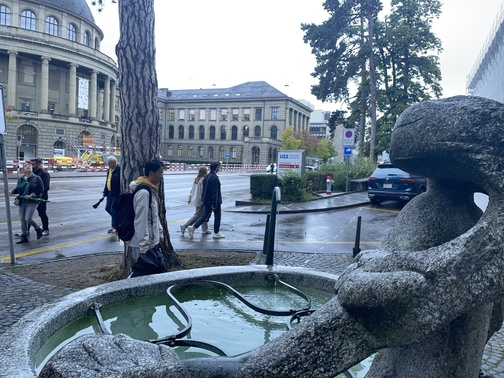} &
    \includegraphics[width=\sz\linewidth]{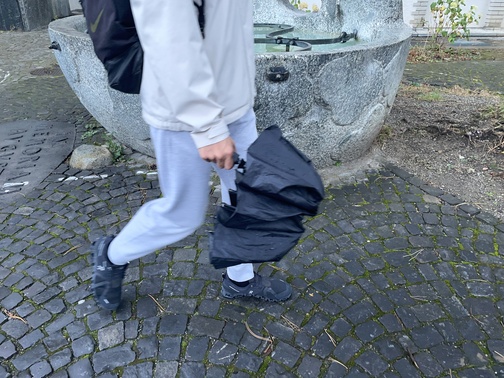} \\
    
    \multicolumn{2}{c|}{ {Mountain}} &
    \multicolumn{2}{c}{ {Fountain}} \\
    \includegraphics[width=\sz\linewidth]{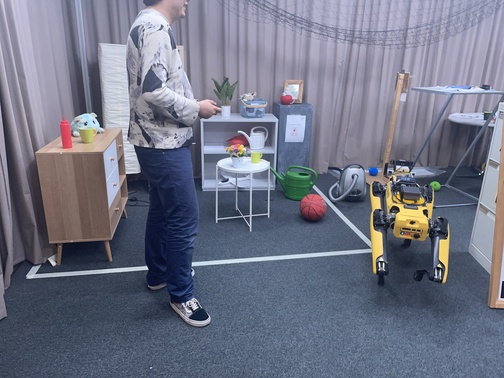} &
    \includegraphics[width=\sz\linewidth]{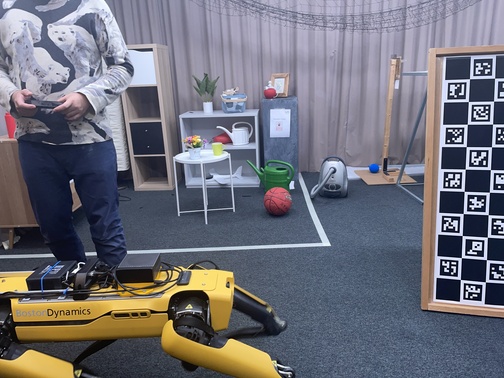} &
    \includegraphics[width=\sz\linewidth]{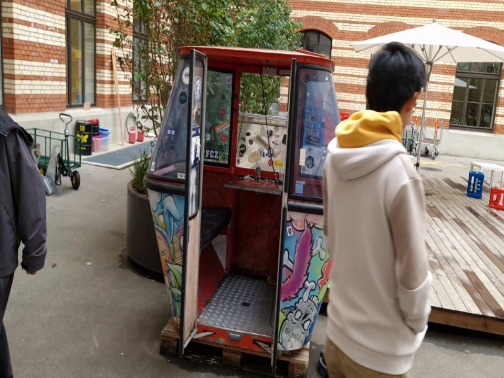} &
    \includegraphics[width=\sz\linewidth]{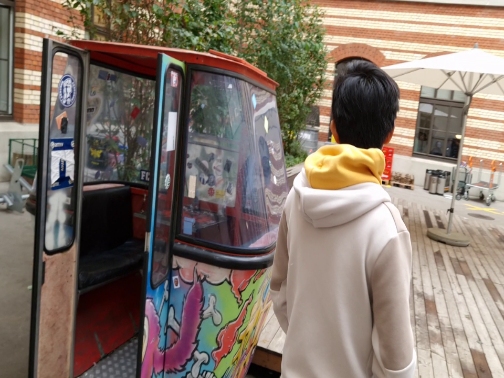} \\
    \multicolumn{2}{c|}{Corner} &
    \multicolumn{2}{c}{Patio} \\
    \includegraphics[width=\sz\linewidth]{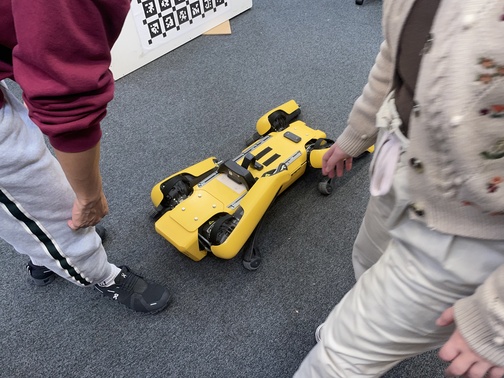} &
    \includegraphics[width=\sz\linewidth]{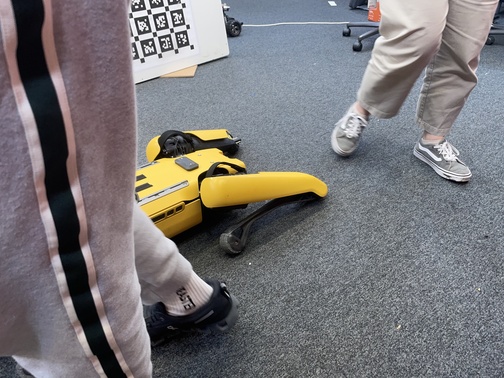} &
    \includegraphics[width=\sz\linewidth]{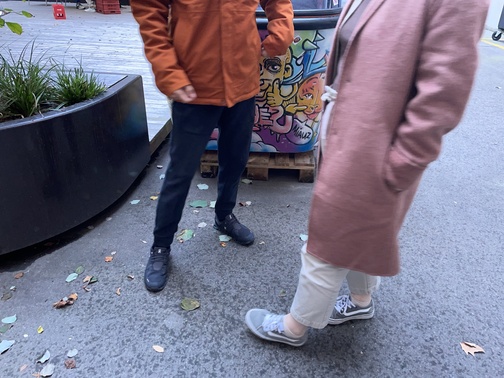} &
    \includegraphics[width=\sz\linewidth]{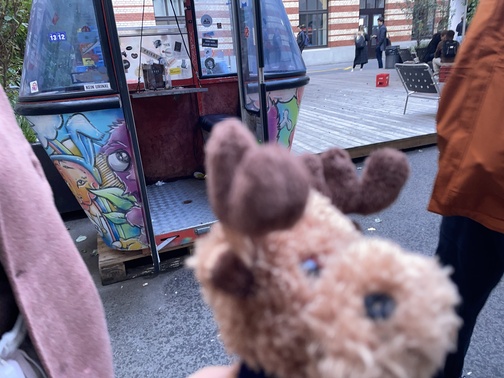} \\
    \multicolumn{2}{c|}{Spot} &
    \multicolumn{2}{c}{Patio-High} \\

\end{tabular} }
\caption{\textbf{\textit{On-the-go} Dataset.} Sample training images showing the distractors in several scenes of our self-captured dataset.
}
\label{fig:self_sample}
\end{figure}

\boldparagraph{Baselines}
We compare our method with \mipNeRFthreesixty, \ddnerf, \NeRFW\footnote{\url{https://github.com/kwea123/nerf_pl/tree/nerfw}}, Ha-NeRF~\cite{chen2022hallucinated}\footnote{\url{https://github.com/rover-xingyu/Ha-NeRF}}, \RobustNeRF\footnote{\url{https://github.com/google-research/multinerf}}, and Mip-NeRF 360 + SAM, a baseline that we design to exclude dynamic objects in images with SAM~\cite{kirillov2023segment}, and train a NeRF on static parts. Refer to supplements for more details.

\boldparagraph{Metrics}
We adopt the widely used PSNR, SSIM~\cite{ssim} and LPIPS~\cite{lpips} for the evaluation of novel view synthesis.

\subsection{Evaluation}

\begin{figure*}[t!]
  \centering
  \footnotesize
  \setlength{\tabcolsep}{0.5pt}
  \newcommand{\sz}{0.136}
  \newcommand{\sza}{0.18} %
  \begin{tabular}{ccccccc|c}
  \multirow{2}{*}{\rotatebox{90}{Mountain}} &
    \includegraphics[width=\sz\linewidth]{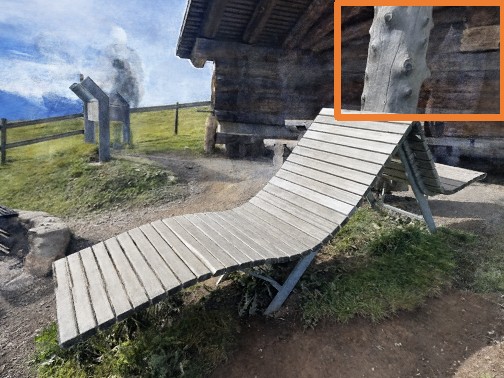} &
    \includegraphics[width=\sz\linewidth]{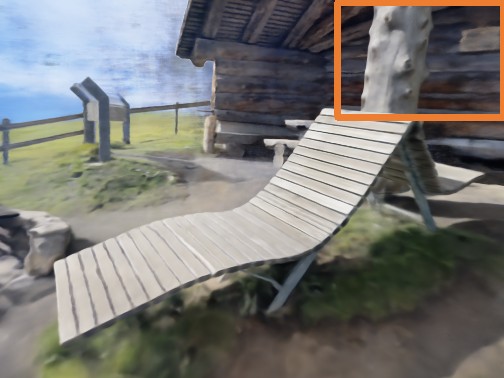} &
    \includegraphics[width=\sz\linewidth]{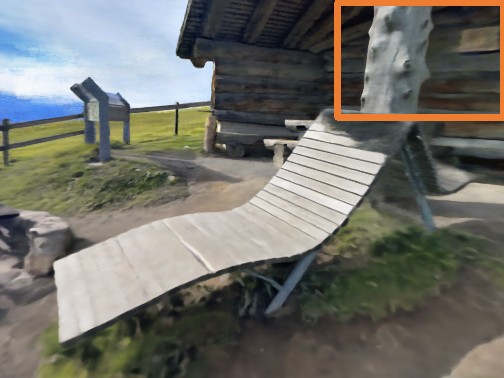} &
    \includegraphics[width=\sz\linewidth]{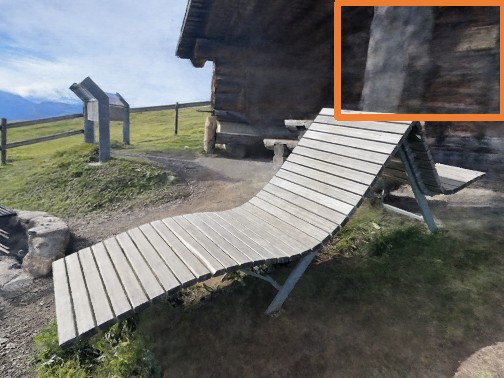} &
    \includegraphics[width=\sz\linewidth]{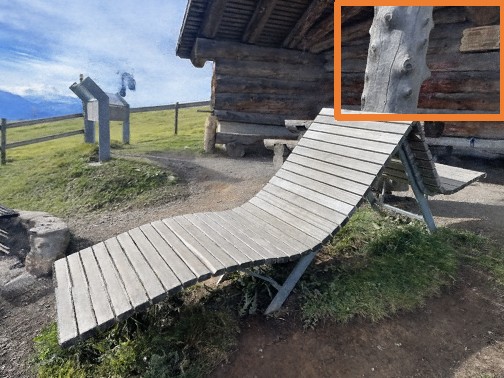} &
    \includegraphics[width=\sz\linewidth]{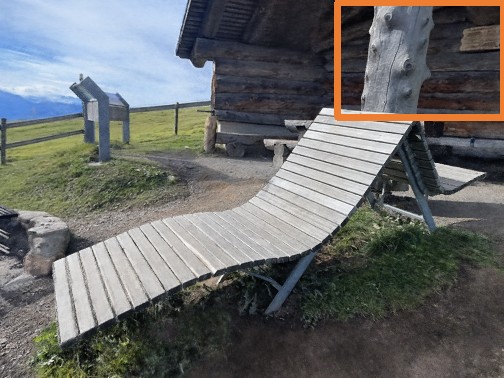}  &
    \includegraphics[width=\sz\linewidth]{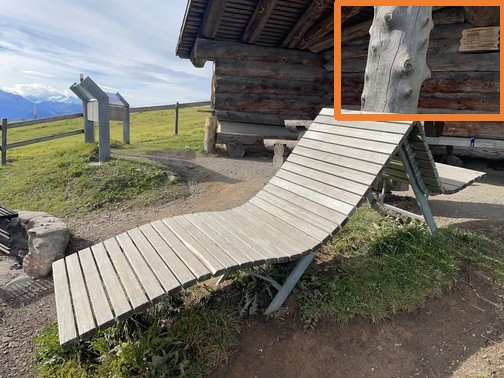} 
     \\
    & \includegraphics[width=\sz\linewidth]{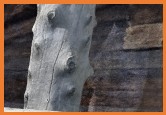} &
    \includegraphics[width=\sz\linewidth]{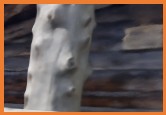} &
    \includegraphics[width=\sz\linewidth]{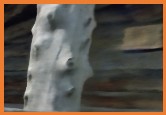} &
    \includegraphics[width=\sz\linewidth]{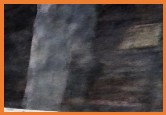} &
    \includegraphics[width=\sz\linewidth]{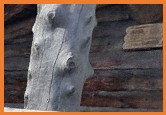} &
    \includegraphics[width=\sz\linewidth]{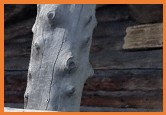}  &
    \includegraphics[width=\sz\linewidth]{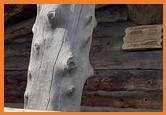} 
    
    \\

    \multirow{2}{*}{\rotatebox{90}{Fountain}} &
    \includegraphics[width=\sz\linewidth]{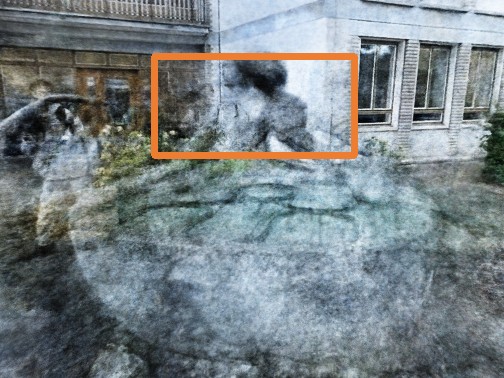} &
    \includegraphics[width=\sz\linewidth]{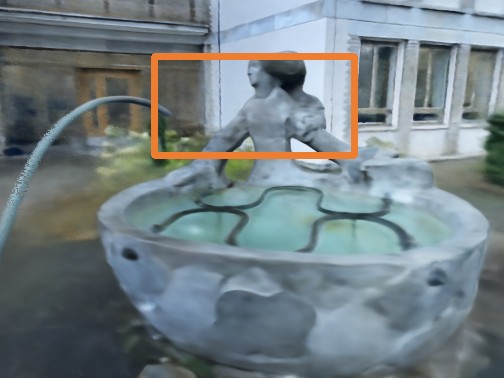} &
    \includegraphics[width=\sz\linewidth]{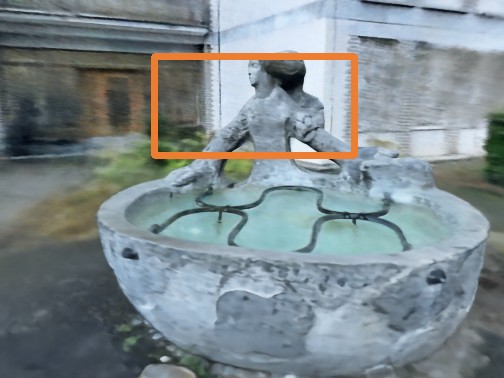} &
    \includegraphics[width=\sz\linewidth]{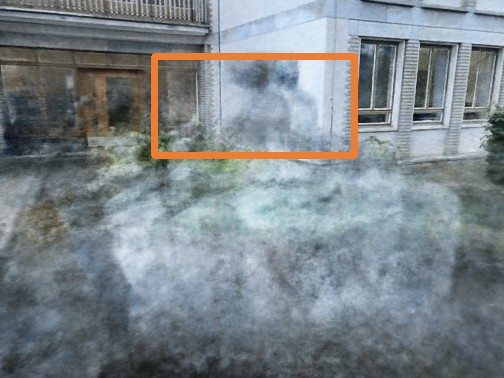} &
    \includegraphics[width=\sz\linewidth]{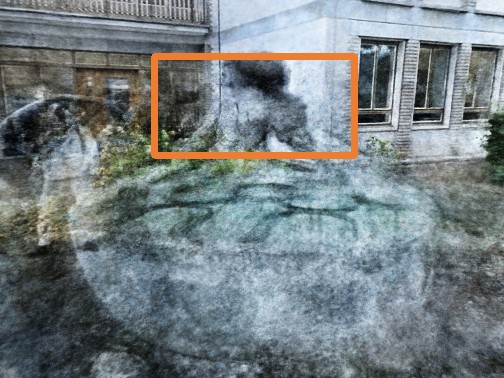} &
    \includegraphics[width=\sz\linewidth]{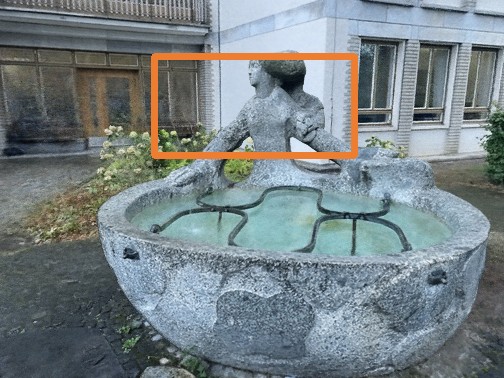}  &
    \includegraphics[width=\sz\linewidth]{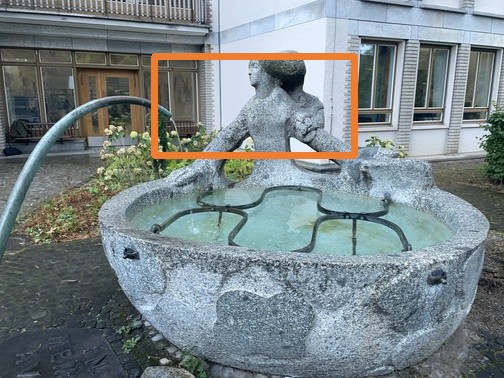} 
     \\
    & \includegraphics[width=\sz\linewidth]{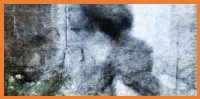} &
    \includegraphics[width=\sz\linewidth]{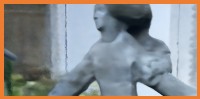} &
    \includegraphics[width=\sz\linewidth]{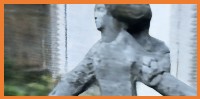} &
    \includegraphics[width=\sz\linewidth]{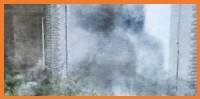} &
    \includegraphics[width=\sz\linewidth]{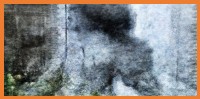} &
    \includegraphics[width=\sz\linewidth]{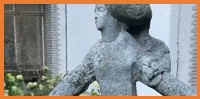}  &
    \includegraphics[width=\sz\linewidth]{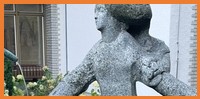} 
    \\
    \multirow{2}{*}{\rotatebox{90}{Corner}} &
    \includegraphics[width=\sz\linewidth]{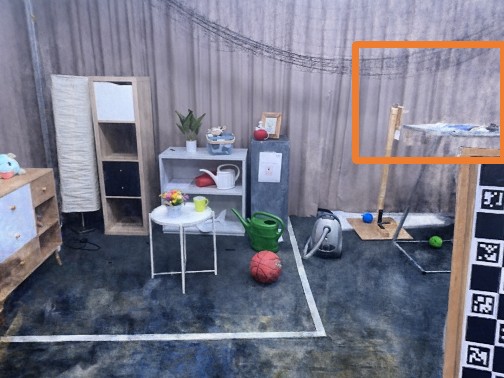} &
    \includegraphics[width=\sz\linewidth]{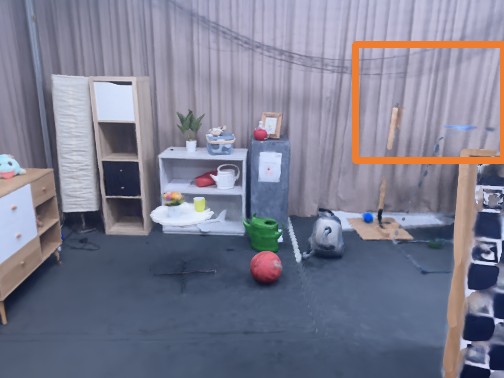} &
    \includegraphics[width=\sz\linewidth]{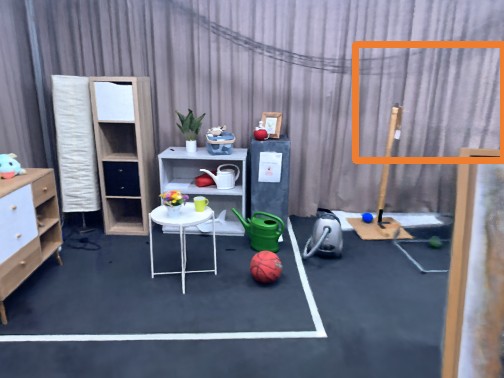} &
    \includegraphics[width=\sz\linewidth]{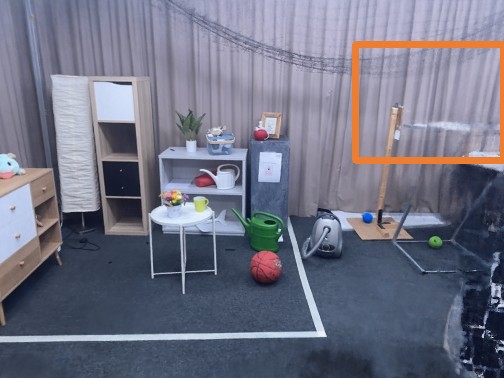} &
    \includegraphics[width=\sz\linewidth]{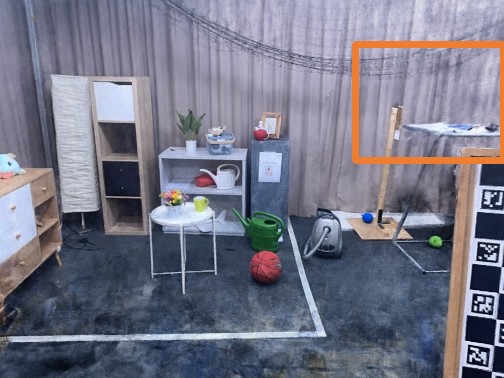} &
    \includegraphics[width=\sz\linewidth]{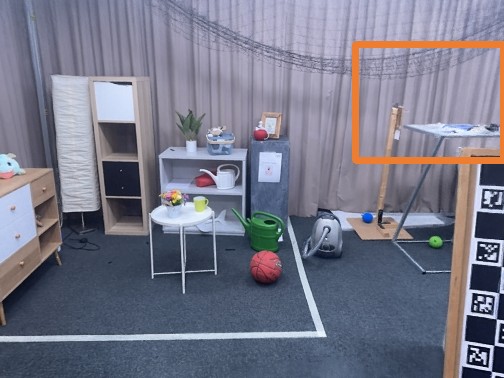}  &
    \includegraphics[width=\sz\linewidth]{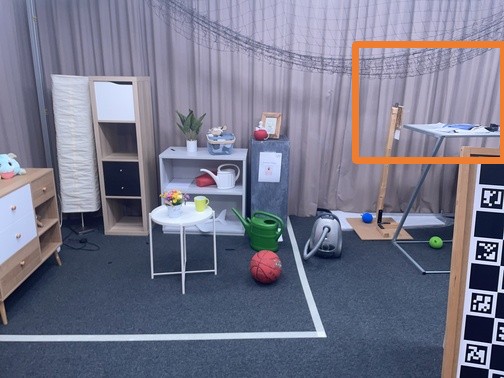} 
     \\
    & \includegraphics[width=\sz\linewidth]{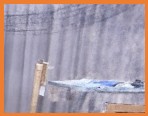} &
    \includegraphics[width=\sz\linewidth]{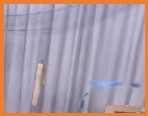} &
    \includegraphics[width=\sz\linewidth]{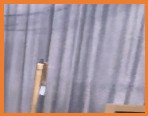} &
    \includegraphics[width=\sz\linewidth]{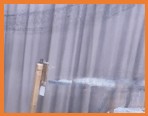} &
    \includegraphics[width=\sz\linewidth]{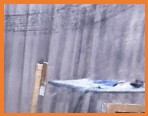} &
    \includegraphics[width=\sz\linewidth]{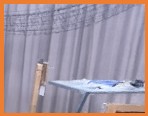}  &
    \includegraphics[width=\sz\linewidth]{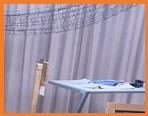} 
    \\
    \multirow{2}{*}{\rotatebox{90}{Patio}} &
    \includegraphics[width=\sz\linewidth]{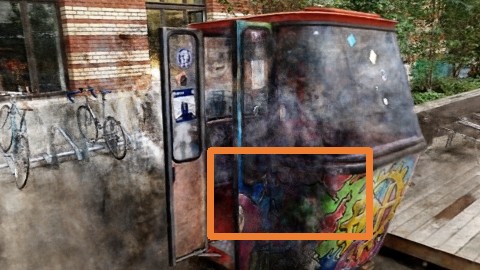} &
    \includegraphics[width=\sz\linewidth]{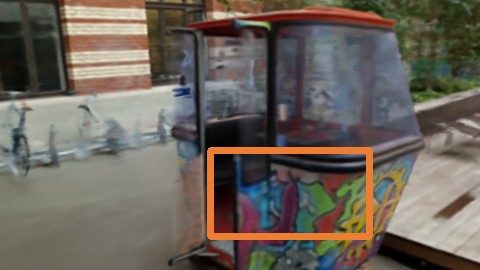} &
    \includegraphics[width=\sz\linewidth]{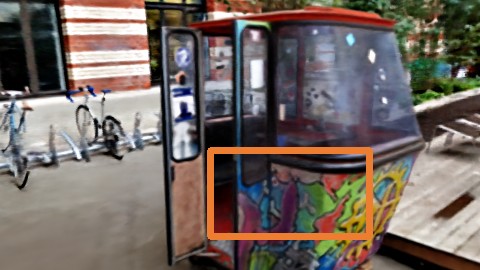} &
    \includegraphics[width=\sz\linewidth]{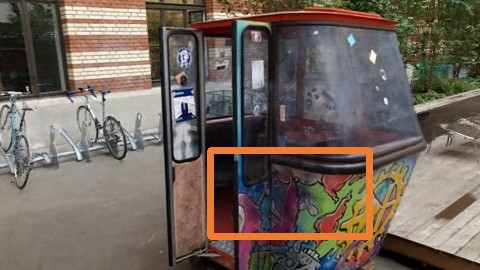} &
    \includegraphics[width=\sz\linewidth]{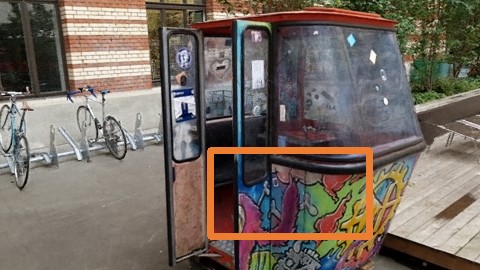} &
    \includegraphics[width=\sz\linewidth]{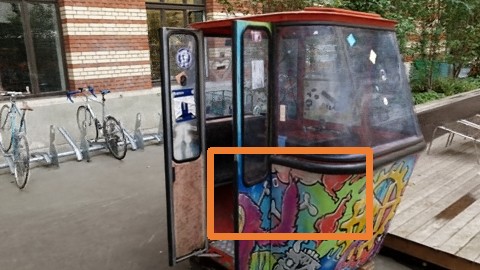}  &
    \includegraphics[width=\sz\linewidth]{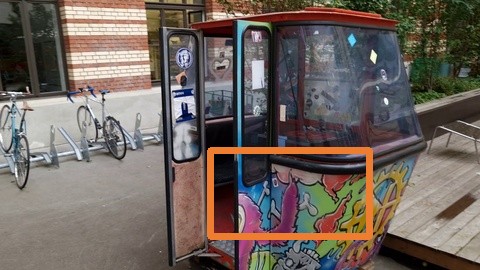} 
     \\
    & \includegraphics[width=\sz\linewidth]{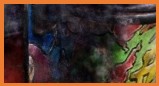} &
    \includegraphics[width=\sz\linewidth]{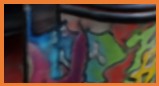} &
    \includegraphics[width=\sz\linewidth]{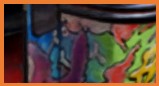} &
    \includegraphics[width=\sz\linewidth]{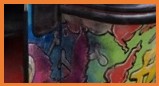} &
    \includegraphics[width=\sz\linewidth]{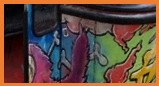} &
    \includegraphics[width=\sz\linewidth]{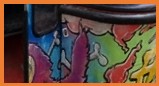}  &
    \includegraphics[width=\sz\linewidth]{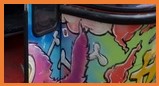} 
    \\
    \multirow{2}{*}{\rotatebox{90}{Spot}} &
    \includegraphics[width=\sz\linewidth]{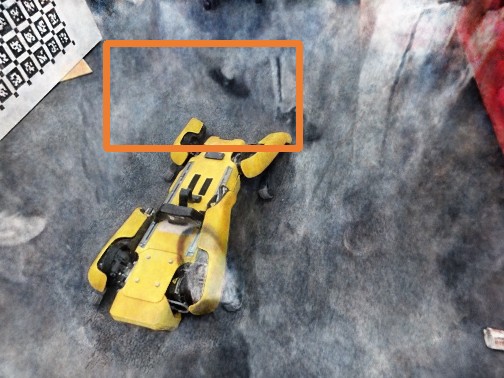} &
    \includegraphics[width=\sz\linewidth]{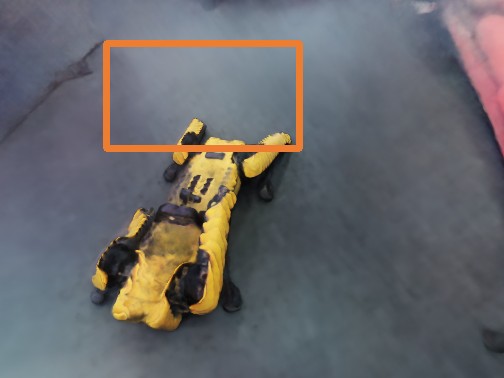} &
    \includegraphics[width=\sz\linewidth]{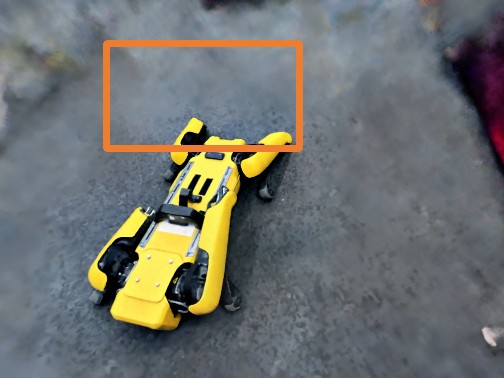} &
    \includegraphics[width=\sz\linewidth]{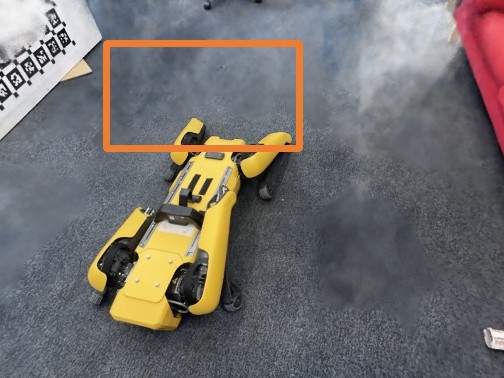} &
    \includegraphics[width=\sz\linewidth]{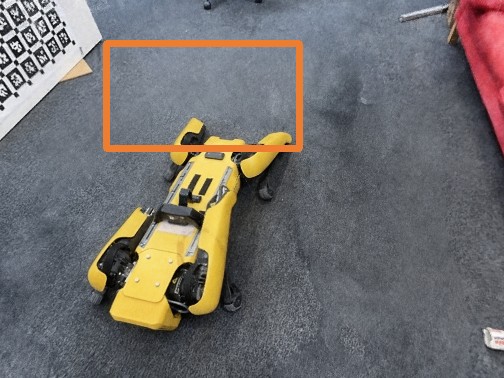} &
    \includegraphics[width=\sz\linewidth]{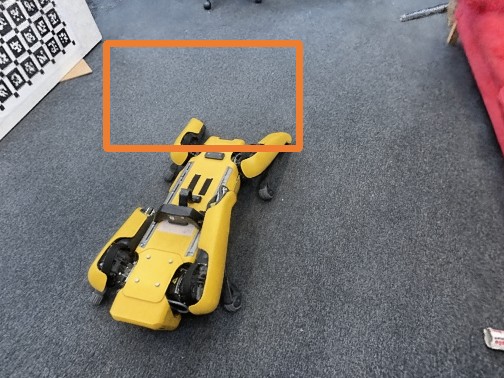}  &
    \includegraphics[width=\sz\linewidth]{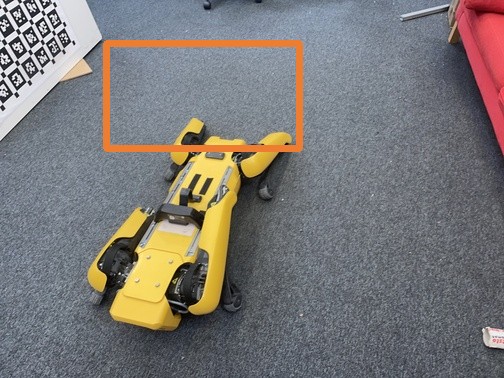} 
     \\
    & \includegraphics[width=\sz\linewidth]{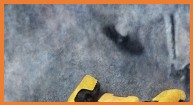} &
    \includegraphics[width=\sz\linewidth]{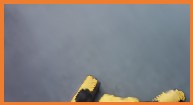} &
    \includegraphics[width=\sz\linewidth]{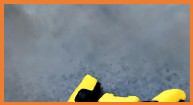} &
    \includegraphics[width=\sz\linewidth]{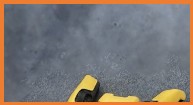} &
    \includegraphics[width=\sz\linewidth]{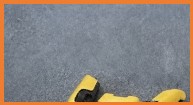} &
    \includegraphics[width=\sz\linewidth]{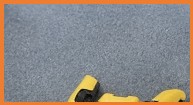}  &
    \includegraphics[width=\sz\linewidth]{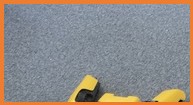} 
    \\
    \multirow{2}{*}{\rotatebox{90}{Patio-High}} &
    \includegraphics[width=\sz\linewidth]{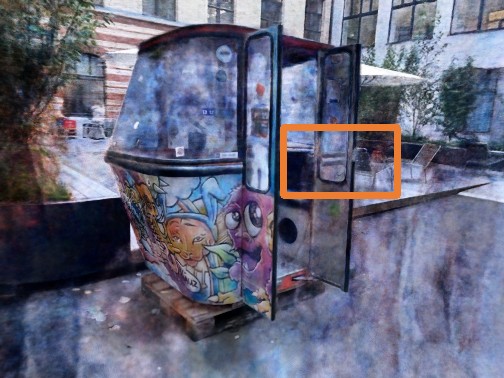} &
    \includegraphics[width=\sz\linewidth]{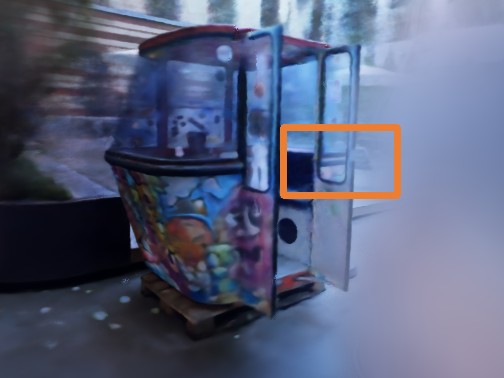} &
    \includegraphics[width=\sz\linewidth]{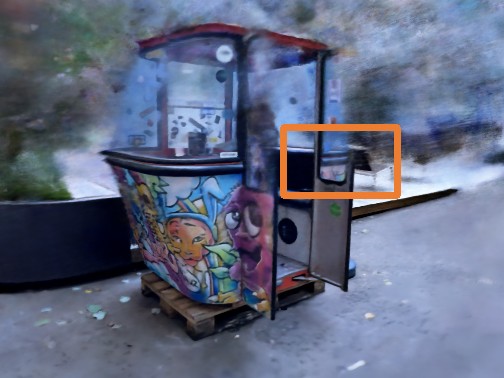} &
    \includegraphics[width=\sz\linewidth]{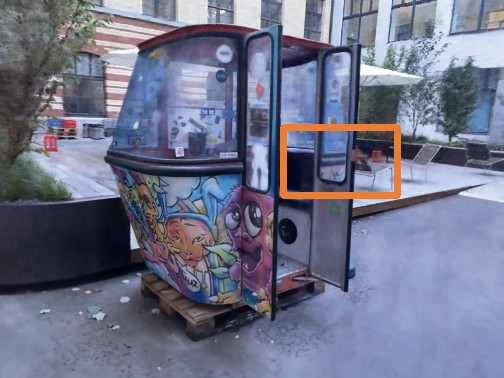} &
    \includegraphics[width=\sz\linewidth]{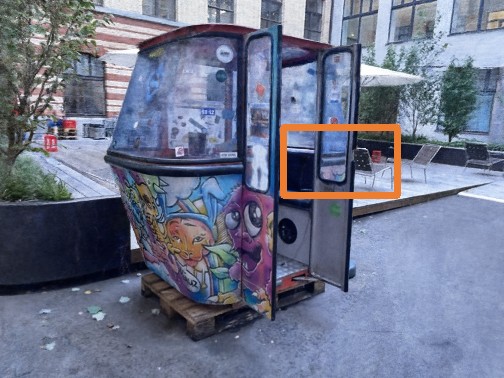} &
    \includegraphics[width=\sz\linewidth]{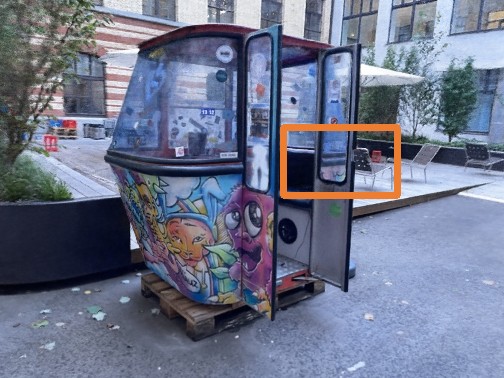}  &
    \includegraphics[width=\sz\linewidth]{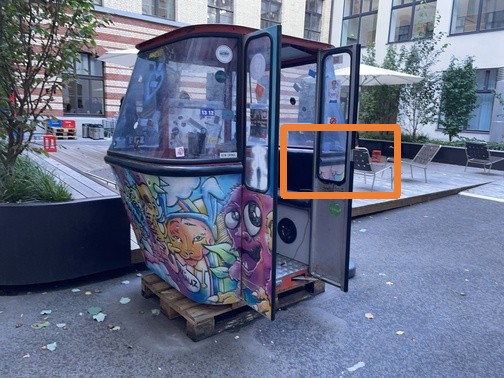} 
     \\
    & \includegraphics[width=\sz\linewidth]{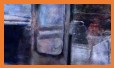} &
    \includegraphics[width=\sz\linewidth]{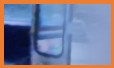} &
    \includegraphics[width=\sz\linewidth]{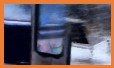} &
    \includegraphics[width=\sz\linewidth]{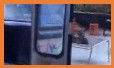} &
    \includegraphics[width=\sz\linewidth]{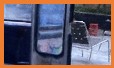} &
    \includegraphics[width=\sz\linewidth]{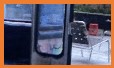}  &
    \includegraphics[width=\sz\linewidth]{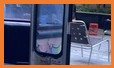} 
    \\
    & {\mipNeRFthreesixty}   & {\NeRFW} & {\hanerf} & {\RobustNeRF} & {Mip-NeRF 360 + SAM} & {\textbf{\ours}}  &  {GT} \\

  \end{tabular} 
  \caption{\textbf{Novel View Synthesis Results on Our \textit{On-the-go} Dataset.} Our method constantly outperforms baseline methods on scenes with various ratios of distractors, from confined indoor scenes with objects to large outdoor scenes.}
  
  \label{fig:self}
\end{figure*} 

\begin{table*}[t]
\setlength{\tabcolsep}{0.1cm}
\resizebox{\linewidth}{!}{
\begin{tabular}{l|ccc|ccc|ccc|ccc|ccc|ccc}
\multicolumn{1}{c|}{} & \multicolumn{6}{c|}{Low Occlusion}  & \multicolumn{6}{c|}{Medium Occlusion} & \multicolumn{6}{c}{High Occlusion} \\
\multicolumn{1}{c|}{}  & \multicolumn{3}{c|}{Mountain}  & \multicolumn{3}{c|}{Fountain} & \multicolumn{3}{c|}{Corner}  & \multicolumn{3}{c|}{Patio} & \multicolumn{3}{c|}{Spot} & \multicolumn{3}{c}{Patio-High} \\
& \lpips & \ssim & \psnr& \lpips & \ssim & \psnr& \lpips & \ssim & \psnr&  \lpips & \ssim & \psnr &  \lpips & \ssim & \psnr & \lpips & \ssim & \psnr \\
\midrule
\mipNeRFthreesixty &  0.295 &  0.601 & 19.64 & 0.556 &  0.290 & 13.91 &  0.345 &  0.660 & 20.41 &  0.421 &  0.503 & 15.48 & 0.469 & 0.306 & 17.82 & 0.486 & 0.432 & 15.73\\
\NeRFW  &  0.491 &  0.492 & 18.07 &  0.546 & 0.410 & 17.20 &  0.349 &  0.708 & 20.21 & 0.445 &  0.532 & 17,55 & 0.690 & 0.384 & 16.40 & 0.606 & 0.349 & 12.99\\
\hanerf  &  0.499 &  0.485 & 18.64 & 0.569 & 0.393 & 16.71 &  0.367 &  0.684 & 19.23 & 0.393 & 0.543 & 16.82 & 0.599 & 0.460 & 17.85 & 0.505 & 0.463 & 16.67\\
\RobustNeRF & 0.383 & 0.496 & 17.54 & 0.576 & 0.318 & 15.65  & 0.244 & 0.764 & 23.04 & 0.251 & 0.718 & 20.39  & 0.391 & 0.625 & 20.65  & 0.366 & 0.578 & 20.54\\ 
Mip-NeRF 360 + SAM  &\textbf{ 0.258} & 0.642 & \textbf{20.20} & 0.556 & 0.287 & 13.65  & 0.332 & 0.670 & 20.65 & 0.227 & 0.738 & 20.83 & 0.323 & 0.542 & 21.08 & 0.326 & 0.576 & 20.13 \\
\textbf{\ours} & 0.259 & \textbf{0.644} & 20.15 &\textbf{0.314} & \textbf{0.609} & \textbf{20.11} & \textbf{0.190} & \textbf{0.806} & \textbf{24.22} & \textbf{0.219} & \textbf{0.754} & \textbf{20.78} & \textbf{0.189} & \textbf{0.787} & \textbf{23.33} & \textbf{0.235} & \textbf{0.718} & \textbf{21.41} \\ 
\end{tabular}
}

\caption{\textbf{Novel View Synthesis Results on Our \textit{On-the-go} Dataset.} We show quantitative comparison between our methods and baselines.}
\label{table:self}
\end{table*}

\boldparagraph{\textit{On-the-go} Dataset}
We extend our evaluation on our \textit{On-the-go} dataset, as depicted in \figref{fig:self_sample} and \tabref{table:self}. Compared to our method, RobustNeRF often fails to retain fine details in low to medium-occlusion scenarios, and struggles to eliminate distractors in high-occlusion settings. 
Besides, we notice that even after tuning the hyperparameter of outlier ratios for highly-occluded scenes, RobustNeRF still shows inferior performance.
Please refer to the supplements.

Unlike RobustNeRF, NeRF-W and Ha-NeRF show proficiency in removing distractors at low and medium occlusion levels, but this effectiveness comes at the cost of reduced image quality. This trade-off is a consequence of its transient embedding approach, as discussed in~\cite{pan2022activenerf, robustnerf}.
Furthermore, NeRF-W and Ha-NeRF struggle notably at higher occlusion ratios. In such cases, their per-image transient embeddings are unable to adequately model distractors, leading to a noticeable performance drop.
The Mip-NeRF 360 combined with SAM method works well in simple scenes like \textit{Mountain}, where distractors are easy to segment. However, its effectiveness diminishes in more complex scenes. In contrast, we exhibit versatility across scenes with various occlusion ratios, and can consistently produce high-quality renderings.

\boldparagraph{Comparison on RobustNeRF Dataset~\cite{robustnerf}}
As shown in \tabref{table:robust}, our method exhibits superior performance quantitatively and qualitatively over all baselines.
RobustNeRF's hard-thresholding approach tends to overlook complex structures with limited observations, such as the shoes and carpet in the \textit{Android} scene. Moreover, we observed that they underperform in scenarios involving plane surfaces with view-dependent effects, \eg the wooden texture on the table with view-dependent highlight in \textit{Statue} scene.
Note that Mip-NeRF 360 + SAM requires a tedious process of manually selecting every distractor in each image using SAM~\cite{kirillov2023segment}, but it still struggles with capturing thin structures, shadows, and reflections.

\begin{table}[t]
  \centering
  \footnotesize
  \setlength{\tabcolsep}{0.07cm}
  \resizebox{\linewidth}{!}{
  \begin{tabular}{l|ccc|ccc}
  & \multicolumn{3}{c|}{Statue} 
      & \multicolumn{3}{c}{Android} 
  \\
                              & \lpips & \ssim & \psnr & \lpips & \ssim & \psnr    \\
  \midrule
  \mipNeRFthreesixty     & 0.36  & 0.66 & 19.09 &  0.40  & 0.65 & 19.35       \\
  \ddnerf & 0.48  & 0.49 &19.09 &   0.43 & 0.57 & 20.61   \\
  \RobustNeRF & 0.28 & 0.75 & 20.89 &  0.31 & 0.65 & 21.72   \\
  \RobustNeRFStar  &  0.27 & 0.73  & 21.13 & 0.22 & 0.73 & 22.83 \\
  Mip-NeRF 360 + SAM &   \textbf{0.23} &  0.74 & 21.30 & 0.23 & 0.71 & 22.62 \\
  \textbf{\ours}     &  0.24 &\textbf{ 0.77} & \textbf{21.58} &\textbf{ 0.21} &\textbf{ 0.75} & \textbf{23.50}\\
  \hline
  \end{tabular}
  }

  \setlength{\tabcolsep}{0.5pt}
  \newcommand{\sz}{0.235}
  \begin{tabular}{ccccc}
  \\
  \multirow{2}{*}{\rotatebox{90}{Android}} &
    \includegraphics[width=\sz\linewidth]{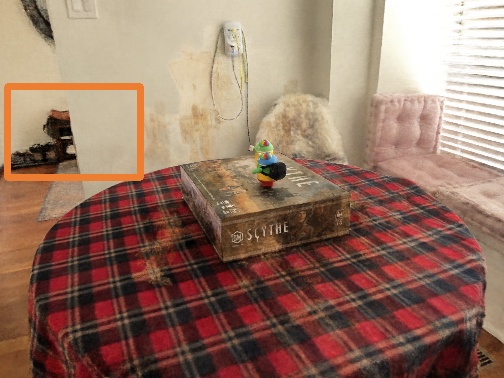} &
    \includegraphics[width=\sz\linewidth]{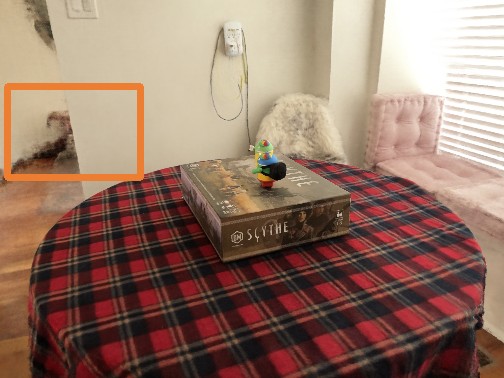} &
    \includegraphics[width=\sz\linewidth]{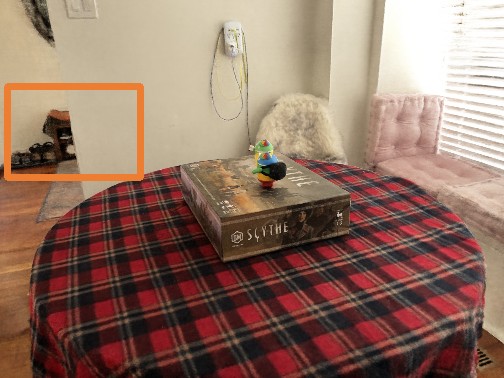} &
    \includegraphics[width=\sz\linewidth]{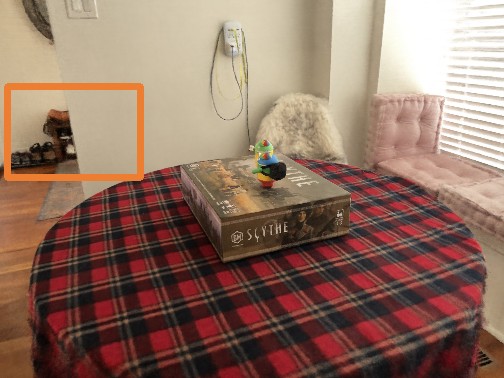} 
    \\
    & \includegraphics[width=\sz\linewidth]{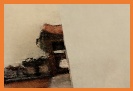} &
    \includegraphics[width=\sz\linewidth]{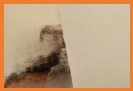} &
    \includegraphics[width=\sz\linewidth]{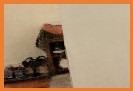} &
    \includegraphics[width=\sz\linewidth]{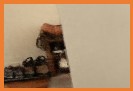} 
    \\
    \multirow{2}{*}{\rotatebox{90}{Statue}} &
    \includegraphics[width=\sz\linewidth]{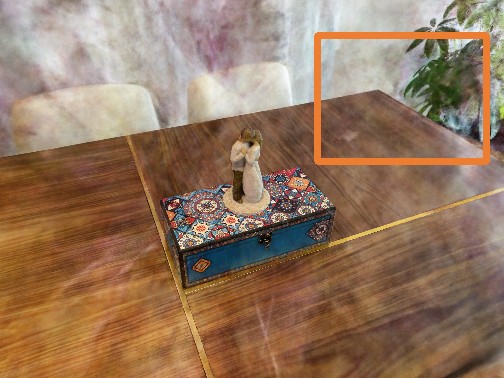} & 
    \includegraphics[width=\sz\linewidth]{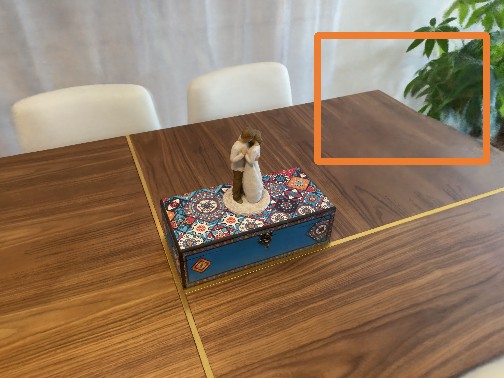} &
    \includegraphics[width=\sz\linewidth]{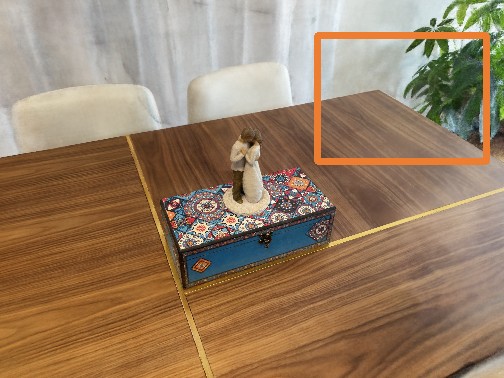} & 
    \includegraphics[width=\sz\linewidth]{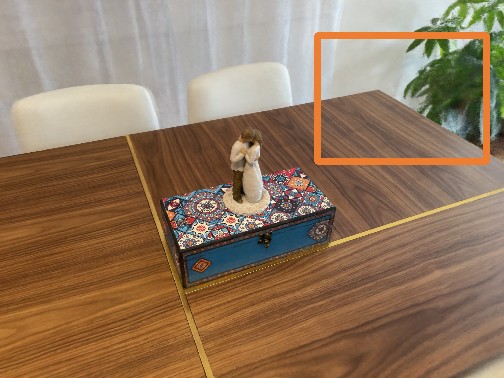} 
    \\
    & \includegraphics[width=\sz\linewidth]{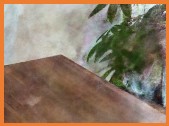} & 
    \includegraphics[width=\sz\linewidth]{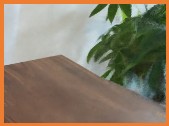} &
    \includegraphics[width=\sz\linewidth]{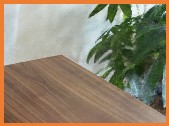} & 
    \includegraphics[width=\sz\linewidth]{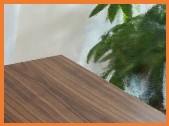} 
    \\
    & {Mip-NeRF 360}   & {RobustNeRF$^*$} &  {\fontsize{6.5}{7}\selectfont Mip-NeRF360+SAM} & \textbf{Ours}  \\
  \end{tabular} 
  \caption{\textbf{Novel View Synthesis Results on the RobustNeRF Dataset.} The numbers for \mipNeRFthreesixty, \ddnerf and \RobustNeRF are taken from~\cite{robustnerf}. \RobustNeRFStar denotes our own run using the official code release.}

  \label{table:robust}
\end{table}

\subsection{Ablation Study}
All ablations are conducted on the challenging highly-occluded ``Patio-High'' scene in our \textit{On-the-go} dataset. 

\paragraph{Patch Dilation}
Here we test different dilation rates for our dilation patch sampling, as shown in~\tabref{tab:ablation_dilation}.
Within a range from 1 to 4, a higher dilation rate results in much faster convergence and better rendering quality. 
This verifies our hypothesis in~\secref{method:dilated_patch} that increasing the contextual information within patches can effectively boost performance.
However, when the dilation rate is above 4, uncertainty optimization begins to collapse.
It is likely because higher dilation rates cause patches to lose semantic information. 
This occurs as the sampling now becomes more akin to random sampling, negatively impacting the learning of uncertainty.
Further details and analysis on patch size and dilation rate across different sequences are available in the supplements.

\paragraph{Loss Functions}
In \tabref{tab:ablation}, we ablate on different training losses. 
In (b), SSIM proves more adept at differentiating distractors with static elements compared to $\ell_2$ loss. 
In (c), we train the uncertainty MLP and NeRF together. 
This results in a significant performance drop, indicating the effectiveness of our decoupled training approach.
Moreover, we find from (a) that omitting \(\mathcal{L}_{reg}\) will negatively impact the rendering quality of certain views.
Additional studies on various sequences are available in the supplements.

\begin{table}[t]
  \centering
  \setlength{\tabcolsep}{0.05cm}
  \begin{tabular}{cc}
      \resizebox{0.16\textwidth}{!}{
          
\begin{tabular}{l|cccc}
& \lpips & \ssim & \psnr \\
\midrule
1 &  0.451 & 0.515 & 17.82  \\
2 &  0.262 & 0.692 & 20.70 & \\
\textbf{4}  & \textbf{0.235} & \textbf{0.718} & \textbf{21.41} &  \\ 
8 & 0.392 & 0.529 & 18.22  \\ 
16 & 0.477 & 0.439 & 16.08  \\ 
\end{tabular}
}&
      \begin{minipage}{0.3\textwidth}\includegraphics[width=1\textwidth]{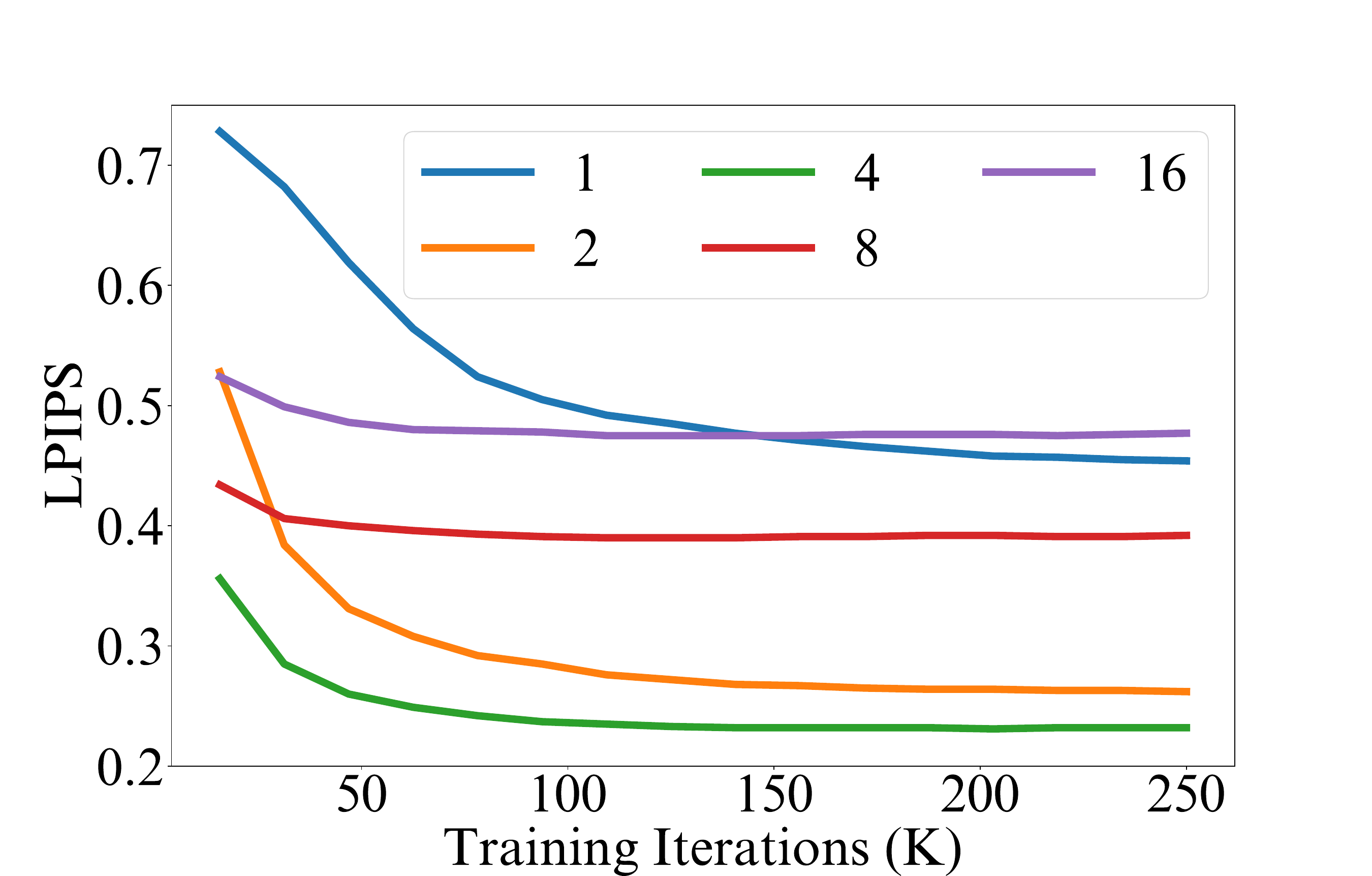}
      \end{minipage}
  \end{tabular}
  \caption{\textbf{Ablations on Patch Dilation Rates}. Comparisons of various dilation rates for the dilated patch sampling, with a patch size of $32\times32$.}
  \label{tab:ablation_dilation}
\end{table}

\begin{table}[t]
    \centering
    \footnotesize
    \newcommand{\sz}{0.195}
    \newcommand{\sza}{0.1729} %
    \setlength{\tabcolsep}{8pt}
    \resizebox{0.8\linewidth}{!}{\begin{tabular}{l|ccc}
& \lpips & \ssim & \psnr \\
\midrule
(a) w/o $\mathcal{L}_\text{reg}$ &  0.261 & 0.698 & 21.02\\
(b) $\ell_2$ in $\mathcal{L}_\text{uncer}$ & 0.437 &  0.492&  17.13 \\
(c) $\mathcal{L}_\text{uncer}$ for NeRF & 0.496 & 0.437 & 16.70\\
\textbf{\ours} & \textbf{0.235} & \textbf{0.718} & \textbf{21.41}  \\ 
\end{tabular}}
    \setlength{\tabcolsep}{0.5pt}
    \begin{tabular}{ccccccc} 
    \\
    \includegraphics[width=\sz\linewidth]{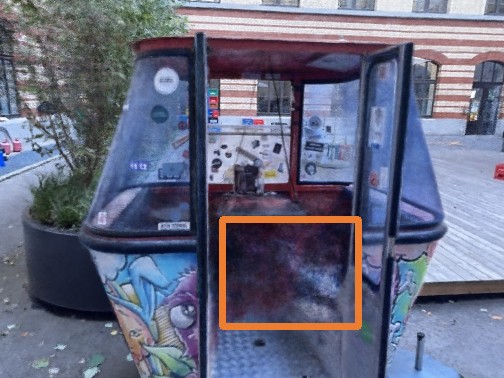} & 
    \includegraphics[width=\sz\linewidth]{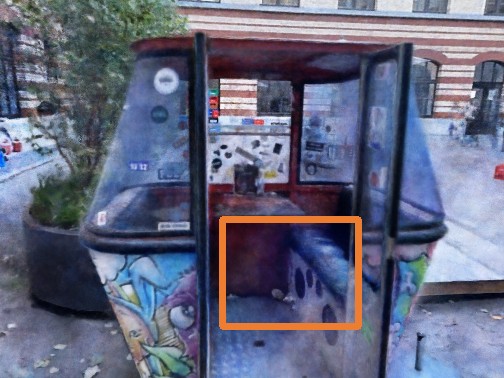} & 
    \includegraphics[width=\sz\linewidth]{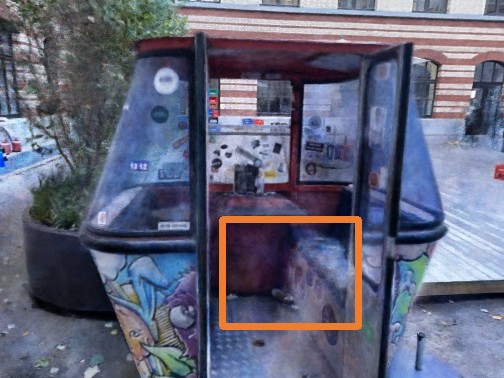} & 
    \includegraphics[width=\sz\linewidth]{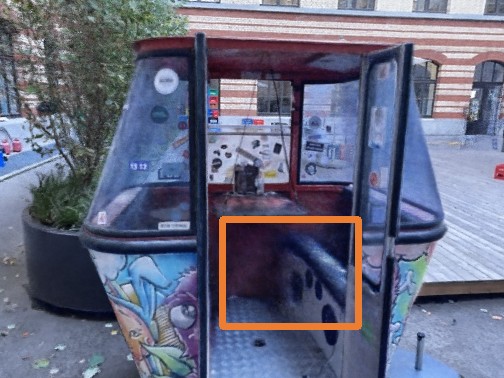} & 
    \includegraphics[width=\sz\linewidth]{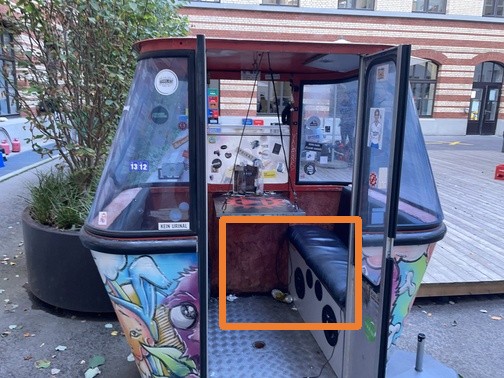} \\
    \includegraphics[width=\sz\linewidth]{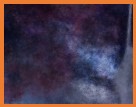} & 
    \includegraphics[width=\sz\linewidth]{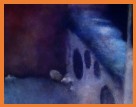} & 
    \includegraphics[width=\sz\linewidth]{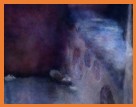} & 
    \includegraphics[width=\sz\linewidth]{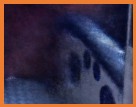} & 
    \includegraphics[width=\sz\linewidth]{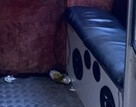} \\
    {(a)} & {(b)} & {(c)} &  {\textbf{\ours}} &  {GT}\\

    \end{tabular} 
    \caption{\textbf{Ablations on Loss Functions.} We compare different loss choices for training our system. 
    }
    \label{tab:ablation}
  \end{table}

\subsection{Analysis}
 
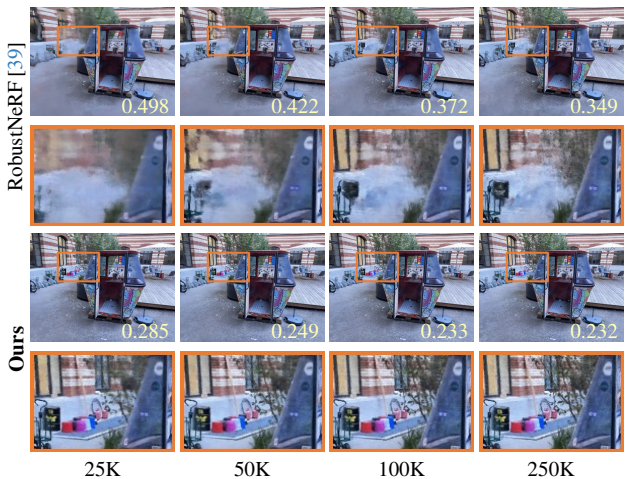
\begin{figure}[t]
  \centering
  \footnotesize
  \setlength{\tabcolsep}{1.0pt}
  \newcommand{\sz}{0.23}
  \newcommand{\sza}{0.1729} %
  \begin{tabular}{ccccc}
  \multirow{2}{*}{\rotatebox{90}{\parbox[c]{1.2cm}{\RobustNeRF}}} & 
  \begin{tikzpicture}
    \node[anchor=south west,inner sep=0] (image) at (0,0) {
      \includegraphics[width=\sz\linewidth]{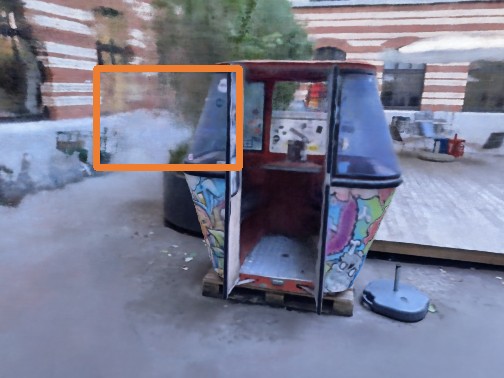} %
    };
    \node[anchor=south east,inner sep=2pt, text=lightyellow, fill=none, yshift=-1pt] at (image.south east) {0.498};
  \end{tikzpicture}
  &
  \begin{tikzpicture}
    \node[anchor=south west,inner sep=0] (image) at (0,0) {
      \includegraphics[width=\sz\linewidth]{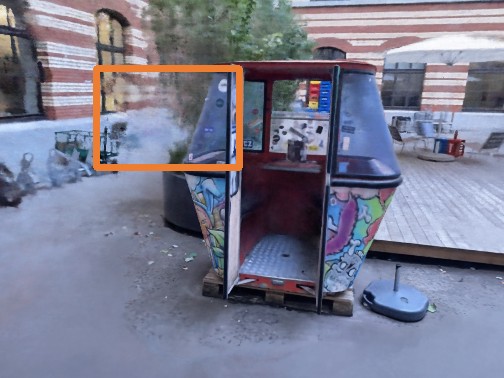} %
    };
    \node[anchor=south east,inner sep=2pt, text=lightyellow, fill=none, yshift=-1pt] at (image.south east) {0.422};
  \end{tikzpicture}
  &
  \begin{tikzpicture}
    \node[anchor=south west,inner sep=0] (image) at (0,0) {
      \includegraphics[width=\sz\linewidth]{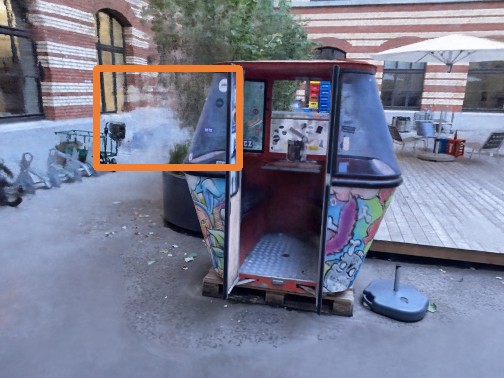} %
    };
    \node[anchor=south east,inner sep=2pt, text=lightyellow, fill=none, yshift=-1pt] at (image.south east) {0.372};
  \end{tikzpicture}
  &
  \begin{tikzpicture}
    \node[anchor=south west,inner sep=0] (image) at (0,0) {
      \includegraphics[width=\sz\linewidth]{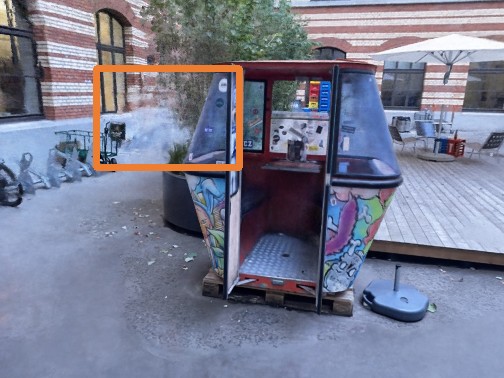} %
    };
    \node[anchor=south east,inner sep=2pt, text=lightyellow, fill=none, yshift=-1pt] at (image.south east) {0.349};
  \end{tikzpicture}
 \\
 &
  \includegraphics[width=\sz\linewidth]{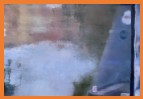} &
  \includegraphics[width=\sz\linewidth]{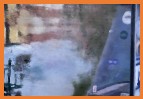} &
  \includegraphics[width=\sz\linewidth]{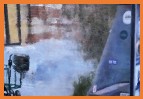} &
  \includegraphics[width=\sz\linewidth]{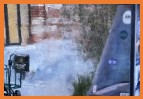}  \\
  \multirow{2}{*}{\rotatebox{90}{\textbf{\ours}}} & 
  \begin{tikzpicture}
    \node[anchor=south west,inner sep=0] (image) at (0,0) {
      \includegraphics[width=\sz\linewidth]{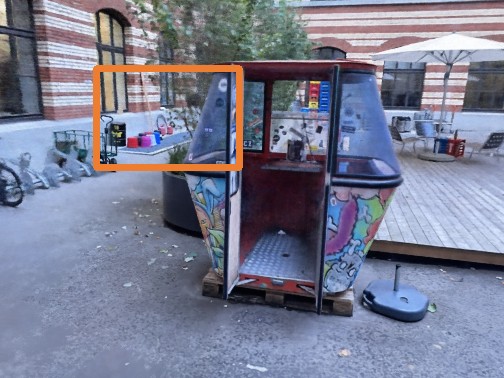} %
    };
    \node[anchor=south east,inner sep=2pt, text=lightyellow, fill=none, yshift=-1pt] at (image.south east) {0.285};
  \end{tikzpicture}
  &
  \begin{tikzpicture}
    \node[anchor=south west,inner sep=0] (image) at (0,0) {
      \includegraphics[width=\sz\linewidth]{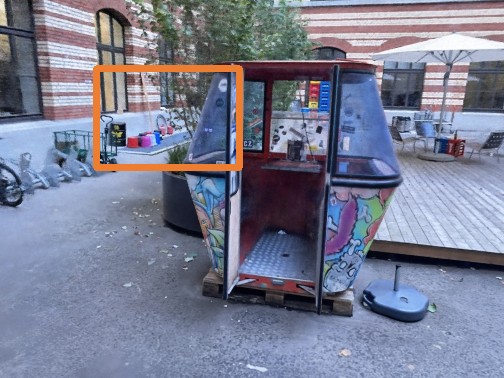} %
    };
    \node[anchor=south east,inner sep=2pt, text=lightyellow, fill=none, yshift=-1pt] at (image.south east) {0.249};
  \end{tikzpicture}
  &
  \begin{tikzpicture}
    \node[anchor=south west,inner sep=0] (image) at (0,0) {
      \includegraphics[width=\sz\linewidth]{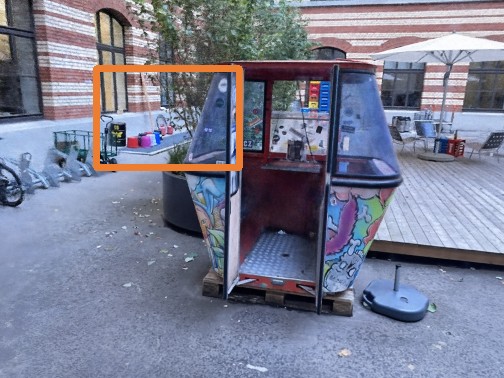} %
    };
    \node[anchor=south east,inner sep=2pt, text=lightyellow, fill=none, yshift=-1pt] at (image.south east) {0.233};
  \end{tikzpicture}
  &
  \begin{tikzpicture}
    \node[anchor=south west,inner sep=0] (image) at (0,0) {
      \includegraphics[width=\sz\linewidth]{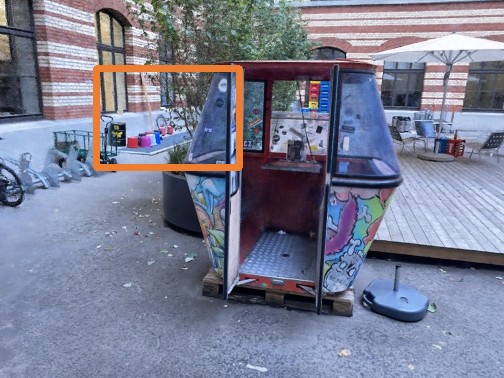} %
    };
    \node[anchor=south east,inner sep=2pt, text=lightyellow, fill=none, yshift=-1pt] at (image.south east) {0.232};
  \end{tikzpicture}
 \\
 &
  \includegraphics[width=\sz\linewidth]{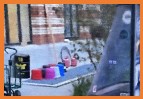} &
  \includegraphics[width=\sz\linewidth]{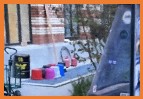} &
  \includegraphics[width=\sz\linewidth]{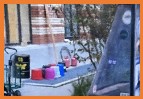} &
  \includegraphics[width=\sz\linewidth]{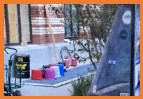}  \\
  & 25K & 50K & 100K & 250K \\
  \end{tabular} 
  \caption{\textbf{Convergence Speed Comparison.} LPIPS metrics are included in images. Our method can already capture better details at 25K iterations than RobustNeRF at 250K iterations.
  } 
  
  \label{fig:converge}
\end{figure}

\boldparagraph{Fast Convergence}\label{para:fast_conv}
\figref{fig:converge} presents a comparison between RobustNeRF and ours during training processes.
Thanks to our uncertainty prediction pipeline and dilated patch sampling, we show notably faster convergence. It can be noticed that we can already capture fine details from the early stages of training, see ours at 25K and RobustNeRF at 250K.

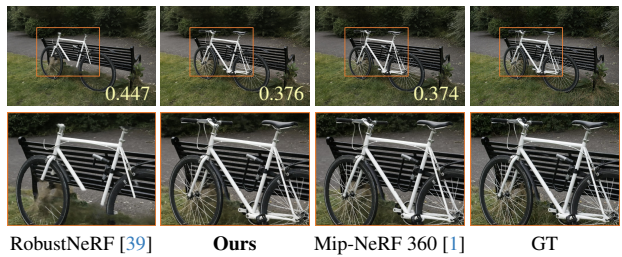
\begin{figure}
  \centering
  \footnotesize
  \setlength{\tabcolsep}{1pt}
  \newcommand{\sz}{0.25}
  \newcommand{\sza}{0.1729} %
  \resizebox{\linewidth}{!}{
  \begin{tabular}{cccc}
  \begin{tikzpicture}
    \node[anchor=south west,inner sep=0] (image) at (0,0) {
      \includegraphics[width=\sz\linewidth]{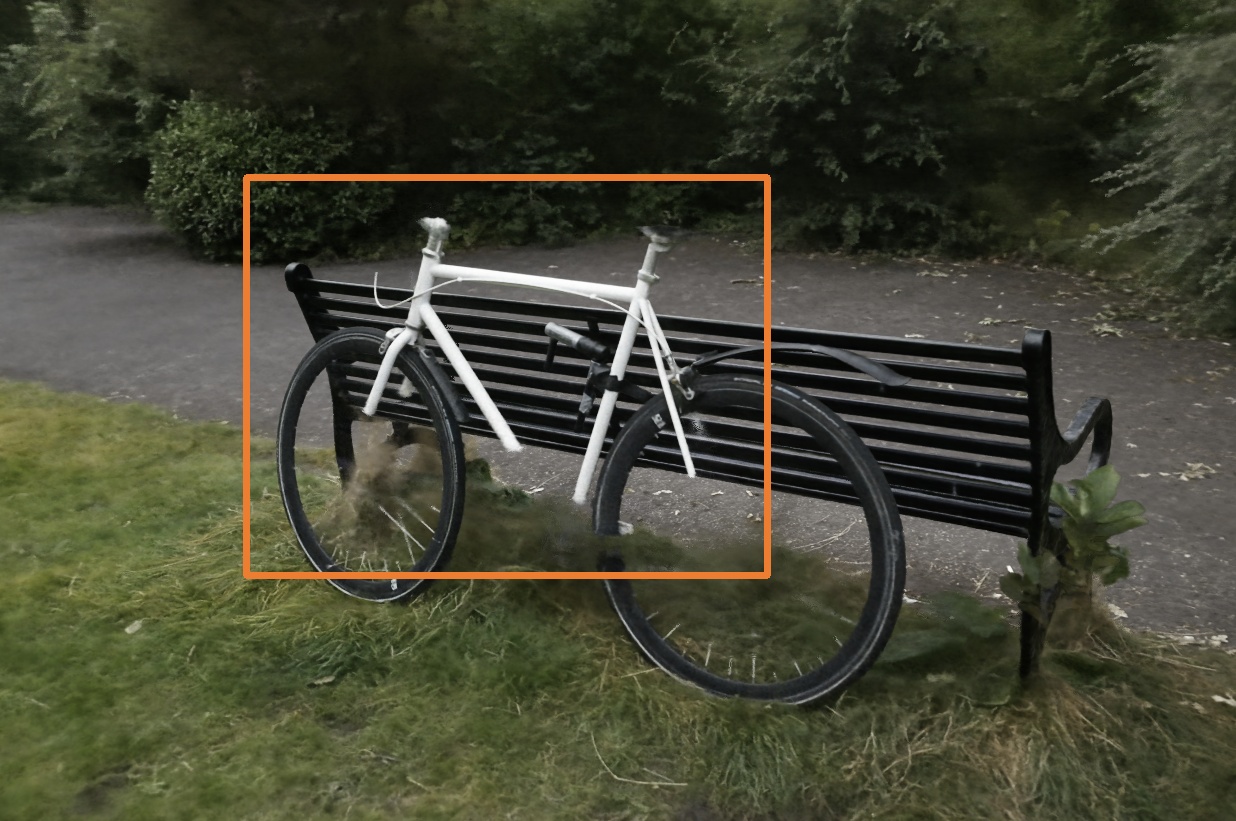} %
    };
    \node[anchor=south east,inner sep=2pt, text=lightyellow, fill=none, yshift=0pt] at (image.south east) {0.447};
  \end{tikzpicture}
    &
  \begin{tikzpicture}
    \node[anchor=south west,inner sep=0] (image) at (0,0) {
      \includegraphics[width=\sz\linewidth]{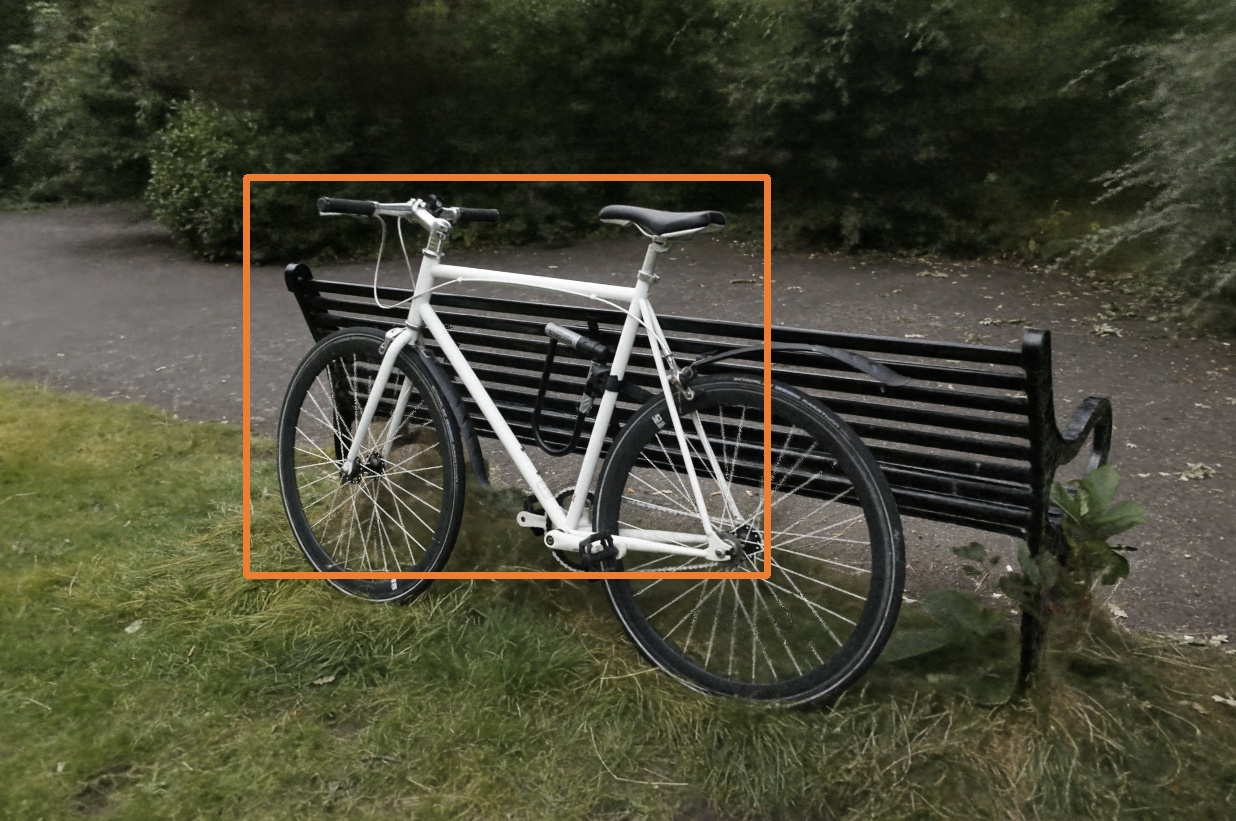} %
    };
    \node[anchor=south east,inner sep=2pt, text=lightyellow, fill=none, yshift=0pt] at (image.south east) {0.376};
  \end{tikzpicture}
    &
  \begin{tikzpicture}
    \node[anchor=south west,inner sep=0] (image) at (0,0) {
      \includegraphics[width=\sz\linewidth]{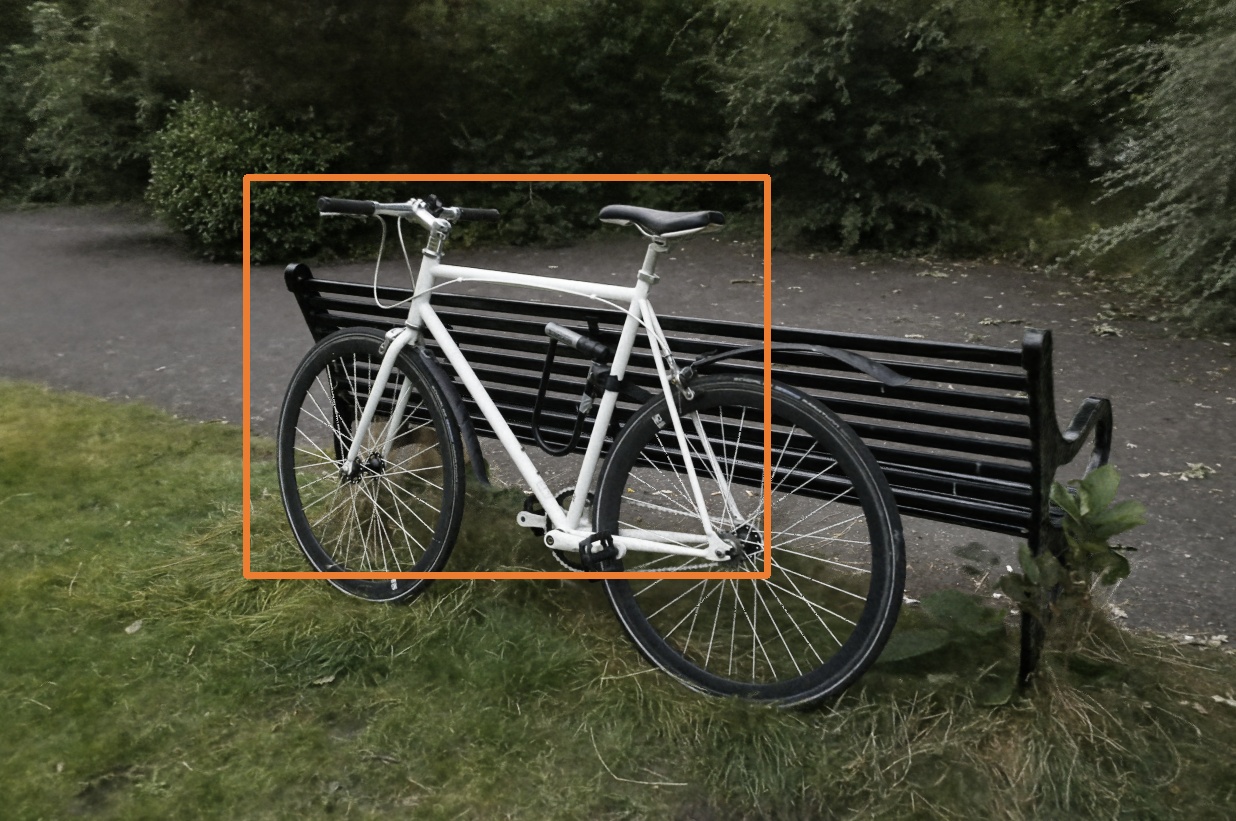} %
    };
    \node[anchor=south east,inner sep=2pt, text=lightyellow, fill=none, yshift=0pt] at (image.south east) {0.374};
  \end{tikzpicture}
    &

    \includegraphics[width=\sz\linewidth]{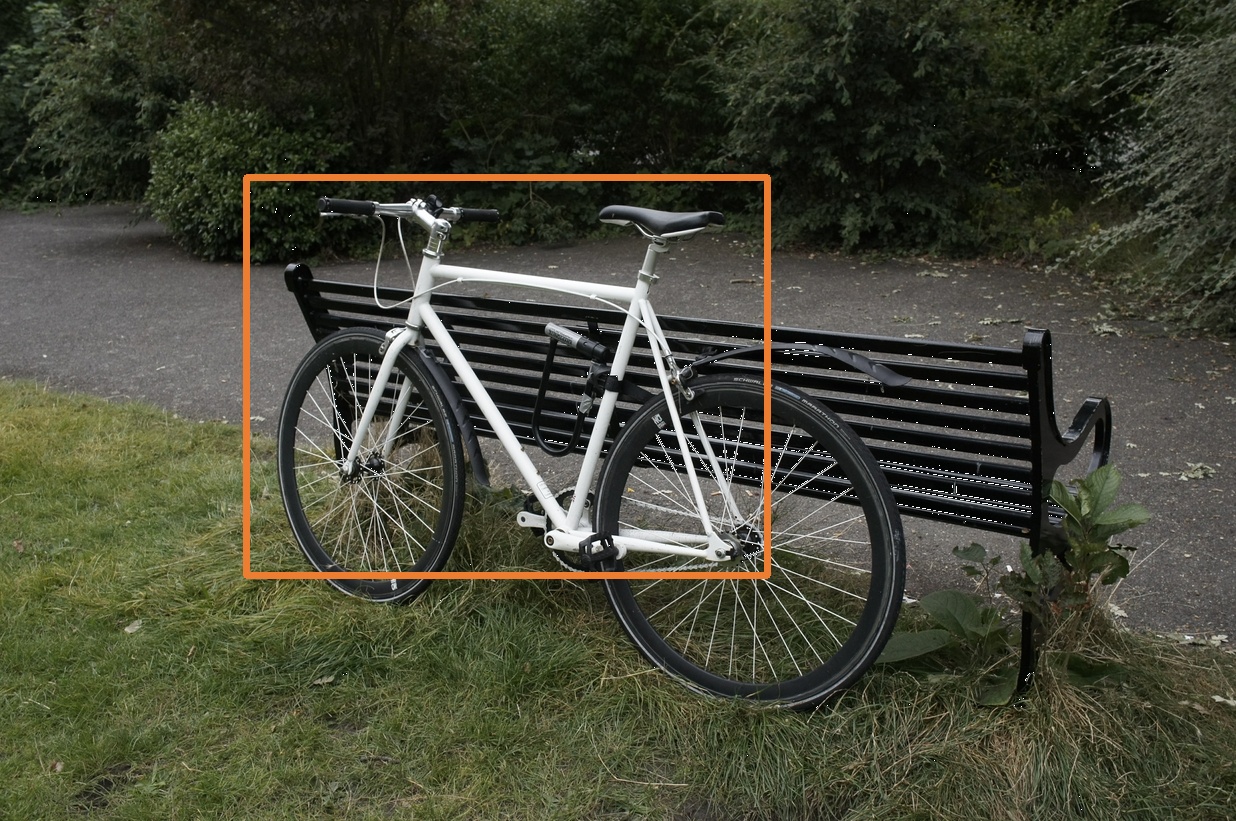} 
    \\
    \includegraphics[width=\sz\linewidth]{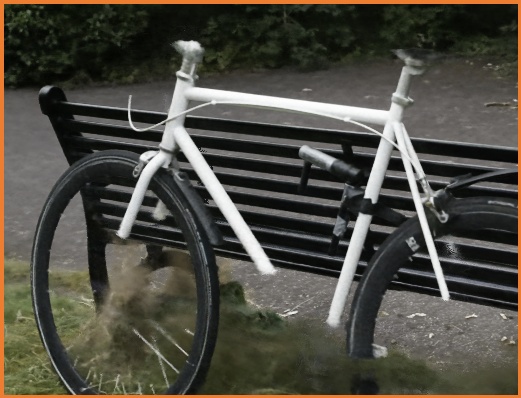} &
    \includegraphics[width=\sz\linewidth]{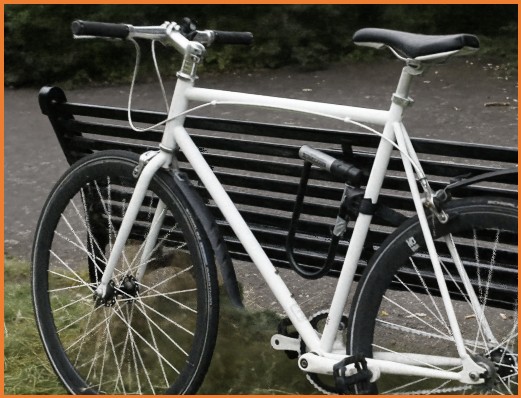} &
    \includegraphics[width=\sz\linewidth]{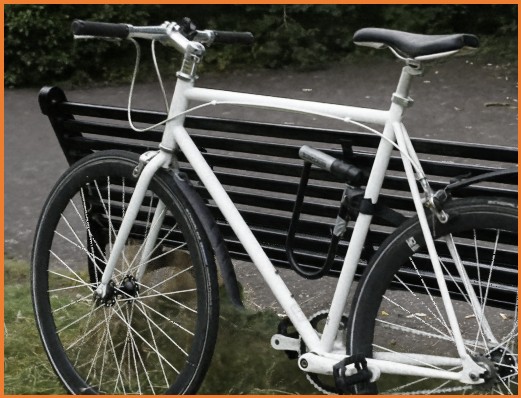} &
    \includegraphics[width=\sz\linewidth]{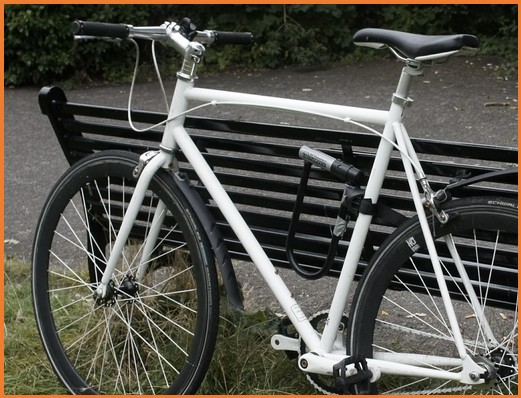} 
    \\

    \RobustNeRF & \textbf{Ours} & \mipNeRFthreesixty & GT  \\
\end{tabular} }

\caption{\textbf{Performance on Static Scenes.} LPIPS metrics are included in images. Our performance is much better than RobustNeRF and on par with the SOTA method~\cite{mipnerf360}.}
\vspace{0.5em}
\label{fig:static}

\end{figure}

\paragraph{Applicability to Static Scenes}
After showcasing our efficacy in building a NeRF from dynamic scenes, we explore whether it is directly adaptable to static scenes.
We evaluate using a static scene from the~\mipNeRFthreesixty dataset. 
As illustrated in \figref{fig:static}, we indeed achieve great performance as~\mipNeRFthreesixty. In contrast, RobustNeRF fails to capture certain parts of the bicycle, since one of their key designs involves omitting at least some portions of a scene.

\begin{figure}[t]
  \centering
  \footnotesize
  \setlength{\tabcolsep}{1.5pt}
  \newcommand{\sz}{0.24}
  \newcommand{\sza}{0.1729} %
  \begin{tabular}{cc|cc}
  \includegraphics[width=\sz\linewidth]{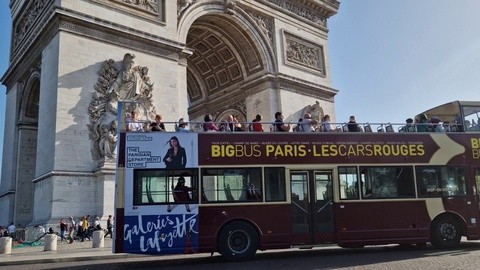} &
  \includegraphics[width=\sz\linewidth]{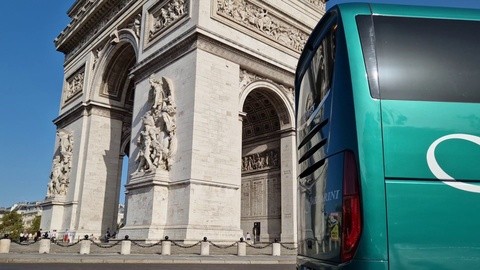} &
  \includegraphics[width=\sz\linewidth]{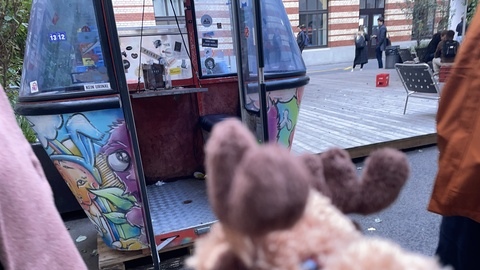} &
  \includegraphics[width=\sz\linewidth]{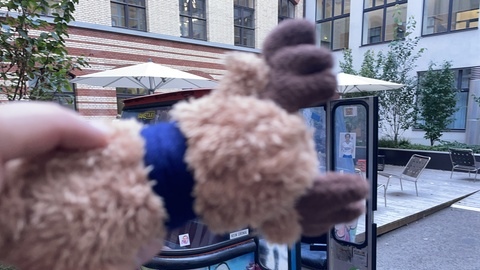}  \\
  \includegraphics[width=\sz\linewidth]{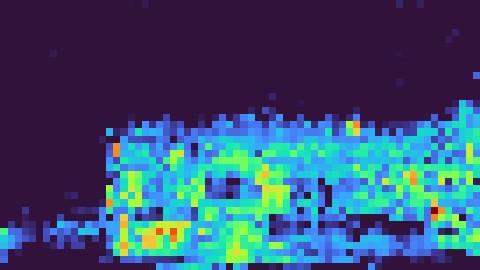} &
  \includegraphics[width=\sz\linewidth]{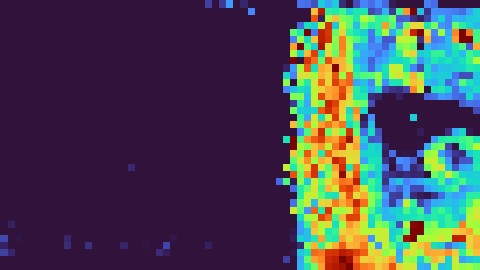} &
  \includegraphics[width=\sz\linewidth]{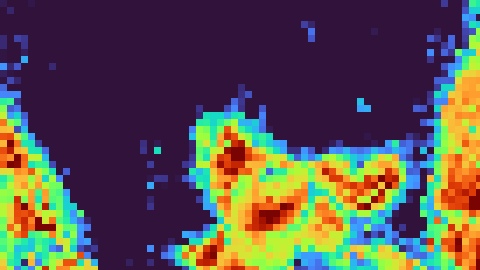} &
  \includegraphics[width=\sz\linewidth]{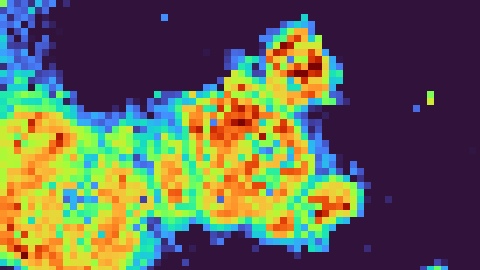}  \\
  \includegraphics[width=\sz\linewidth]{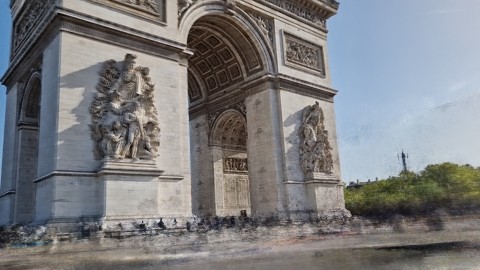} &
  \includegraphics[width=\sz\linewidth]{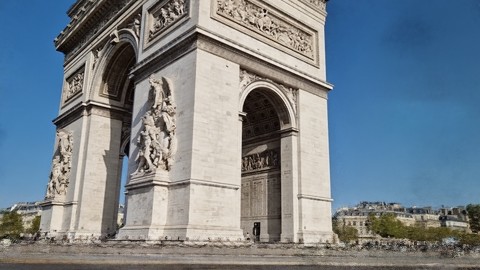} &
  \includegraphics[width=\sz\linewidth]{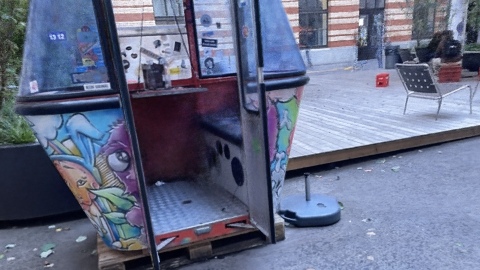} &
  \includegraphics[width=\sz\linewidth]{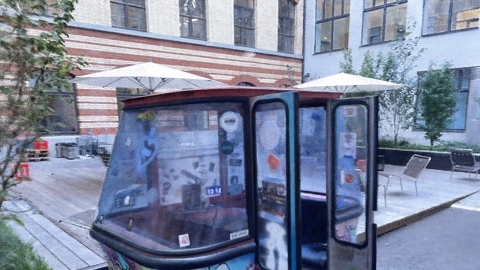}  \\
  
  \multicolumn{2}{c|}{Arc de Triomphe} &
    \multicolumn{2}{c}{Patio-High} \\
  \end{tabular}\vspace{-0.5em} 
  \caption{\textbf{Handling Large Obstructions.} From top to bottom: input frames, our uncertainty maps, our rendering results.
  }
  \label{fig:obstruct}
\end{figure}

\paragraph{Large Obstructions}
In~\figref{fig:obstruct}, we further show that our method can faithfully model the large obstructions with our predicted uncertainty, and effectively remove them.

\section{Conclusions}
\label{sec:conclusion}

We introduce NeRF~\emph{On-the-go}, a versatile method that enables effective and efficient distractor removal in dynamic real-world scenes containing various levels of distractors.
Our method represents a step towards realizing the full potential of NeRF in practical, in-the-wild applications.

\boldparagraph{Limitation}
While our method shows robustness on diverse real-world scenes, we suffer in predicting correct uncertainties for regions with strong view-dependent effects, such as highly reflective surfaces like windows and metals. 
Integrating additional prior knowledge into the optimization process could potentially be beneficial.

\paragraph{Acknowledgements}
We thank the Max Planck ETH Center for Learning Systems (CLS) for supporting Songyou Peng. 
We also thank Yiming Zhao, Yidan Gao and Clément Jambon for helpful discussions.

\newpage
{
    \small
    \bibliographystyle{ieeenat_fullname}
    \bibliography{main}
}

\clearpage
\maketitlesupplementary

\setcounter{section}{0}
\setcounter{figure}{0}
\setcounter{table}{0}
\renewcommand\thesection{\Alph{section}}
\renewcommand\thetable{\Alph{table}}
\renewcommand\thefigure{\Alph{figure}}

In this \textbf{supplementary document}, we first provide additional details in~\secref{sec:supp_implementation}, then, we further provide additional experiment results, more thorough ablation studies and performance analysis in ~\secref{sec:sup_exp}.
We also provide a \textbf{supplementary video} where we show additional visual comparisons.

\section{Details}\label{sec:supp_implementation}

\subsection{Dataset Details}

\paragraph{Synthetic Dataset}
We evaluate on the synthetic dataset~\cite{greff2022kubric} provided in \ddnerf. This dataset includes five sequences with floating objects in the room generated by Kubric~\cite{greff2021kubric}. Upon careful examination, we notice that the training and test images within the \textit{Chair} scene are misaligned in terms of their coordinate systems, therefore we decide to temporarily exclude this particular scene.

\begin{table*}[t]
\resizebox{\linewidth}{!}{
\begin{tabular}{l|ccc|ccc|ccc|ccc}

\multicolumn{1}{c}{}  & \multicolumn{3}{c|}{Car}  & \multicolumn{3}{c|}{Cars} & \multicolumn{3}{c|}{Bag}  & \multicolumn{3}{c}{Pillow} \\
& \lpips & \mssim & \psnr& \lpips & \mssim & \psnr& \lpips & \mssim & \psnr&  \lpips & \mssim & \psnr \\
\midrule
NeRF-W \cite{nerfw} &  0.218 &  0.814 & 24.23 &  0.243 &  0.873 & 24.51 &  0.139 &  0.791 & 20.65 &  0.088 &  0.935 & 28.24 \\
NSFF \cite{nsff} & 0.200 & 0.806 & 24.90 & 0.620 & 0.376 & 10.29 & 0.108 & 0.892 & 25.62 & 0.782 & 0.343 & 4.55 \\
NeuralDiff \cite{neuraldiff}  &  0.065 &  0.952 & 31.89 &  0.098 &  0.921 & 25.93 &  0.117 &  0.910 & 29.02 &  0.565 &  0.652 & 20.09 \\
\ddnerf & 0.062 & 0.975 & 34.27 & 0.090 & 0.953 & 26.27 & 0.076 & 0.979 & 34.14 & 0.076 & 0.979 & 36.58       \\
\RobustNeRF & \bf{0.013} & 0.988 & 37.73 & 0.063 & 0.957 & 26.31 & \bf{0.006} & \bf{0.995} & \bf{41.82} & \bf{0.018} & \bf{0.990} & \bf{38.95} \\ 
\textbf{\ours} & 0.023 & \bf{0.989} & \bf{39.83} & \bf{0.035} & \bf{0.982} & \bf{27.00} & 0.016 & 0.993 & 39.50 & 0.039 & 0.986 & 38.41 \\ 

\end{tabular}}
  \caption{\textbf{Novel view synthesis results on the Kubric Dataset.} The numbers for baseline methods are taken from ~\cite{robustnerf}.}
  \label{table:kubric}
\end{table*}

\paragraph{RobustNeRF Dataset}
As illustrated in the original RobustNeRF, there are unintentional changes throughout the capturing process (both the training and test set) for the dataset, including the tablecloth movement in the {\textit Android} scene and the curtain in the {\textit Statue} scene, which may adversely affect the performance of SAM-based methods. In contrast, both \RobustNeRF and our method can naturally accommodate these unintentional changes.

\paragraph{\textit{On-the-go} Dataset}
\textit{On-the-go} dataset is acquired with an assortment of devices, including an iPhone 12, a Samsung Galaxy S22 and a DJI Mini 3 Pro drone. During the capture of each sequence, the exposure, white balance, and ISO are fixed. This dataset features a wide range of dynamic objects including pedestrians, cyclists, strollers, toys, cars, robots, and trams), along with diverse occlusion ratios ranging from 5\% to 30\%. This diversity ensures a rich and challenging environment for our assessments. The resolution of images captured by the iPhone 12 and DJI drone(\textit{Drone} sequence) is 4032\(\times\)3024, whereas the resolution of sequences captured by the Samsung Galaxy S22(\textit{Arc de Triomphe} and \textit{Patio} sequence) is 1920\(\times\)1080.

\subsection{Implementation Details of NeRF On-the-go}
Our work is built upon the \mipNeRFthreesixty codebase \footnote{\url{https://github.com/google-research/multinerf}}. In addition to our proposed loss, we keep the original distortion loss and interval loss in \mipNeRFthreesixty. 
We run our method on a server with an AMD EPYC 9554 64-core processor and 4 NVIDIA RTX 4090 GPUs. For each scene, we run 250000 iterations with a batch size of 16384, which typically takes 12 hours to finish. Through our assessment, we observed that our model, after only one hour of training, already demonstrated superior quality compared to RobustNeRF, even after it underwent 12 hours of training. We downsample images by 8x to keep it the same as RobustNeRF (except \textit{Arc de Triomphe} and \textit{Patio} is downsampled by 4x to make it roughly the same as RobustNeRF).
We select the dilated sample patches with a size of $32\times32$ and a dilation rate of 4.
The SSIM window size is $5\times5$. For hyperparameters in loss terms, we set \(\lambda_1 = 100, \lambda_2 = 0.5, \lambda_3 = 0.5, \lambda_4 = 0.1\) for all datasets. 

\subsection{Baseline Details}

\paragraph{\RobustNeRF}
For our own run of \RobustNeRF, we enable the appearance embedding (GLO) since it delivers consistently better results as illustrated in \RobustNeRF as shown in \tabref{table:robust}.

\begin{figure*}[t]
    \centering
    \footnotesize
    \setlength{\tabcolsep}{1.5pt}
    \newcommand{\sz}{0.23}
    \newcommand{\sza}{0.1729} %
    \begin{tabular}{cccc}
      \includegraphics[width=\sz\linewidth]{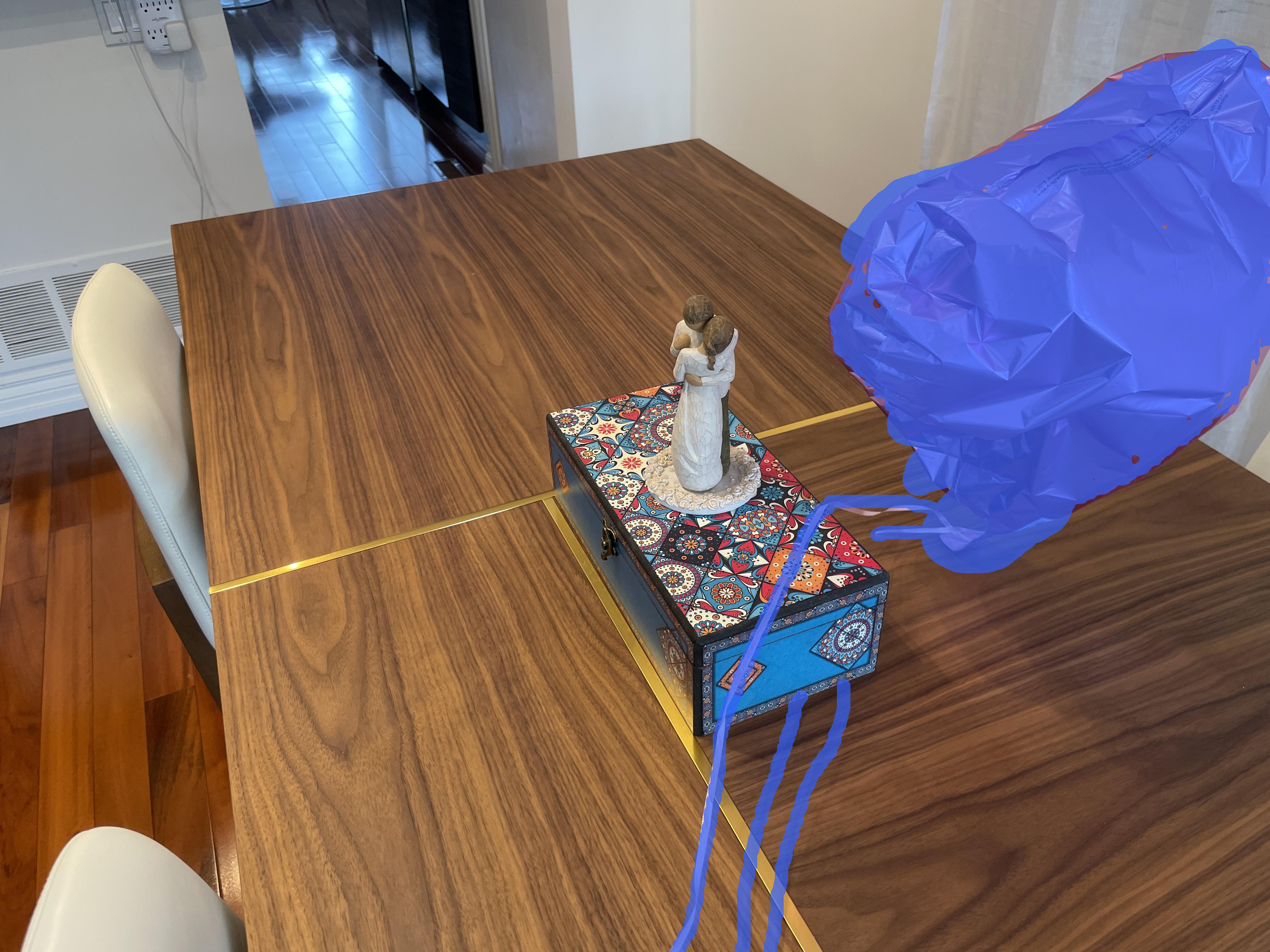} &
      \includegraphics[width=\sz\linewidth]{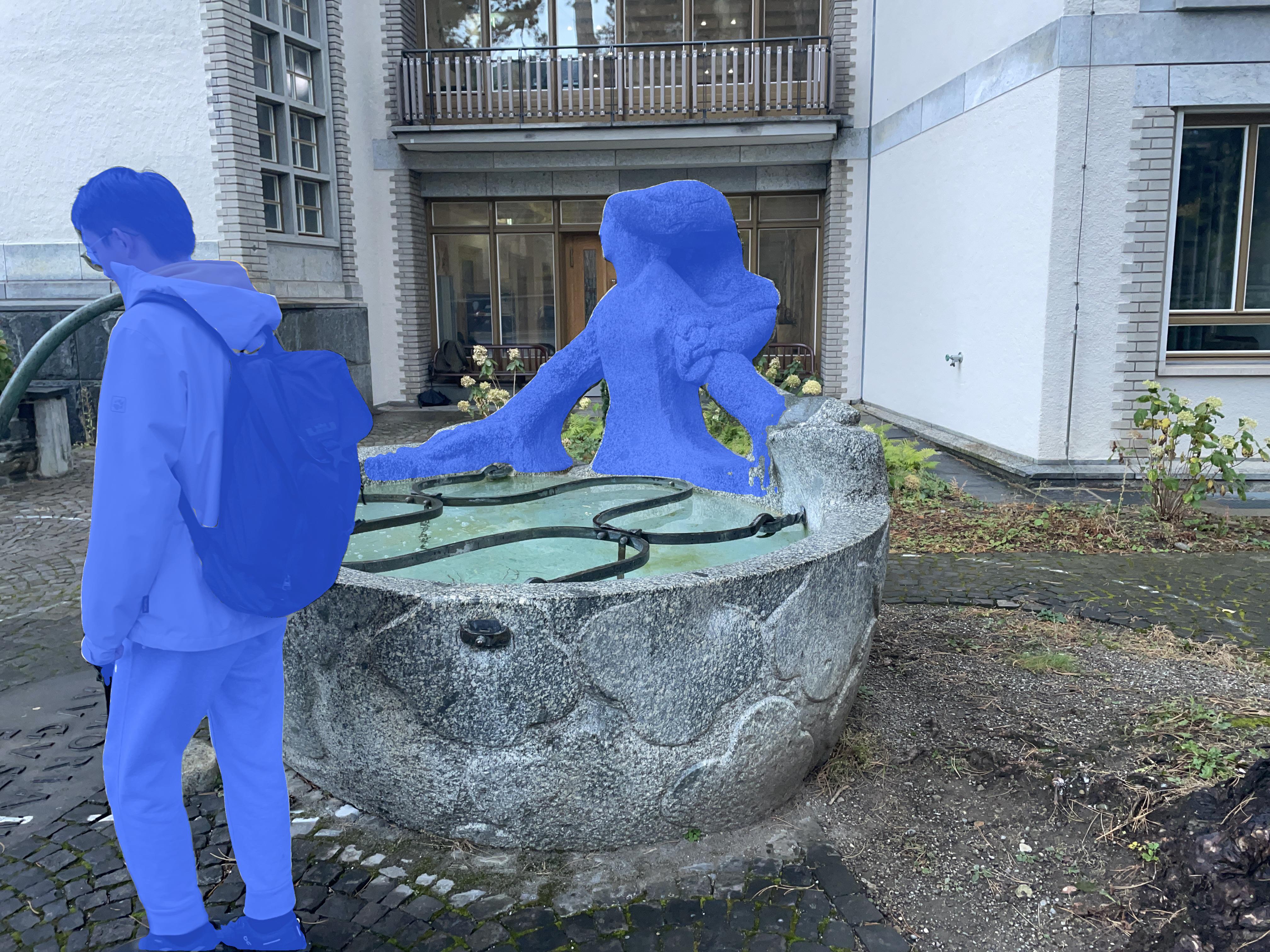} &
      \includegraphics[width=\sz\linewidth]{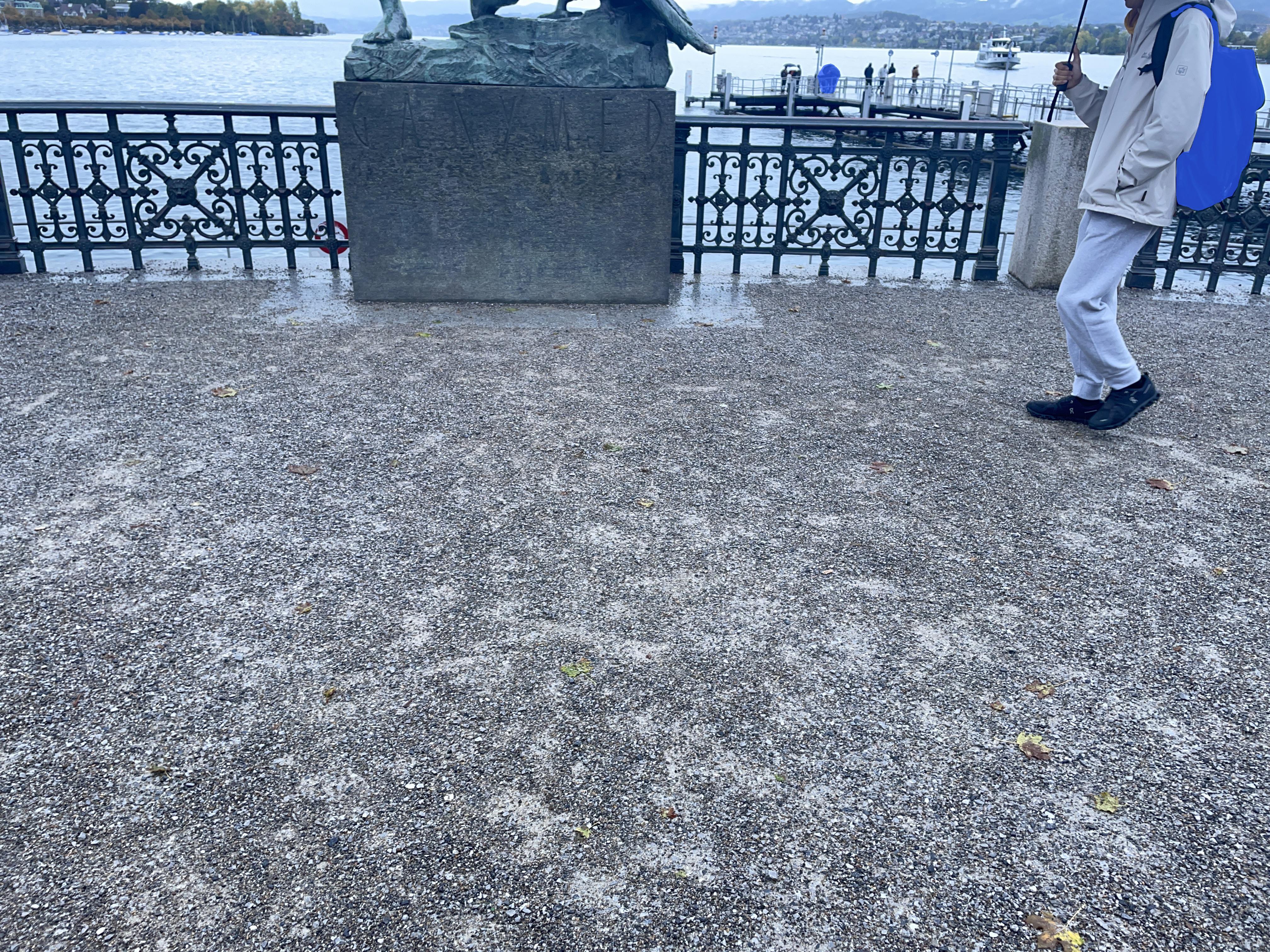} &
      \includegraphics[width=\sz\linewidth]{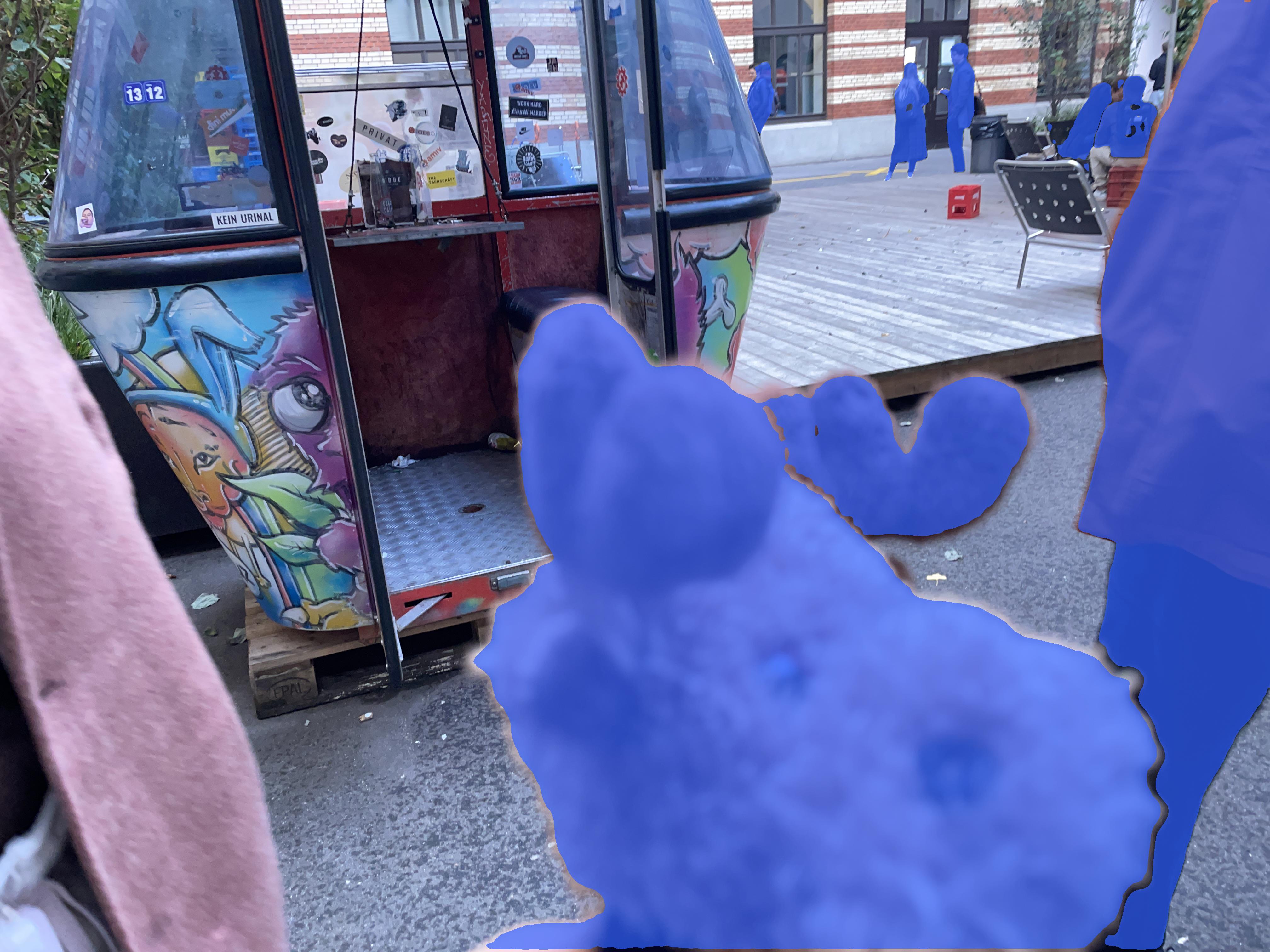} 
       \\
    \end{tabular} 
    \caption{\textbf{Sample Masking Results of \mipNeRFthreesixty + SAM.} The predicted dynamic segments are highlighted in blue. Although state-of-the-art methods for object detection and instant segmentation are used with known dynamic object categories, they still have incorrect predictions, overlooked objects, or incomplete segmentation of objects.
    }
    \label{fig:sam_sample}
  \end{figure*}

\paragraph{Mip-NeRF 360 + SAM}
This is a baseline that we introduce for evaluation.
For \RobustNeRF dataset, we use an interactive tool\footnote{https://github.com/open-mmlab/playground} to click each distractor in every image.
For \textit{On-the-go} dataset, we pre-identify the dynamic objects' categories and consider this as an oracle for this method.  To detect the dynamic object's bounding box, we employed YOLOv8\footnote{\url{https://github.com/ultralytics/ultralytics.git}} to generate the bounding box for the distractors. Following this, Segment Anything Model (SAM)~\cite{kirillov2023segment} is applied with the detected bounding box to get the corresponding segmentation. In the absence of a 'robot' class in YOLOv8, we identify the robot in the {\textit Spot} scene by selecting the bounding box encompassing the largest area of yellow. Some imperfect masking results are shown in \figref{fig:sam_sample}, primarily attributable to factors such as partial observation, reflections of distractors, and ambiguous classifications, like the categorization of a statue as a human.

\section{Additional Experiments}\label{sec:sup_exp}
\subsection{Evaluation}

\paragraph{Kubric Dataset~\cite{greff2022kubric}}
We evaluate on Kubric synthetic dataset provided in \ddnerf, with qualitative results shown in \tabref{table:kubric}. Our performance aligns with RobustNeRF, this is due to saturation on this simple dataset. We include the result of this dataset solely for the sake of a comprehensive evaluation.

\paragraph{Comparison on RobustNeRF Dataset~\cite{robustnerf}}
In this section, we present the results obtained from the {\textit BabyYoda} scene, as summarized in Table~\ref{table:babyyoda}. Our methodology yields improved outcomes compared to the open-source implementation of RobustNeRF. However, these results do not quite reach the performance levels reported in the original RobustNeRF paper. We didn't put this result in the main paper because the distractors in this dataset varies across all images, which doesn't fit our setting.

\begin{table}
  \centering
  \setlength{\tabcolsep}{0.07cm}
  \begin{tabular}{l|ccc }
  & \multicolumn{3}{c}{BabyYoda} 
  \\  & \lpips & \ssim & \psnr \\
  \midrule
  \RobustNeRF & 0.20 & 0.83 & 30.87 \\
  \RobustNeRFStar  &  0.31 & 0.81  & 29.19 \\
  \textbf{\ours}     &  0.24 & 0.83 & 29.96 \\
  \end{tabular}
  \caption{\textbf{Novel View Synthesis Results on the BabyYoda Scene of RobustNeRF dataset.} \RobustNeRFStar denotes our own run using the official code release. Our method is superior compared with \RobustNeRFStar, although it does not quite achieve the results reported in the RobustNeRF paper.
  }
  \label{table:babyyoda}
\end{table}

\begin{figure*}
  \centering
  \footnotesize
  \setlength{\tabcolsep}{0.5pt}
  \newcommand{\sz}{0.15}
  \newcommand{\sza}{0.1729} %
  \begin{tabular}{ccccc|c}
  \multirow{2}{*}{\rotatebox{90}{Arc de Triomphe}} &
    \includegraphics[width=\sz\linewidth]{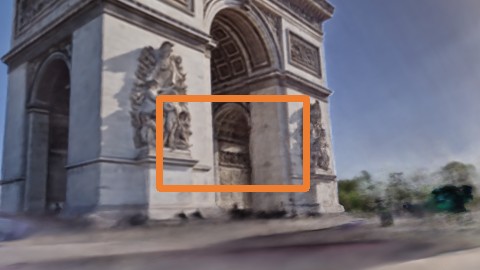} &
    \includegraphics[width=\sz\linewidth]{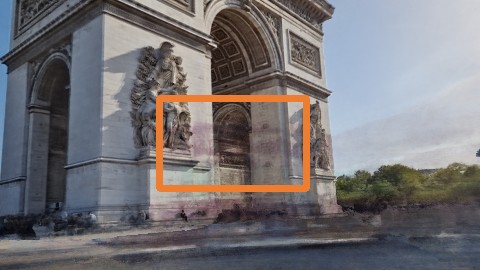} &
    \includegraphics[width=\sz\linewidth]{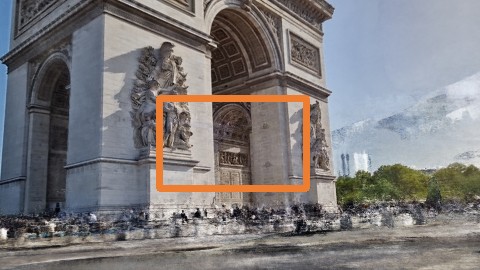} &
    \includegraphics[width=\sz\linewidth]{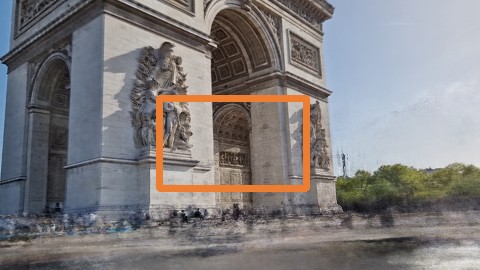}  &
    \includegraphics[width=\sz\linewidth]{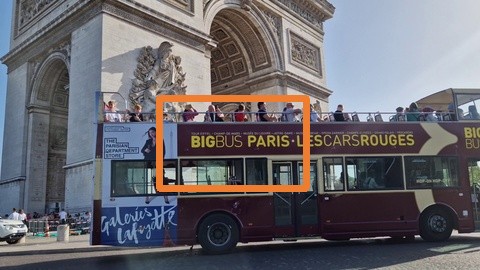} 
     \\
    & 
    \includegraphics[width=\sz\linewidth]{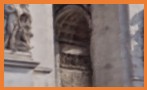} &
    \includegraphics[width=\sz\linewidth]{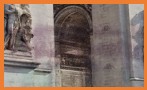} &
    \includegraphics[width=\sz\linewidth]{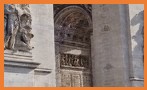} &
    \includegraphics[width=\sz\linewidth]{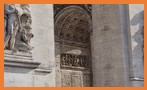}  &
    \includegraphics[width=\sz\linewidth]{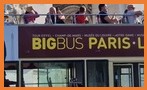} 
    
    \\

    \multirow{2}{*}{\rotatebox{90}{Statue}} &
    \includegraphics[width=\sz\linewidth]{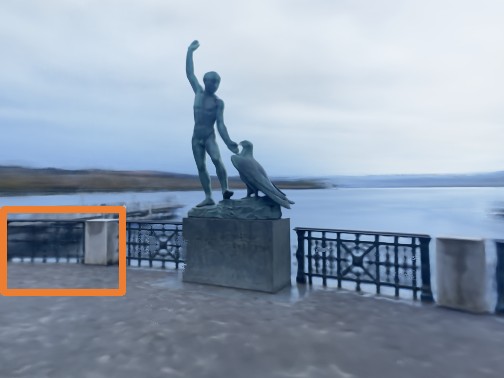} &
    \includegraphics[width=\sz\linewidth]{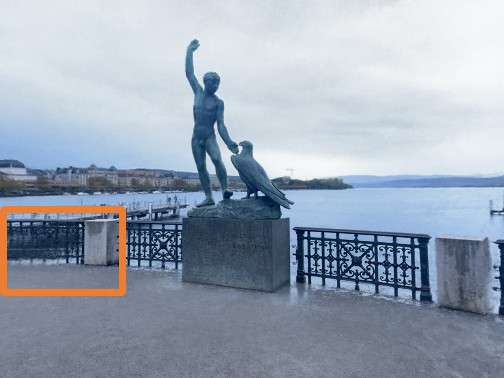} &
    \includegraphics[width=\sz\linewidth]{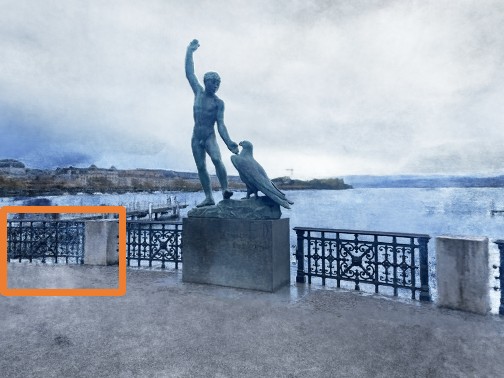} &
    \includegraphics[width=\sz\linewidth]{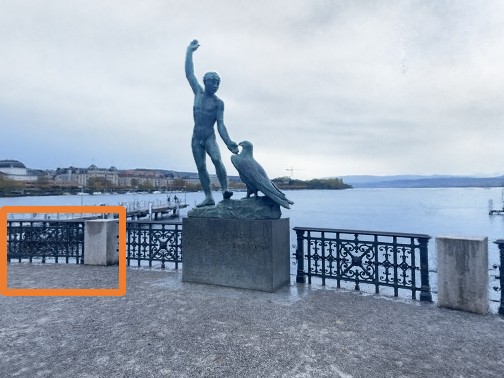}  &
    \includegraphics[width=\sz\linewidth]{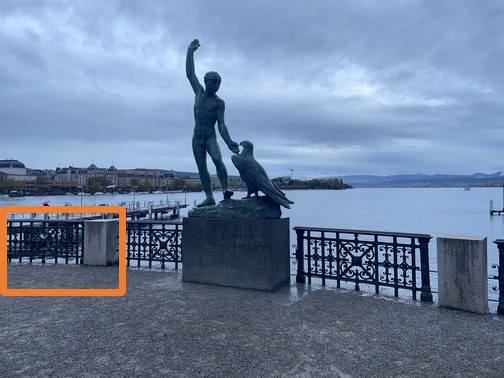} 
     \\
    & 
    \includegraphics[width=\sz\linewidth]{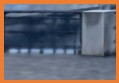} &
    \includegraphics[width=\sz\linewidth]{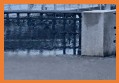} &
    \includegraphics[width=\sz\linewidth]{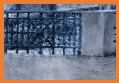} &
    \includegraphics[width=\sz\linewidth]{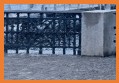}  &
    \includegraphics[width=\sz\linewidth]{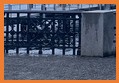} 
    \\
    \multirow{2}{*}{\rotatebox{90}{Drone}} &
    \includegraphics[width=\sz\linewidth]{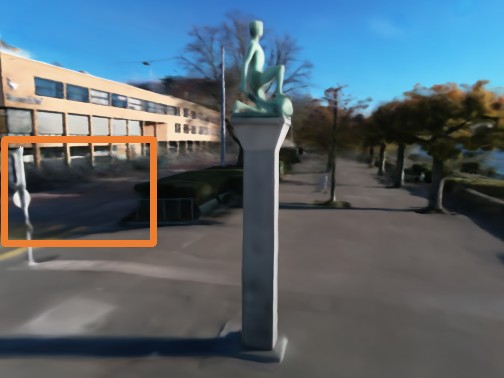} &
    \includegraphics[width=\sz\linewidth]{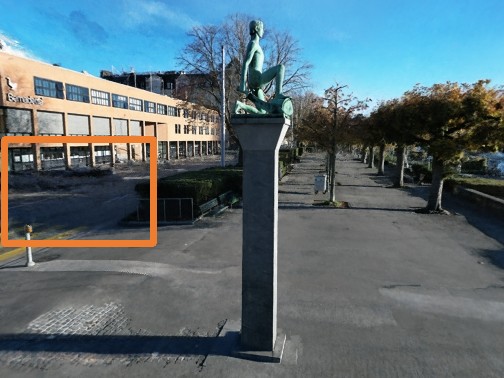} &
    \includegraphics[width=\sz\linewidth]{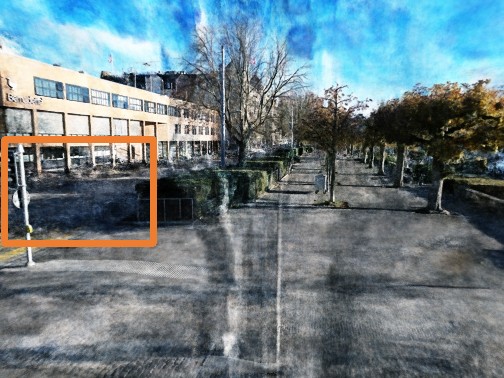} &
    \includegraphics[width=\sz\linewidth]{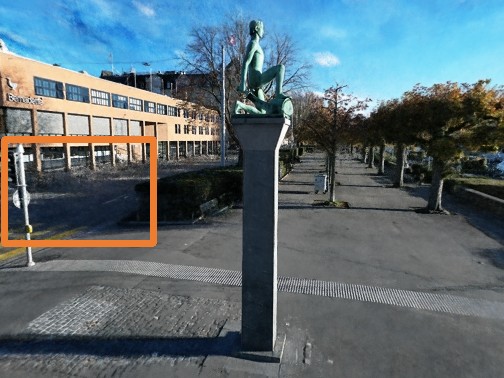}  &
    \includegraphics[width=\sz\linewidth]{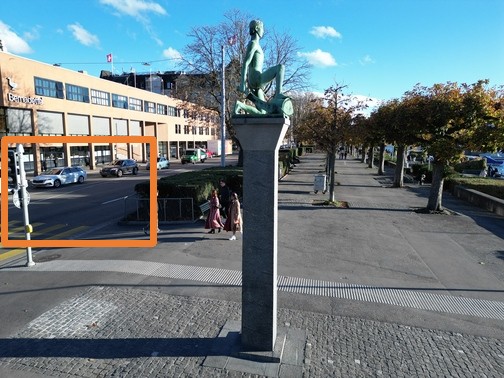} 
     \\
    & 
    \includegraphics[width=\sz\linewidth]{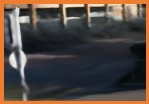} &
    \includegraphics[width=\sz\linewidth]{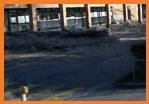} &
    \includegraphics[width=\sz\linewidth]{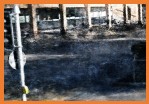} &
    \includegraphics[width=\sz\linewidth]{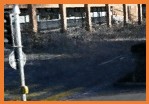}  &
    \includegraphics[width=\sz\linewidth]{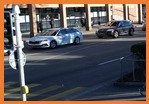} 
    \\
    \multirow{2}{*}{\rotatebox{90}{Station}} &
    \includegraphics[width=\sz\linewidth]{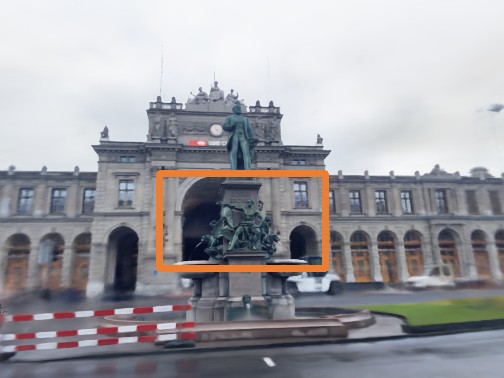} &
    \includegraphics[width=\sz\linewidth]{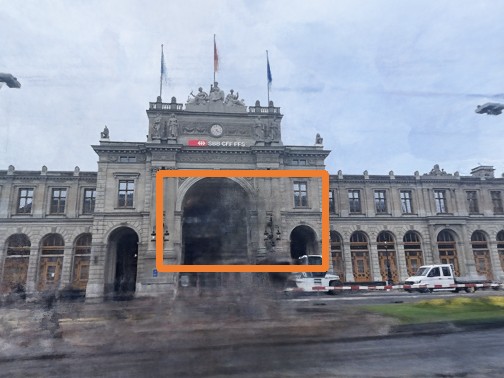} &
    \includegraphics[width=\sz\linewidth]{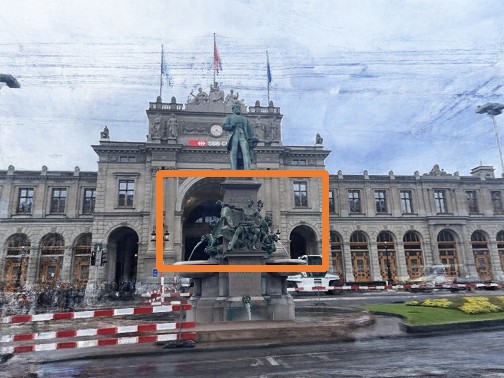} &
    \includegraphics[width=\sz\linewidth]{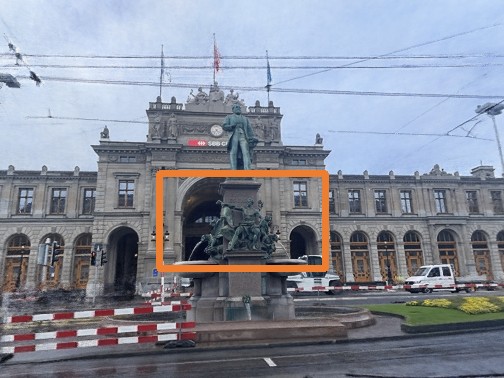}  &
    \includegraphics[width=\sz\linewidth]{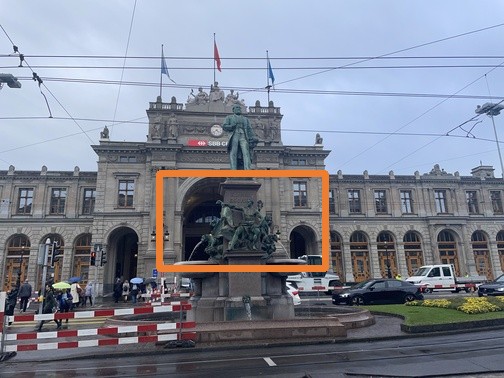} 
     \\
    & 
    \includegraphics[width=\sz\linewidth]{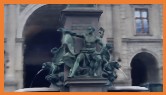} &
    \includegraphics[width=\sz\linewidth]{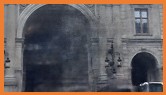} &
    \includegraphics[width=\sz\linewidth]{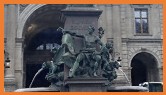} &
    \includegraphics[width=\sz\linewidth]{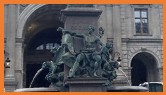}  &
    \includegraphics[width=\sz\linewidth]{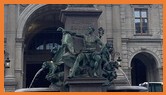} 
    \\
    \multirow{2}{*}{\rotatebox{90}{Tree}} &
    \includegraphics[width=\sz\linewidth]{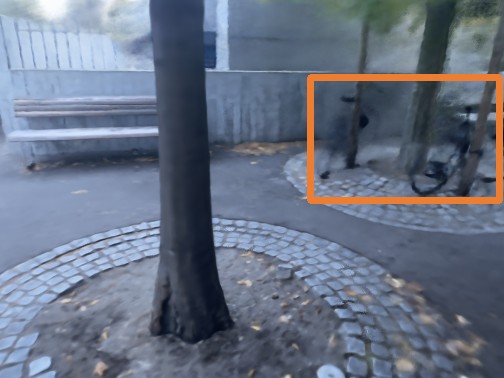} &
    \includegraphics[width=\sz\linewidth]{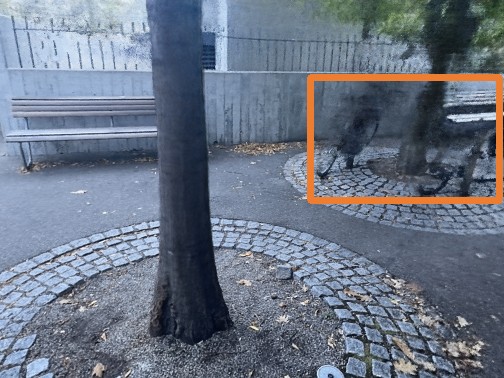} &
    \includegraphics[width=\sz\linewidth]{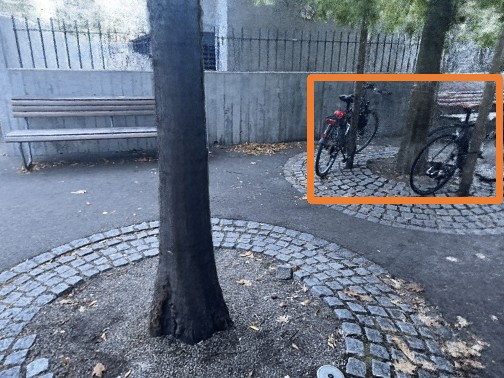} &
    \includegraphics[width=\sz\linewidth]{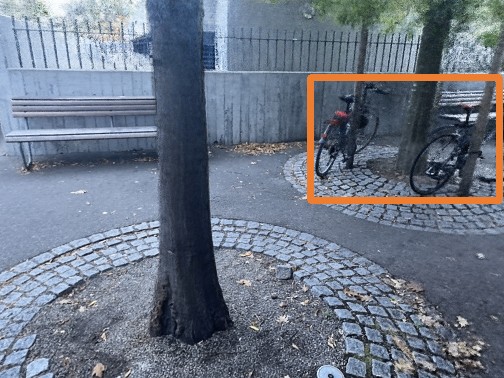}  &
    \includegraphics[width=\sz\linewidth]{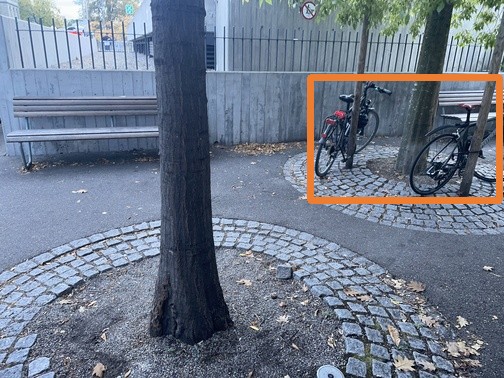} 
     \\
    & 
    \includegraphics[width=\sz\linewidth]{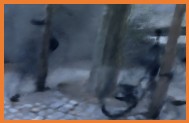} &
    \includegraphics[width=\sz\linewidth]{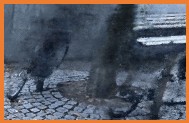} &
    \includegraphics[width=\sz\linewidth]{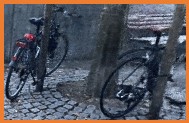} &
    \includegraphics[width=\sz\linewidth]{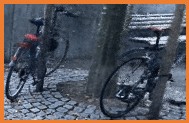}  &
    \includegraphics[width=\sz\linewidth]{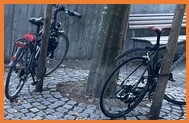} 
    \\
    \multirow{2}{*}{\rotatebox{90}{Train}} &
    \includegraphics[width=\sz\linewidth]{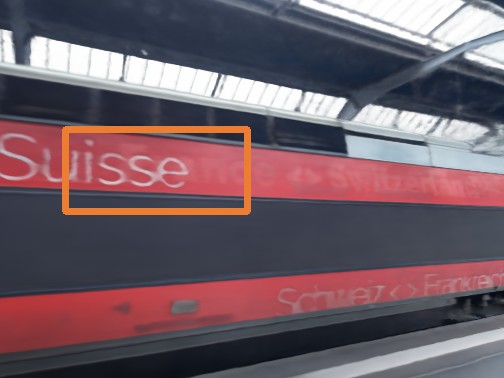} &
    \includegraphics[width=\sz\linewidth]{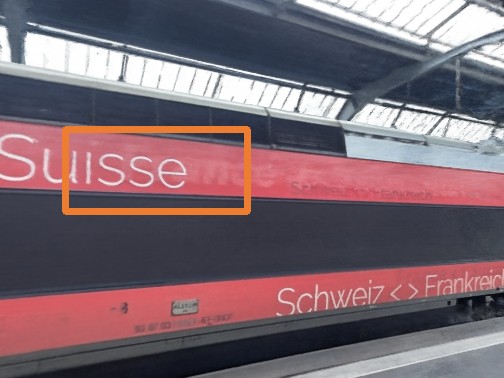} &
    \includegraphics[width=\sz\linewidth]{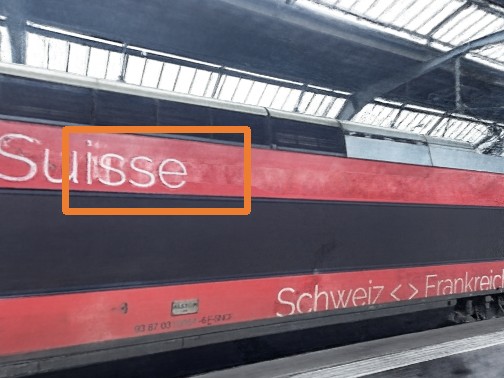} &
    \includegraphics[width=\sz\linewidth]{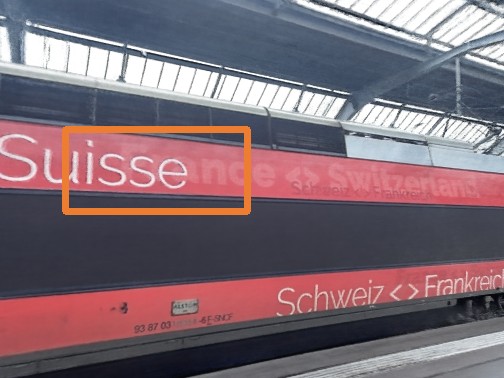}  &
    \includegraphics[width=\sz\linewidth]{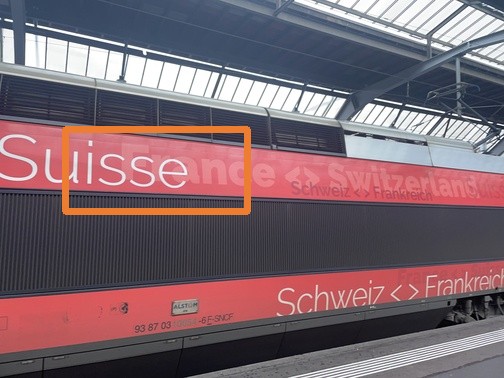} 
     \\
    & 
    \includegraphics[width=\sz\linewidth]{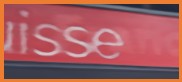} &
    \includegraphics[width=\sz\linewidth]{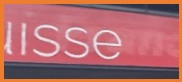} &
    \includegraphics[width=\sz\linewidth]{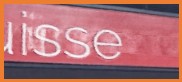} &
    \includegraphics[width=\sz\linewidth]{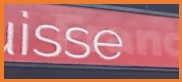}  &
    \includegraphics[width=\sz\linewidth]{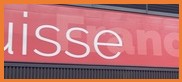} 
    \\
    & {\NeRFW} & {\RobustNeRF} & {Mip-NeRF 360 + SAM} & {\textbf{\ours}}  &  {GT} \\

  \end{tabular} 
  \caption{\textbf{Additional Novel View Synthesis Results on Our \textit{On-the-go} Dataset.} For GT, we show captured test views that might contain some dynamic objects due to restrictions of the capture environment. }
  
  \label{fig:supp_self}
\end{figure*}

\paragraph{\textit{On-the-go} Dataset} Additional qualitative results of \textit{On-the-go} dataset are shown in \figref{fig:supp_self}. Our method consistently outperforms all baseline methods in various environments. The performance of different baseline methods closely aligns with the sequences depicted in the \tabref{fig:self}. While \NeRFW is capable of removing distractors, it does so at the expense of detail loss. \RobustNeRF, due to its threshold-based nature, occasionally fails to preserve thin structures. Furthermore, Mip-NeRF 360 + SAM struggles due to the imperfect segmentation, as illustrated in \figref{fig:sam_sample}.

\subsection{Ablation Study}

\paragraph{Loss Ablation} To evaluate the effectiveness of our loss functions, we conduct a supplementary loss ablation on a low occlusion scene (\textit{Tree}) as presented in \tabref{table:tree_loss}. 
While \tabref{tab:ablation} in the main paper is evaluated on a high occlusion sequence, \tabref{table:tree_loss} is evaluated on a low occlusion sequence.
We find that for both occlusion scenarios, each component of our method contributes to the overall performance enhancement. Although in scenarios with relatively low occlusion, the design choice (b) still can achieve satisfactory quality except for certain views, the performance drop is more pronounced in high occlusion scenarios. Furthermore, in both occlusion scenarios, we observe that (c) $\mathcal{L}_\text{uncer}$ for NeRF exhibits a significant performance decline. This decline can primarily be attributed to our SSIM formulation, which is tailored more toward optimizing uncertainty rather than scene representation.

\begin{table}
    \centering
    \newcommand{\sz}{0.195}
    \newcommand{\sza}{0.1729} %
    \setlength{\tabcolsep}{8pt}
    \resizebox{0.8\linewidth}{!}{\begin{tabular}{l|ccc}
& \lpips & \ssim & \psnr \\
\midrule
(a) w/o $\mathcal{L}_\text{reg}$ &  0.244 & 0.703 & 20.19\\
(b) $\ell_2$ in $\mathcal{L}_\text{uncer}$ & 0.240 & 0.709 & 20.53 \\
(c) $\mathcal{L}_\text{uncer}$ for NeRF & 0.354 & 0.601 & 18.84\\
\textbf{\ours} & \textbf{0.226} & \textbf{0.718} & \textbf{20.68}  \\ 
\end{tabular}}
\caption{\textbf{Ablations on Loss Functions.} We compare different loss choices for training on the {\textit Tree} sequence.}
\label{table:tree_loss}
\end{table}

\paragraph{Dilated Patch Ablation} We continue to test various dilation rates on a low occlusion scene {\textit Tree} in \tabref{table:conv_ssim} with patch size fixed to be \(32\times 32\). 
We observe that the performance closely resembles that of high occlusion scenes as depicted in \tabref{tab:ablation_dilation}.
Notably, unlike in high occlusion situations, a dilation rate of 8 is able to sustain performance without collapsing. Nevertheless, to maintain consistency in hyperparameter settings across all occlusion scenarios, we set the dilation rate at 4.
\begin{table}
    \centering
    \begin{tabular}{l|cccc}
& \lpips & \ssim & \psnr \\
\midrule
1 &  0.363 & 0.592 & 18.51  \\
2 &  0.257 & 0.694 & 20.07 & \\
\textbf{4 (Ours)}  & \textbf{0.226} & \textbf{0.718} & 20.68 &  \\ 
8 & 0.235 & 0.714 & \textbf{20.69}  \\ 
16 & 0.248 & 0.702 & 20.37  \\ 
\end{tabular}
    \caption{\textbf{Ablations on Patch Dilation Rates on the Tree Scene.} Comparisons of various dilated rates for the dilation sampling, with a patch size of $32\times 32$.}
    \label{tab:my_label}
\end{table}

\begin{figure*}
    \centering
    \footnotesize
    \setlength{\tabcolsep}{0.5pt}
    \newcommand{\sz}{0.160}
    \newcommand{\sza}{0.1729} %
    \begin{tabular}{ccccccc} 
    \includegraphics[width=\sz\linewidth]{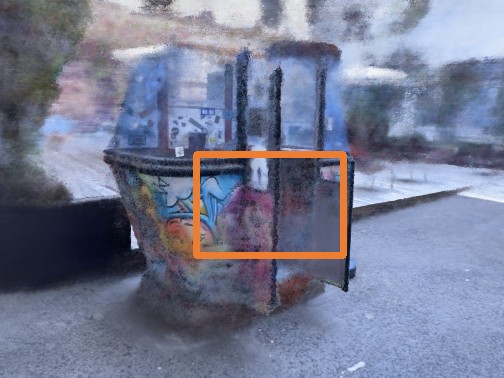} & 
    \includegraphics[width=\sz\linewidth]{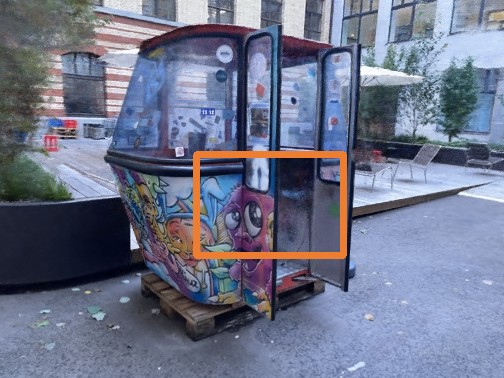} & 
    \includegraphics[width=\sz\linewidth]{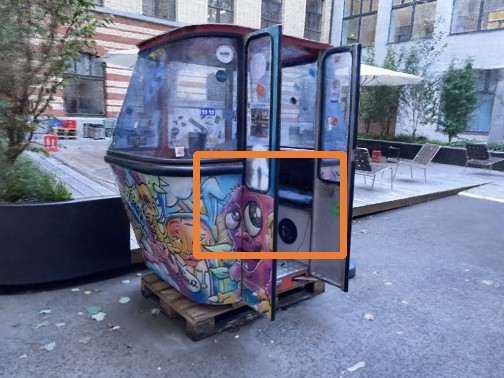} & 
    \includegraphics[width=\sz\linewidth]{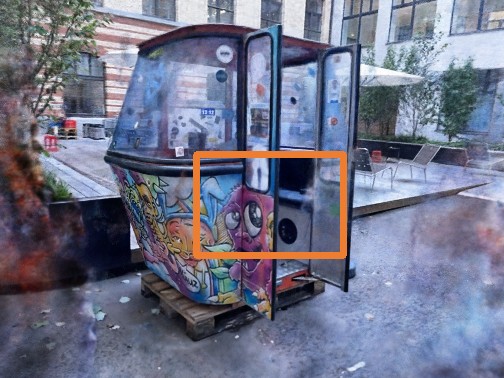} & 
    \includegraphics[width=\sz\linewidth]{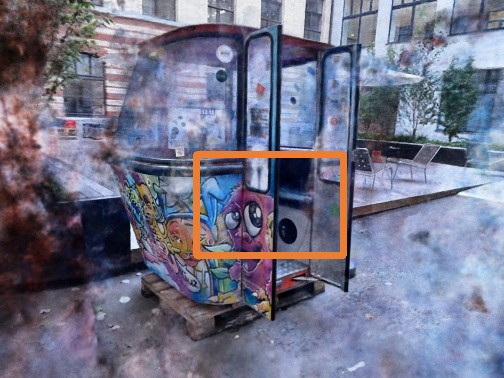} & 
    \includegraphics[width=\sz\linewidth]{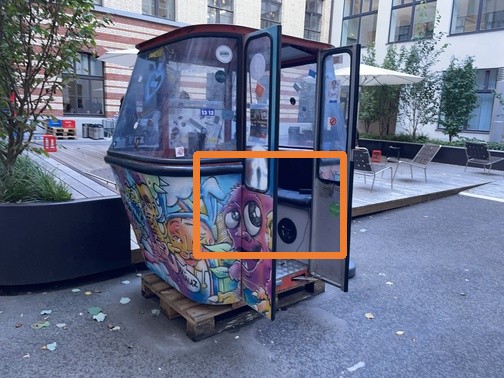} \\
    \includegraphics[width=\sz\linewidth]{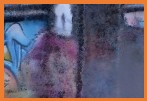} & 
    \includegraphics[width=\sz\linewidth]{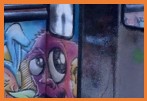} & 
    \includegraphics[width=\sz\linewidth]{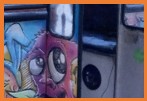} & 
    \includegraphics[width=\sz\linewidth]{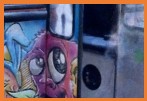} & 
    \includegraphics[width=\sz\linewidth]{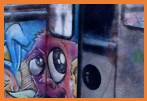} & 
    \includegraphics[width=\sz\linewidth]{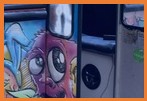} \\

    {1} & {2} & {\textbf{4 (Ours)}} &  {8} & {16} & {GT}\\

    \end{tabular} 
    \caption{\textbf{Ablations on Dilation Rate with a Patch Size at $32\times32$.}  A dilation rate of 4 results in superior rendering quality.  }
    \label{fig:ablation_dilation}
  \end{figure*}

Due to the space constraints in the main paper, the qualitative results of \tabref{tab:ablation_dilation} are shown in  \figref{fig:ablation_dilation}. These qualitative results align with the trends observed in \tabref{tab:ablation_dilation}, indicating that a lower uncertainty ratio \((<4)\) effectively removes distractors but reduces the reconstruction quality, whereas a higher dilation ratio \((>4)\) tends to reintroduce the distractors due to the loss of local information.

\paragraph{Feature Extraction Module}
In this paragraph, we change the feature extractor module \(\mathcal{E}\) to Resnet-50~\cite{he2016identity} and DINOv1~\cite{caron2021emerging} as detailed in \tabref{table:feature}. We find that there are negligible differences between DINOv1 and DINOv2. However, we observe that the Resnet-50 features are less effective in removing distractors. 
We attribute this difference to the Resnet features' emphasis on color information, in contrast to the DINO features that prioritize instance information, essential for efficient distractor removal.

\begin{table}
  \centering
  \setlength{\tabcolsep}{0.1cm}
  \begin{tabular}{l|ccc}
  \\  & \lpips & \ssim & \psnr \\
  \midrule
  ResNet-50  &  0.480 & 0.444  & 16.16 \\
  DINOv1  &  0.237 & \textbf{0.720}  & 21.36 \\
  \textbf{DINOv2 (Ours)}     &   \textbf{0.235} & 0.718 & \textbf{21.41}  \\
  \end{tabular}
  \caption{\textbf{Novel View Synthesis Results with Different Feature Extraction Module.}}
  \label{table:feature}
\end{table}

\subsection{Analysis}

\paragraph{Our SSIM Formulation}
In this section, we will show the mathematical proof that our method can impose a larger uncertainty difference between distractors and static backgrounds. To simplify notation, we denote the $L(P, \hat{P}), C(P, \hat{P}), S(P, \hat{P})$ in \eqnref{eq:ssim} as $l, c, s$.
\begin{proof}
Let \(l_1, c_1, s_1\) represent the luminance, contrast, and structure similarity between the distractor patch and the ground-truth patch. Similarly, \(l_2, c_2, s_2\) represent these similarities for the distractor-free patch and ground truth patch. Therefore, we have the following conditions:

\begin{equation}
    \begin{aligned}\label{eq:range}
        0 < l_1 < l_2 < 1, \\
        0 < c_1 < c_2 < 1, \\
        0 < s_1 < s_2 < 1.
    \end{aligned}
\end{equation}

Our assumptions in \eqnref{eq:range} are directly grounded in the properties proved in the original SSIM paper (Section \uppercase\expandafter{\romannumeral3}.B). In such cases, the similarity between rendered patches and ground truth would naturally decrease. Our empirical results also support this validity: our modified SSIM loss consistently outperforms the original one in various datasets.

To prove that our reformulation in \eqnref{eq:ssim_new} places greater emphasis on the differences between dynamic and static elements compared to \eqnref{eq:ssim}, we need to demonstrate the following inequality:

\begin{equation}\label{eq:inequality}
    \frac{(1-l_1)(1-c_1)(1-s_1)}{(1-l_2)(1-c_2)(1-s_2)} > \frac{1-l_1 \cdot c_1 \cdot s_1}{1-l_2 \cdot c_2 \cdot s_2}.
\end{equation}

The left-hand side of this equation of the ratio of our SSIM formulation between distractors and static backgrounds, and the right-hand side is the ratio of conventional SSIM Loss. This can be equivalently expressed as:

\begin{equation}\label{eq:expanded_inequality}
    \frac{(1-l_1)(1-c_1)(1-s_1)}{1-l_1 \cdot c_1 \cdot s_1} > \frac{(1-l_2)(1-c_2)(1-s_2)}{1-l_2 \cdot c_2 \cdot s_2}.
\end{equation}

Taking the natural logarithm of both sides, we get:

\begin{equation}
\begin{aligned}\label{eq:log_inequality}
    &\ln\left(\frac{(1-l_1)(1-c_1)(1-s_1)}{1-l_1 \cdot c_1 \cdot s_1}\right) > \\
    &\ln\left(\frac{(1-l_2)(1-c_2)(1-s_2)}{1-l_2 \cdot c_2 \cdot s_2}\right).
\end{aligned}
\end{equation}

We aim to prove that the function \(f(x, y, z) = \ln\left(\frac{(1-x)(1-y)(1-z)}{1-xyz}\right)\) is monotonically decreasing for \(0 < x, y, z < 1\). 
Given the function's symmetry across variables, it is sufficient to take the partial derivative with respect to one variable,
say \(x\), and show that it is negative. The partial derivative of \(f(x, y, z)\) with respect to \(x\) is given by:

\begin{equation}
\begin{aligned}\label{eq:partial_derivative}
    \frac{\partial f(x, y, z)}{\partial x} &= -\frac{1}{1-x} + \frac{yz}{1-xyz} \\
    &= \frac{yz - 1}{(1-x)(1-xyz)}.
\end{aligned}
\end{equation}

Given \(0 < x, y, z < 1\), both terms \(1-x\) and \(1-xyz\) are positive. Since \(yz < 1\) (as both \(y\) and \(z\) are less than 1), the numerator \(yz - 1\) is negative. Therefore, the entire expression for \(\frac{\partial f(x, y, z)}{\partial x}\) is less than zero:

\begin{equation}
    \frac{\partial f(x, y, z)}{\partial x} < 0.
\end{equation}

This implies that \(f(x, y, z)\) is monotonically decreasing with respect to \(x\) in the given domain. By the symmetry of \(f\), the same holds for \(y\) and \(z\), completing the proof.

\end{proof}

We compare the effectiveness of the conventional SSIM formulation and our modified SSIM approach in the \textit{Patio-High} scene as shown in \tabref{table:conv_ssim}.
Our SSIM formulation can successfully remove distractors while conventional SSIM fails to do so.
\begin{table}[t]
  \centering
  \setlength{\tabcolsep}{0.1cm}
  \begin{tabular}{l|ccc}
  \\  & \lpips & \ssim & \psnr \\
  \midrule
  Conventional SSIM  &  0.455 & 0.459  & 16.33 \\
  \textbf{\ours}     &   \textbf{0.235} & \textbf{0.718} & \textbf{21.41}  \\
  \end{tabular}
  \caption{\textbf{Novel View Synthesis Results on the Patio-High Scene of \textit{On-the-go} dataset.}}
  \label{table:conv_ssim}
\end{table}

\begin{figure}[]
  \centering
  \includegraphics[width=1\linewidth]{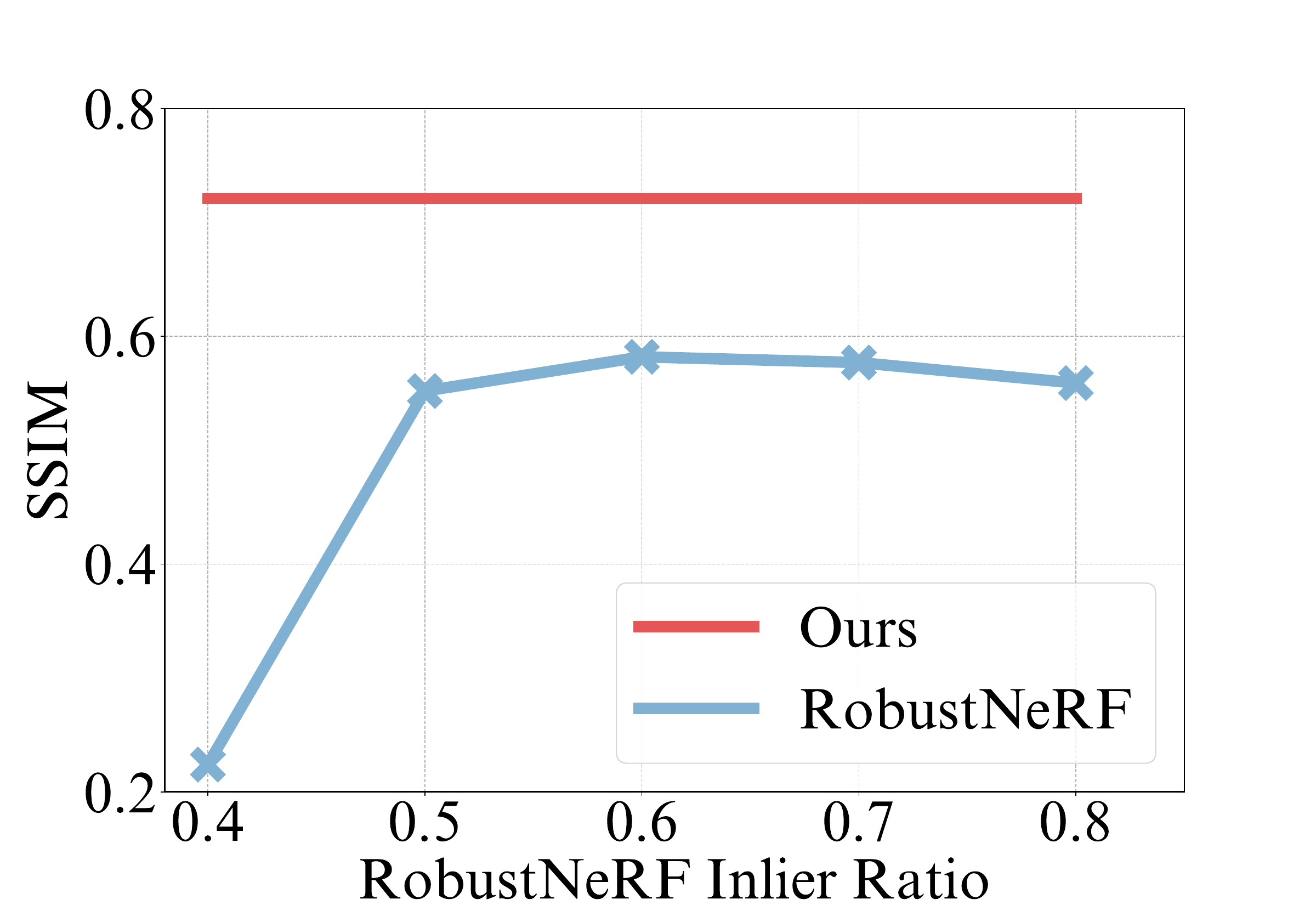}
  \caption{\textbf{The Performance of \RobustNeRF under Different Inlier Ratios Compared to Our Method.
  } 
  }  
  \label{fig:inlier_curve}
\end{figure}

\paragraph{Parameter-tuning Free}
Here we show our method's superiority against \RobustNeRF that no explicit outlier ratio assignment is required for training on scene {\textit Patio-High}. As shown in \figref{fig:inlier_curve}, multiple experiments with different ratios need to be run for \RobustNeRF to gain its best performance. However, our method does not need any hyperparameter tuning and still archives much better results than \RobustNeRF. 

\begin{figure}[]
  \centering
  \includegraphics[width=0.9\linewidth]{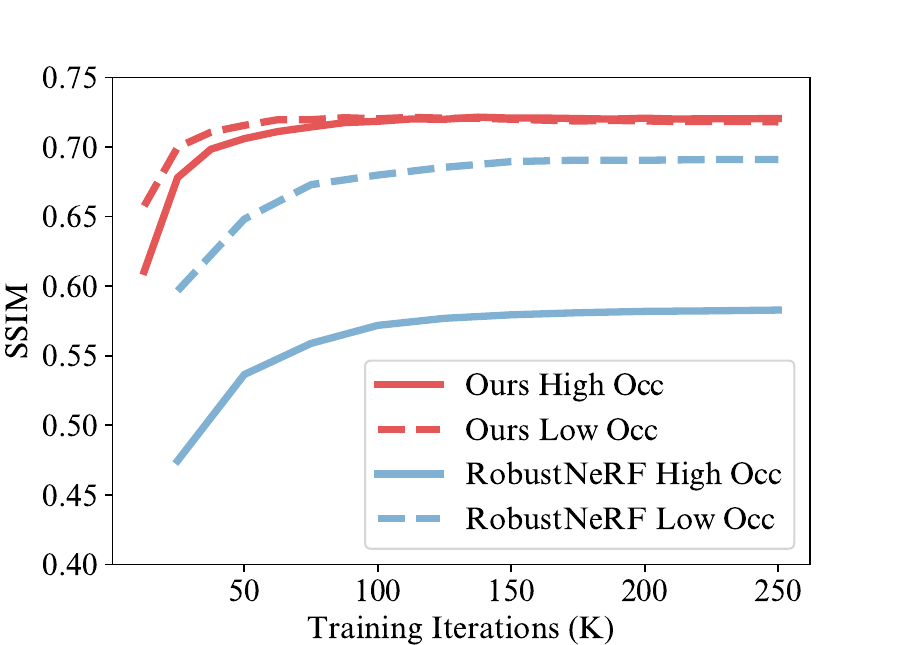}
  \caption{\textbf{SSIM Evaluation Metrics across Training Iterations under Different Occlusion Conditions.}}
  
  \label{fig:converge_curve}
\end{figure}

\paragraph{Fast Convergence}
In \figref{fig:converge_curve}, we show the convergence curve comparison between \RobustNeRF and our method under different occlusion conditions({\textit Tree} and {\textit Patio-High}), using SSIM metrics as the basis for comparison. Our method demonstrates significantly faster convergence — nearly one magnitude faster — and exhibits markedly better performance after reaching convergence.

\begin{figure}[]
  \centering
  \tiny
  \setlength{\tabcolsep}{1.5pt}
  \newcommand{\sz}{0.21}
  \begin{tabular}{cccc}
 
     Mip-NeRF360+SAM & RobustNeRF & Ours & GT \\  
    \includegraphics[width=\sz\linewidth]{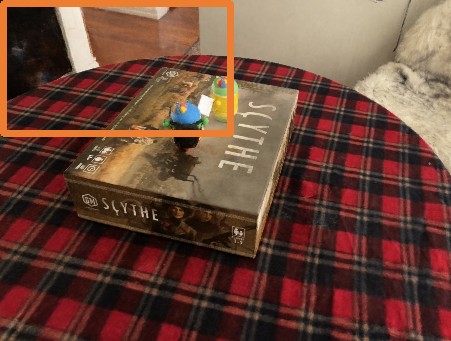} &          
    \includegraphics[width=\sz\linewidth]{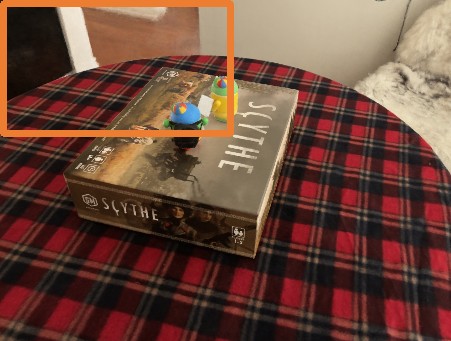} &          
    \includegraphics[width=\sz\linewidth]{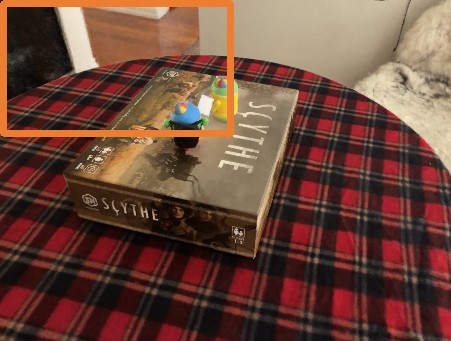} &          
    \includegraphics[width=\sz\linewidth]{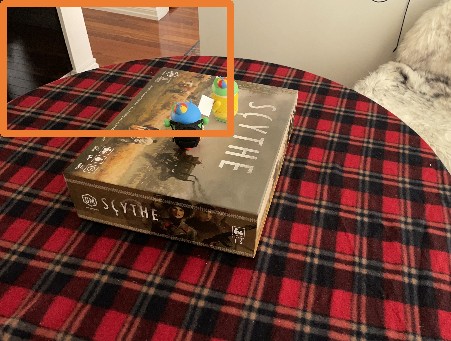}  \\

    \includegraphics[width=\sz\linewidth]{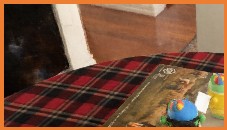} &
    \includegraphics[width=\sz\linewidth]{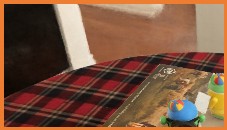} &  
    \includegraphics[width=\sz\linewidth]{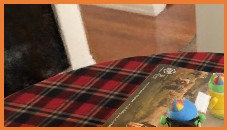} &   
    \includegraphics[width=\sz\linewidth]{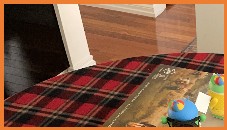} 
  \end{tabular}
  \caption{\textbf{Failure cases.} }
  \label{fig:fail}
\end{figure}

\paragraph{Failure Case}
Similar to baseline methods, we also struggle in regions with strong view-dependent effects, see~\figref{fig:fail}. 
Moreover, inherited from the limitation of our base model Mip-NeRF360, we also require sufficient training views.
Our performance will degrade significantly when the training views become sparse.

\end{document}